\documentclass[10pt,journal,compsoc]{IEEEtran}

\ifCLASSOPTIONcompsoc
  \usepackage[nocompress]{cite}
\else
  \usepackage{cite}
\fi
\ifCLASSINFOpdf
\else
\fi

\usepackage{url}
\usepackage{times}
\usepackage{epsfig}
\usepackage{graphicx}
\usepackage{amsmath}
\usepackage{amssymb}
\usepackage{multirow}
\usepackage{colortbl}
\usepackage{color}
\usepackage{threeparttable}
\usepackage{booktabs}
\usepackage{adjustbox}
\usepackage{subfig}
\usepackage{caption}
\usepackage[misc]{ifsym}
\usepackage{xcolor,colortbl}
\usepackage{makecell}
\newcommand*{\belowrulesepcolor}[1]{%
  \noalign{%
    \kern-\belowrulesep 
    \begingroup 
      \color{#1}%
      \hrule height\belowrulesep 
    \endgroup 
    \vspace{-0.03mm}
  }%
} 
\newcommand*{\aboverulesepcolor}[1]{%
  \noalign{%
  \vspace{-0.03mm}
    \begingroup 
      \color{#1}%
      \hrule height\aboverulesep 
    \endgroup 
    \kern-\aboverulesep 
  }%
}
\newcommand{\vspacefigtext}{\vspace{-3mm}}

\usepackage{pifont}
\newcommand{\cmark}{\ding{51}}%
\newcommand{\xmark}{\ding{55}}%

\usepackage{xspace}
\makeatletter
\DeclareRobustCommand\onedot{\futurelet\@let@token\@onedot}
\def\@onedot{\ifx\@let@token.\else.\null\fi\xspace}
\def\eg{\emph{e.g}\onedot} 
\def\ie{\emph{i.e}\onedot} 
 
\def\etc{\emph{etc}\onedot} 
 
\def\etal{\emph{et al}\onedot}

\usepackage[linesnumbered,lined,boxed,commentsnumbered,ruled]{algorithm2e}
\usepackage[pagebackref,breaklinks=true,colorlinks,citecolor=blue,urlcolor=blue,linkcolor=blue,bookmarks=false]{hyperref}
\def\etal{\textit{et al.}}
\begin{document}
%

\title{Transformer-Based Visual Segmentation:\\ A Survey}

%
%
%
%
\author{Xiangtai Li,
        Henghui Ding,
        Haobo Yuan,
        Wenwei Zhang,
        Jiangmiao Pang, \\
        Guangliang Cheng, 
        Kai Chen,
        Ziwei Liu,
        Chen Change Loy
\IEEEcompsocitemizethanks{\IEEEcompsocthanksitem X. Li, H. Ding, W. Zhang, H. Yuan, and C. Loy are with the S-Lab, Nanyang Technological University, Singapore. xiangtai94@gmail.com, henghui.ding@gmail.com, \{wenwei.zhang, ccloy\}@ntu.edu.sg. 
\IEEEcompsocthanksitem J. Pang and K. Chen are with Shanghai AI Laboratory, Shanghai, China. pangjiangmiao@gmail.com, chenkai@pjlab.org.cn.
\IEEEcompsocthanksitem G. Cheng is with the University of Liverpool, UK.
\IEEEcompsocthanksitem Corresponding: Guangliang Cheng.(Guangliang.Cheng@liverpool.ac.uk)
}
}

%
%

\markboth{IEEE TRANSACTIONS ON PATTERN ANALYSIS AND MACHINE INTELLIGENCE}
{Shell \MakeLowercase{\textit{et al.}}: Bare Advanced Demo of IEEEtran.cls for IEEE Computer Society Journals}
%




\IEEEtitleabstractindextext{
\begin{abstract}
Visual segmentation seeks to partition images, video frames, or point clouds into multiple segments or groups. This technique has numerous real-world applications, such as autonomous driving, image editing, robot sensing, and medical analysis. Over the past decade, deep learning-based methods have made remarkable strides in this area. Recently, transformers, a type of neural network based on self-attention originally designed for natural language processing, have considerably surpassed previous convolutional or recurrent approaches in various vision processing tasks. Specifically, vision transformers offer robust, unified, and even simpler solutions for various segmentation tasks. This survey provides a thorough overview of transformer-based visual segmentation, summarizing recent advancements. We first review the background, encompassing problem definitions, datasets, and prior convolutional methods. Next, we summarize a meta-architecture that unifies all recent transformer-based approaches. Based on this meta-architecture, we examine various method designs, including modifications to the meta-architecture and associated applications. We also present several specific subfields, including 3D point cloud segmentation, foundation model tuning, domain-aware segmentation, efficient segmentation, and medical segmentation. Additionally, we compile and re-evaluate the reviewed methods on several well-established datasets. Finally, we identify open challenges in this field and propose directions for future research. The project page can be found at \url{https://github.com/lxtGH/Awesome-Segmentation-With-Transformer}. 
\end{abstract}

}

\maketitle

\IEEEdisplaynontitleabstractindextext

%
\IEEEpeerreviewmaketitle

\section{Introduction}

\IEEEPARstart{V}{isual} segmentation aims to group pixels of the given image or video into a set of semantic regions. It is a fundamental problem in computer vision and involves numerous real-world applications, such as robotics, automated surveillance, image/video editing, social media, autonomous driving, \emph{etc}. 
Starting from the hand-crafted features~\cite{malik2001contour,shi2000normalized} and classical machine learning models~\cite{stella2003multiclass,schroff2008object,kass1988snakes}, segmentation problems have been involved with a lot of research efforts. During the last ten years, deep neural networks, Convolution Neural Networks (CNNs)~\cite{russakovsky2015imagenet,resnet,simonyan2014very}, such as Fully Convolutional Networks (FCNs)~\cite{long2015fully,chen2017deeplab,zhao2017pyramid,ding2018context} have achieved remarkable successes for different segmentation tasks and led to much better results. Compared to traditional segmentation approaches, CNNs based approaches have better generalization ability. Because of their exceptional performance, CNNs and FCN architecture have been the basic components in the segmentation research works.

Recently, with the success of natural language processing (NLP), transformer~\cite{vaswani2017attention} is introduced as a replacement for recurrent neural networks~\cite{LSTM}. 
Transformer contains a novel self-attention design and can process various tokens in parallel.
Then, based on transformer design, BERT~\cite{BERT} and GPT-3~\cite{GPT3} scale the model parameters up and pre-train with huge unlabeled text information. They achieve strong performance on many NLP tasks, accelerating the development of transformers into the vision community. Recently, researchers applied transformers to computer vision (CV) tasks. 
Early methods~\cite{zhao2018psanet,wang2018nonlocal} combine the self-attention layers to augment CNNs. 
Meanwhile, several works~\cite{zhao2020exploring,hu2019local} used pure self-attention layers to replace convolution layers. 
After that, two remarkable methods boost the CV tasks. One is \textit{vision transformer (ViT)}~\cite{VIT}, which is a pure transformer that directly takes the sequences of image patches to classify the full image. It achieves state-of-the-art performance on multiple image recognition datasets. 
Another is \textit{detection transformer (DETR)}~\cite{detr}, which introduces the concept of object query. Each object query represents one instance. The object query replaces the complex anchor design in the previous detection framework, which simplifies the pipeline of detection and segmentation. Then, the following works adopt improved designs on various vision tasks, including representation learning~\cite{liu2021swin,MaskedAutoencoders2021}, object detection~\cite{zhu2020deformabledetr}, segmentation~\cite{wang2020maxDeeplab}, low-level image processing~\cite{chen2021pre}, video understanding~\cite{bertasius2021space}, 3D scene understanding~\cite{point_transformer}, and image/video generation~\cite{pan20213d}. 

As for visual segmentation, recent state-of-the-art methods are all based on transformer architecture. Compared with CNN approaches, most transformer-based approaches have simpler pipelines but stronger performance. Because of a rapid upsurge in transformer-based vision models, there are several surveys on vision transformer~\cite{han2022survey,khan2022transformers,lin2022survey}. However, most of them mainly focus on general transformer design and its application on several specific vision tasks~\cite{selva2022videotransformer_survey,lahoud20223d_transformer_survey,xu2022multimodal}. Meanwhile, there are previous surveys on the deep-learning-based segmentation~\cite{minaee2021image,hao2020brief,zhou2023survey}. However, to the best of our knowledge, there are \textit{no surveys} focusing on using vision transformers for visual segmentation or query-based object detection. We believe it would be beneficial for the community to summarize these works and keep tracking this evolving field.

\begin{figure*}[!t]
	\centering
	\includegraphics[width=0.95\linewidth]{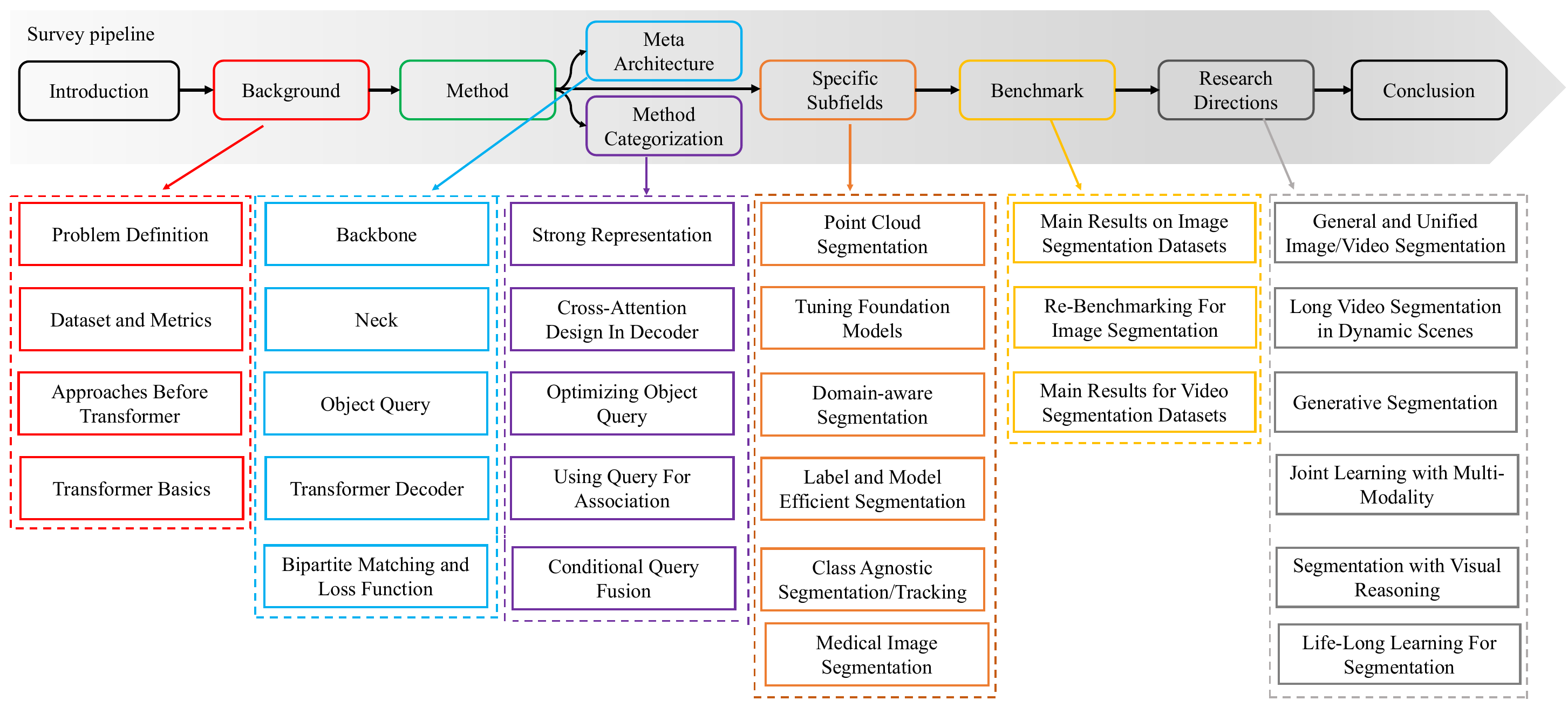}
	\caption{A diagram that summarizes this survey. Different colors represent specific sections. Best viewed in color.}
	\label{fig:survey_pipline}
\end{figure*}

\vspace{0.5cm}
\noindent$\bullet$
\textbf{Contribution.}
In this survey, we systematically introduce recent advances in transformer-based visual segmentation methods. 
We start by defining the task, datasets, and CNN-based approaches and then move on to transformer-based approaches, covering existing methods and future work directions. 
Our survey groups existing representative works from a more technical perspective of the method details. 
In particular, for the main review part, we first summarize the core framework of existing approaches into a meta-architecture in Sec.~\ref{sec:method_meta}, which is an extension of DETR~\cite{detr}. 
By changing the components of the meta-architecture, we divide existing approaches into six categories in Sec.~\ref{sec:method_categorization}, including Representation Learning, Interaction Design in Decoder, Optimizing Object Query, Using Query For Association, and Conditional Query Generation.

Moreover, we also survey closely related specific subfields, including point cloud segmentation, tuning foundation models, domain-aware segmentation, data/model efficient segmentation, class agnostic segmentation and tracking, and medical segmentation. 
We also evaluate the performance of influential works published in top-tier conferences and journals on several widely used segmentation benchmarks. 
Additionally, we provide an overview of previous CNN-based models and relevant literature in other areas, such as object detection, object tracking, and referring segmentation in the background section.

\noindent$\bullet$
\textbf{Scope.} This survey will cover several mainstream segmentation tasks, including semantic segmentation, instance segmentation, panoptic segmentation, and their variants, such as video and point cloud segmentation. 
Additionally, we cover related subfields in Sec.~\ref{sec:method_downstream_beyond}. 
We focus on transformer-based approaches and only review a few closely related CNN-based approaches for reference. 
Although there are many preprints or published works, we only include the most representative works.

\noindent$\bullet$
\textbf{Organization.} The rest of the survey is organized as follows.
Overall, Fig.~\ref{fig:survey_pipline} shows the pipeline of our survey. 
We first introduce the background knowledge on problem definition, datasets, and CNN-based approaches in Sec.~\ref{sec:background}.
Then, we review representative papers on transformer-based segmentation methods in Sec.~\ref{sec:method_survey} and Sec.~\ref{sec:method_downstream_beyond}. 
We compare the experiment results in Sec.~\ref{sec:benchmark}.
Finally, we raise the future directions in Sec.~\ref{sec:future_work} and conclude the survey in Sec.~\ref{sec:conclusion}. 
We provide more benchmarks and details in the appendix.

\vspace{-2mm}
\section{Background}
\label{sec:background}
%
In this section, we first present a unified problem definition of different segmentation tasks. Then, we detail the common datasets and evaluation metrics. Next, we present a summary of previous approaches before the transformer. Finally, we present a review of basic concepts in transformers. To facilitate understanding of this survey, we list the brief notations in Tab.~\ref{tab:conception_notation} for reference.

\begin{table}
\centering
\tiny
\caption{Notation and abbreviations used in this survey.}
\label{tab:conception_notation}
\begin{adjustbox}{width=0.50\textwidth}
\begin{tabular}{c c c }
\toprule[0.15em]
\textbf{Notations}  & Descriptions \\ 
\midrule[0.15em]
SS / IS /PS &  Semantic Segmentation / Instance Segmentation / Panoptic Segmentation \\
VSS / VIS / VPS & Video Semantic / Instance / Panoptic Segmentation \\
DVPS &  Depth-aware Panoptic Segmentation \\
PPS & Part-aware Panoptic Segmentation \\
PCSS / PCIS / PCPS & Point Cloud Semantic /Instance / Panoptic Segmentation  \\
RIS / RVOS & Referring Image  Segmentation / Referring Video Object Segmentation \\
VLM & Vision Language Model  \\
VOS / (V)OD & Vido Object Segmentation / (Video) Object Detection \\
MOTS & Multi-Object Tracking, and Segmentation \\
\hline
CNN / ViTs & Convolution Neural Network / Vision Transformer \\
SA / MHSA / MLP & Self-Attention / Multi-Head Self Attention / Multi-Layer Perceptron  \\
(Deformable) DETR & (Deformable)  DEtection TRansformer  \\
\hline
mIoU & mean Intersection over Union (SS, VSS) \\
PQ/VPQ & Panoptic Quality / Video Panoptic Quality (PS, VPS) \\
mAP & mean Average Precision (IS, VIS) \\
STQ & Segmentation and Tracking Quality (VPS)\\
\bottomrule
\end{tabular}
\end{adjustbox}
\end{table}

\subsection{Problem Definition}
\label{sec:problem_defination}

\begin{figure}[!t]
	\centering
	\includegraphics[width=0.95\linewidth]{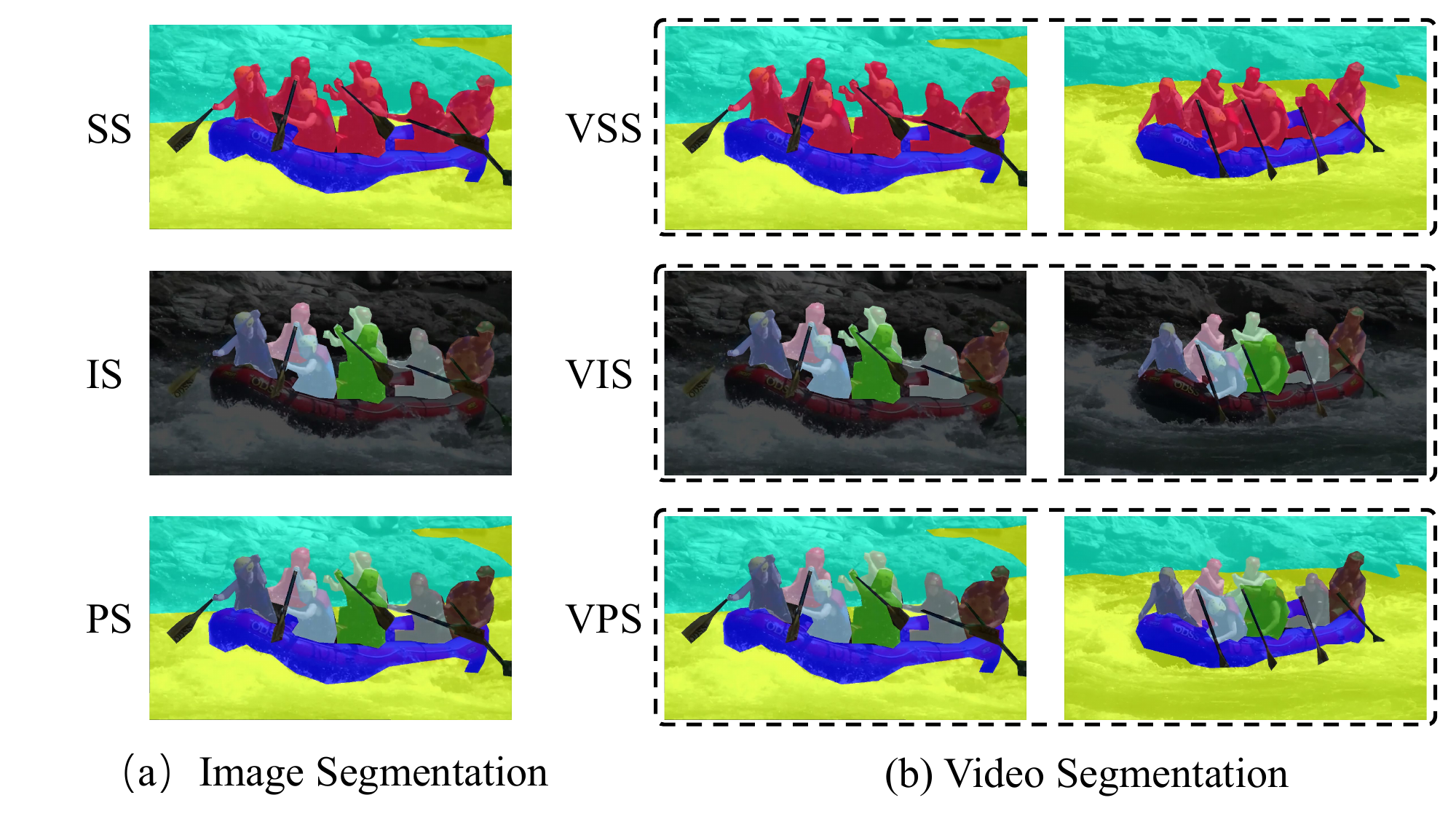}
	\caption{Illustration of different segmentation tasks. The examples are sampled from the VIP-Seg dataset~\cite{miao2022large}. For (V)SS, the same color indicates the same class. For (V)IS and (V)PS, different instances are represented by different colors.}\vspace{-3mm}
    \label{fig:seg_tasks}
\end{figure}

\noindent$\bullet$
\textbf{Image Segmentation.} Given an input image $ I \in \mathbb{R}^{H\times {W}\times 3}$, the goal of image segmentation is to output a group of masks $\{y_i\}_{i=1}^G = \{(m_i, c_i)\}_{i=1}^G \,$ where $c_i$ denotes the ground truth class label of the binary mask $m_i$ and $G$ is the number of masks, ${H}\times {W}$ are the spatial size. According to the scope of class labels and masks, image segmentation can be divided into three different tasks, including semantic segmentation (SS), instance segmentation (IS), and panoptic segmentation (PS), as shown in Fig.~\ref{fig:seg_tasks} (a). For SS, the classes may be foreground objects (thing) or background (stuff), and each class only has one binary mask that indicates the pixels belonging to this class.
Each SS mask does not overlap with other masks. For IS, each class may have more than one binary mask, and all the classes are foreground objects. Some IS masks may overlap with others. For PS, depending on the class definition, each class may have a different number of masks. For the countable thing class, each class may have multiple masks for different instances. For the uncountable stuff class, each class only has one mask. Each PS mask does not overlap with other masks. One can understand image segmentation from the pixel view. 
Given an input $I\in\mathbb{R}^{H\times {W}\times 3}$, the output of image segmentation is a two-channel 
dense segmentation map $S= \{k_{j},c_{j}\}_{j=1}^{H\times W}$. In particular, $k$ indicates the identity of the pixel $j$, and $c$ means the class label of pixel $j$. 
For SS, the identities of all pixels are zero. 
For IS, each instance has a unique identity. 
For PS, the pixels belonging to the thing classes have a unique identity. The pixel identities of the stuff class are zero. 
From both two perspectives, the PS unifies both SS and IS. We present the visual examples in Fig.~\ref{fig:seg_tasks}.


\noindent$\bullet$
\textbf{Video Segmentation.} Given a video clip input as $ V \in \mathbb{R}^{T\times H\times {W}\times 3}$, where $T$ represents the frame number, the goal of video segmentation is to obtain a mask tube $\{y_i\}_{i=1}^N = \{(m_i, c_i)\}_{i=1}^N \,$, where $N$ is the number of the tube masks $m_i \in {\{0,1\}}^{{T}\times {H}\times {W}}$, and $c_i$ denotes the class label of the tube $m_i$. 
Video panoptic segmentation (VPS) requires temporally consistent segmentation and tracking results for each pixel. 
Each tube mask can be classified into countable thing classes and countless stuff classes. Each thing tube mask also has a unique ID for evaluating tracking performance. For stuff masks, the tracking is zero by default. When $N=C$ and the task only contains stuff classes, and all thing classes have no IDs, VPS turns into video semantic segmentation (VSS). If ${\{y_i\}_{i=1}^N}$ overlap and $C$ only contains the thing classes and all stuff classes are ignored, VPS turns into video instance segmentation (VIS). We present the visual examples that summarize the difference among VPS, VIS, and VSS with $T=2$ in Fig.~\ref{fig:seg_tasks} (b). 


\noindent$\bullet$
\textbf{Related Problems.} Object detection and instance-wise segmentation (IS/VIS/VPS) are closely related tasks. 
Object detection involves predicting object bounding boxes, which can be considered a coarse form of IS. 
After introducing the DETR model, many works have treated object detection and IS as the same task, as IS can be achieved by adding a simple mask prediction head to object detection. 
Similarly, video object detection (VOD) aims to detect objects in every video frame.
In our survey, we also examine query-based object detectors for both object detection and VOD.
Point cloud segmentation is another segmentation task, where the goal is to segment each point in a point cloud into pre-defined categories. 
We can apply the same definitions of semantic segmentation, instance segmentation, and panoptic segmentation to this task, resulting in point cloud semantic segmentation (PCSS), point cloud instance segmentation (PCIS), and point cloud panoptic segmentation (PCPS). 
Referring segmentation is a task that aims to segment objects described in natural language text input. There are two subtasks in referring segmentation: referring image segmentation (RIS), which performs language-driven segmentation, and referring video object segmentation (RVOS), which segments and tracks a specific object in a video based on required text inputs. 
Finally, video object segmentation (VOS) involves tracking an object in a video by predicting pixel-wise masks in every frame, given a mask of the object in the first frame.

\subsection{Datasets and Metrics} 
\label{sec:dataset_metric}

\begin{table*}[t]
\centering
\tiny
\caption{Commonly used datasets and metric for Transformer-based segmentation}\vspace{-2mm}
\begin{adjustbox}{width=1.\textwidth}
\begin{tabular}{c c c c c c} 
  \toprule[0.15em]
     Dataset &  Samples (train/val) &  Task  & Evaluation Metrics & Characterization \\
    \toprule[0.15em]
     Pascal VOC~\cite{pascalvoc} & 1,464 / 1,449 & SS & mIoU & PASCAL Visual Object Classes (VOC) 2012 dataset contains 20 object categories. \\
     Pascal Context~\cite{pascalcontext} & 4,998 / 5,105 & SS & mIoU & PASCAL Context dataset is an extension of PASCAL VOC containing 400+ classes (usually 59 most frequently). \\
     COCO~\cite{coco_dataset} & 118k / 5k & SS / IS / PS &  mIoU / mAP / PQ & MS COCO dataset is a large-scale dataset with 80 thing categories and 91 stuff categories.\\
    ADE20k~\cite{ADE20K} & 20,210 / 2,000 & SS / IS / PS & mIoU / mAP / PQ & ADE20k dataset is a large-scale dataset exhaustively annotated with pixel-level objects and object part labels. \\
    Cityscapes~\cite{cordts2016cityscapes} & 2,975 / 500 & SS / IS / PS & mIoU / mAP / PQ & Cityscapes dataset focuses on semantic understanding of urban street scenes, captured in 50 cities.\\
    Mapillary~\cite{neuhold2017mapillary} & 18k / 2k & SS / PS & mIoU / PQ & Mapillary dataset is a large-scale dataset with accurate high-resolution annotations. \\
    RefCOCO~\cite{RefCOCO} & 42k / 4k & RIS & mIoU & A large-scale dataset for classic reference segmentation based on the COCO. \\
    gRefCOCO~\cite{GRES} & 79k / 8k & RIS & mIoU & A large-scale dataset for generalized referring segmentation based on the COCO. \\
    \hline
    VSPW~\cite{miao2021vspw} & 2,906 / 343 & VSS & mIoU & VPSW is a large-scale high-resolution dataset with long videos focusing on VSS.\\
    Youtube-VIS-2019~\cite{vis_dataset} & 2,238 / 302 & VIS & AP & Extending from Youtube-VOS, Youtube-VIS is with exhaustive instance labels.\\
    OVIS~\cite{OVIS} & 607 / 140 & VIS & AP & A large-scale occluded video instance segmentation benchmark.\\
    VIP-Seg~\cite{miao2022large} & 2,806 / 343 & VPS & VPQ \& STQ & Extending from VSPW, VIP-Seg adds extra instance labels for VPS task.\\
    Cityscape-VPS~\cite{kim2020vps} & 2,400 / 300 & VPS & VPQ & Cityscapes-VPS dataset extracts from the val split of Cityscapes dataset, adding temporal annotations.\\
    KITTI-STEP~\cite{STEP} & 5,027 / 2,981 & VPS & STQ & KITTI-STEP focuses on the long videos in the urban scenes. \\
    DAVIS-2017~\cite{davis2017}& 4,219 / 2,023  & VOS & J / F / J\&F& DAVIS focuses on video object segmentation. \\
    Youtube-VOS~\cite{vos2018}&  3,471 / 474   & VOS & J / F / J\&F& A large-scale video object segmentation benchmark. \\
    MOSE~\cite{MOSE}& 1,507 / 311 & VOS & J / F / J\&F& Tracking and segmenting objects in complex environments.\\
    MeViS~\cite{MeViS}& 1,712 / 140 & RVOS & J / F / J\&F& Tracking and segmenting target objects referred by motion expressions.\\
    \bottomrule[0.1em]
\end{tabular}
\end{adjustbox}
\label{tab:dataset_summary}
\end{table*}

\noindent$\bullet$
\textbf{Commonly Used Datasets.} For image segmentation, the most commonly used datasets are COCO~\cite{coco_dataset}, ADE20k~\cite{ADE20K} and Cityscapes~\cite{cordts2016cityscapes}. For video segmentation, the most used datasets are VSPW~\cite{miao2021vspw} and Youtube-VIS~\cite{vis_dataset}. We will compare several dataset results in Sec.~\ref{sec:benchmark}. More datasets are listed in the Tab.~\ref{tab:dataset_summary}.

\noindent$\bullet$
\textbf{Common Metric.} For SS and VSS, the commonly used metric is mean intersection over union (mIoU), which calculates the pixel-wised Union of Interest between output image and video masks and ground truth masks.
For IS, the metric is mask mean average precision (mAP), which is extended from the object detection via replacing box IoU with mask IoU. 
For VIS, the metric is 3D mAP, which extends mask mAP in a spatial-temporal manner. 
For PS, the metric is the panoptic quality (PQ), which unifies both thing and stuff prediction by setting a fixed threshold 0.5. 
For VPS, the commonly used metrics are video panoptic quality (VPQ) and segmentation tracking quality (STQ). The former extends PQ into temporal window calculation, while the latter decouples the segmentation and tracking in a per-pixel-wised manner. 
Note that there are other metrics, including pixel accuracy and temporal consistency.
For simplicity, we only report the primary metrics used in the literature. We present the detailed formulation of these metrics in the supplementary material.

\subsection{Segmentation Approaches Before Transformer}
\label{sec:segmentation_before_transformer}

\noindent$\bullet$
\textbf{Semantic Segmentation.} Prior to the emergence of {ViT} and {DETR}. SS was typically approached as a dense pixel classification problem, as initially proposed by FCN. Then, the following works are all based on the FCN framework. These methods can be divided into the following aspects, including better encoder-decoder frameworks~\cite{yu2018learning,ding2020semantic}, larger kernels~\cite{peng2017large,SVCNet}, multiscale pooling~\cite{zhao2017pyramid,deeplabv3}, multiscale feature fusion~\cite{ding2018context,shuai2018toward,li2020gated,li2021global}, non-local modeling~\cite{wang2018nonlocal,ocrnet,zhangli_dgcn}, efficient modeling~\cite{sfnet,Li2022SFNetFA,BiSeNet}, and better boundary delineation~\cite{kirillov2020pointrend,BoundaryAware,li2020improving,ebl_he_iccv}. After the transformer was proposed, with the goal of global context modeling, several works design variants of self-attention operators to replace the CNN prediction heads~\cite{DAnet,ocrnet}.

\noindent$\bullet$
\textbf{Instance Segmentation.} IS aims to detect and segment each object, which goes beyond object detection. 
Most IS approaches focus on how to represent instance masks beyond object detection, which can be divided into two categories: top-down approaches~\cite{maskrcnn,tian2020conditional} and bottom-up approaches~\cite{neven2019instanceSeg,de2017semanticInstanceLoss}. 
The former extends the object detector with an extra mask head. The designs of mask heads are various, including FCN heads~\cite{maskrcnn,htc}, diverse mask encodings~\cite{zhang2020MEInst}, and dynamic kernels~\cite{tian2020conditional,bolya2019yolact}. 
The latter performs instance clustering from semantic segmentation maps to form instance masks.
The performance of top-down approaches is closely related to the choice of detector ~\cite{qiao2021detectors}, while bottom-up approaches depend on both semantic segmentation results and clustering methods~\cite{cheng2020panoptic}. 
Besides, there are also several approaches~\cite{chen2019tensormask,wang2020solov2} using gird representation to learn instance masks directly. 
The ideas using kernels and different mask encodings are also extended into several transformer-based approaches, which will be detailed in Sec.~\ref{sec:method_survey}.

\noindent$\bullet$
\textbf{Panoptic Segmentation.} Previous works for PS mainly focus on how to fuse the results of both SS and IS, which treats PS as two independent tasks. 
Based on IS subtask, the previous works can also be divided into two categories: top-down approaches~\cite{xiong2019upsnet,li2020panopticFCN} and bottom-up approaches~\cite{axialDeeplab,cheng2020panoptic}, according to the way to generate instance masks. 
Several works use a shared backbone with multitask heads to jointly learn IS and SS, focusing on mutual task association. Meanwhile, several bottom-up approaches~\cite{axialDeeplab,cheng2020panoptic} use the sequential pipeline by performing instance clustering from semantic segmentation results and then fusing both. In summary, most PS methods include complex pipelines and are highly engineered.

\noindent$\bullet$
\textbf{Video Segmentation.} The research for VSS mainly focuses on better spatial-temporal fusion~\cite{gadde2017semantic} or acceleration using extra cues~\cite{shelhamer2016clockwork,DFF} in the video. 
VIS requires segmenting and tracking each instance. Most VIS approaches~\cite{mask_pro_vis,fu2021compfeat,kim2020vps,li2022improving} focus on learning instance-wised spatial, temporal relation, and feature fusion. Several works learn the 3D temporal embeddings. Like PS, VPS~\cite{kim2020vps} can also be top-down~\cite{kim2020vps} and bottom-up approaches~\cite{vip_deeplab}. 
The top-down approaches learn to link the temporal features and then perform instance association online. In contrast, the bottom-up approaches predict the center map of the near frame and perform instance association in a separate stage. 
Most of these approaches are highly engineering. For example, MaskPro~\cite{mask_pro_vis} adopts state-of-the-art IS segmentation models~\cite{htc}, deformable CNN~\cite{deformablev2}, and offline mask propagation in one system. There are also several video segmentation tasks, including video object segmentation (VOS)~\cite{VOS_data,MOSE}, referring video segmentation~\cite{MeViS}, multi-Object tracking, and segmentation (MOTS)~\cite{voigtlaender2019mots}.

\noindent$\bullet$
\textbf{Point Cloud Segmentation.} This task aims to group point clouds into semantic or instance categories, similar to image and video segmentation. 
Depending on the input scene, it is typically categorized as either indoor or outdoor scenes. 
Indoor scene segmentation mainly includes point cloud semantic segmentation (PSS) and point cloud instance segmentation (PIS). 
PSS is commonly achieved using the Point-Net~\cite{qi2017pointnet,qi2017pointnet++}, while PIS can be achieved through two approaches: top-down approaches~\cite{yang2019learning,yi2019gspn} and bottom-up approaches~\cite{wang2018sgpn,jiang2020pointgroup}. 
The former extracts 3D bounding boxes and uses a mask learning branch to predict masks, while the latter predicts semantic labels and utilizes point embedding to group points into different instances. 
For outdoor scenes, point cloud segmentation can be divided into point-based~\cite{qi2017pointnet,mao2019interpolated} and voxel-based~\cite{hu2020randla,cheng20212} approaches. 
Point-based methods focus on processing individual points, while voxel-based methods divide the point cloud into 3D grids and apply 3D convolution. 
Like panoptic segmentation, most 3D panoptic segmentation methods~\cite{zhou2021panoptic,xu2022sparse,Hong_2021_CVPR,aygun20214d,Zhu_2021_CVPR} first predict semantic segmentation results, separate instances based on these predictions and fuse the two results to obtain the final results.

\subsection{Transformer Basics}
\label{sec:transformer_basic}

\noindent$\bullet$
\textbf{Vanilla Transformer}~\cite{vaswani2017attention} is a seminal model in the transformer-based research field. It is an encoder-decoder structure that takes tokenized inputs and consists of stacked transformer blocks. 
Each block has two sub-layers: a multi-head self-attention (MHSA) layer and a position-wise fully-connected feed-forward network (FFN). 
The MHSA layer allows the model to attend to different parts of the input sequence while the FFN processes the output of the MHSA layer. 
Both sub-layers use residual connections and layer normalization for better optimization.

In the vanilla transformer, the encoder and decoder both use the same architecture. 
However, the decoder is modified to include a mask that prevents it from attending to future tokens during training. 
Additionally, the decoder uses sine and cosine functions to produce positional embeddings, which allow the model to understand the order of the input sequence. 
Subsequent models such as BERT and GPT-2 have built upon its architecture and achieved state-of-the-art results on a wide range of natural language processing tasks.

\noindent$\bullet$
\textbf{Self-Attention.} The core operator of the vanilla transformer is the self-attention (SA) operation. Suppose the input data is a set of tokens $X =[ x_{1}, x_{2}, ..., x_{N}] \in \mathbb{R}^{N \times c}$. $N$ is the token number and $c$ is token dimension.
The positional encoding $P$ may be added into $I = X + P$. The input embedding $I$ goes through three linear projection layers ($W^{q} \in \mathbb{R}^{c \times d}, W^{k} \in \mathbb{R}^{c \times d}, W^{v} \in \mathbb{R}^{c \times d}$) to generate Query (Q), Key (K), and Value (V):
\begin{equation}
    Q = IW^{q}, K=IW^{k}, V=IW^{v},
\end{equation}
where $d$ is the hidden dimension. The Query and Key are usually used to generate the attention map in SA. Then the SA is performed as follows:
\begin{equation}
\label{equ:self_atten}
    O = \mathrm{SA}(Q,K,V) = \mathrm{Softmax}(QK^\intercal)V.
\end{equation}
According to Equ.~\ref{equ:self_atten}, given an input $X$, self-attention allows each token $x_{i}$ to attend to all the other tokens.
Thus, it has the ability of global perception compared with local CNN operator.
Motivated by this, several works~\cite{wang2018nonlocal,chen2018a2net} treat it as a fully-connected graph or a non-local module for visual recognition task.

\noindent$\bullet$
\textbf{Multi-Head Self-Attention.} In practice, multi-head self-attention (MHSA) is more commonly used. 
The idea of MHSA is to stack multiple SA sub-layer in parallel, and the concatenated outputs are fused by a projection matrix $W^{fuse} \in \mathbb{R}^{d \times c}$:
\begin{equation}
    O = \mathrm{MHSA}(Q,K,V) = \mathrm{concat}([\mathrm{SA_{i},..SA_{H}}])W^{fuse},
\end{equation}
where $\mathrm{SA_{i}}= \mathrm{SA}(Q_{i}, K_{i}, V_{i})$ and $H$ is the number of the head. 
Different heads have individual parameters. Thus, MHSA can be viewed as an ensemble of SA.

\noindent$\bullet$
\textbf{Feed-Forward Network.} The goal of feed-forward network (FFN) is to enhance the non-linearity of attention layer outputs. 
It is also called multi-layer perceptron (MLP) since it consists of two successive linear layers with non-linear activation layers. 
\section{Methods: A Survey}
\label{sec:method_survey}

\begin{figure}[!t]
	\centering
	\includegraphics[width=1.0\linewidth]{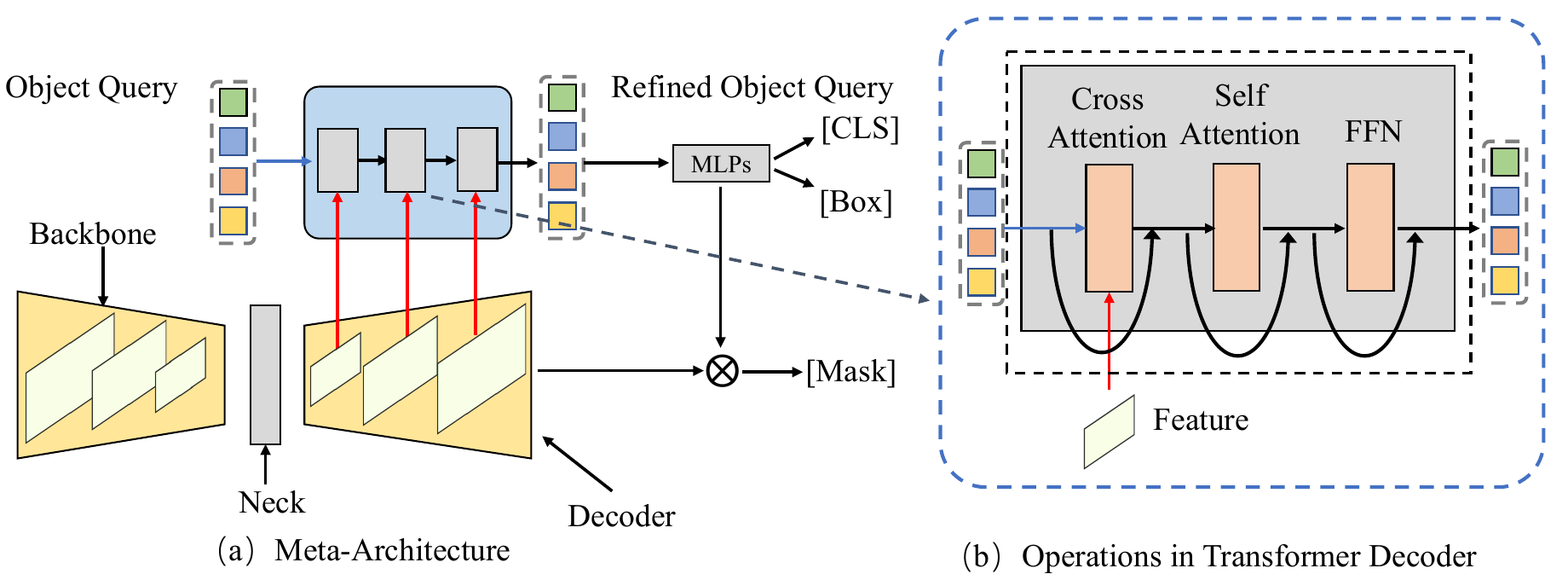}
	\caption{Illustration of (a) meta-architecture and (b) common operations in the decoder.}
    \label{fig:meta_architec}
\end{figure}

In this section, based on DETR-like meta-architecture, we review the key techniques of transformer-based segmentation. 
As shown in Fig.~\ref{fig:meta_architec}, the meta-architecture contains a feature extractor, object query, and a transformer decoder. 
Then, according to the meta-architecture, we survey existing methods by considering the modification or improvements to each component of the meta-architecture in Sec.~\ref{sec:method_strong_representation}, Sec.~\ref{sec:method_interaction_design} and Sec.~\ref{sec:extra_cue_query_learning}. 
Finally, based on such meta-architecture, we present several detailed applications in Sec.~\ref{sec:query_association} and Sec.~\ref{sec:conditional_query_generation}.

\subsection{Meta-Architecture}
\label{sec:method_meta} 

\noindent$\bullet$
\textbf{Backbone.} Before ViTs, CNNs were the standard approach for feature extraction in computer vision tasks. 
To ensure a fair comparison, many research works~\cite{maskrcnn,ren2015faster,detr} used the same CNN models, such as ResNet50~\cite{resnet}. 
Some researchers~\cite{wang2018nonlocal,axialDeeplab} also explored the combination of CNNs with self-attention layers to model long-range dependencies. 
ViT, on the other hand, utilizes a standard transformer encoder for feature extraction. 
It has a specific input pipeline for images, where the input image is split into fixed-size patches, such as 16 $\times$ 16 patches. 
These patches are then processed through a linear embedding layer. Then, the positional embeddings are added to each patch. 
Afterward, a standard transformer encoder encodes all patches. 
It contains multiple multi-head self-attention and feed-forward layers. 
For instance, given an image $I \in \mathbb{R}^{H \times W \times 3}$, ViT first reshapes it into a sequence of flattened 2D patches: $I_{p} \in \mathbb{R}^{N \times P^{2} \times 3}$, where $N$ is the number of patches and $P$ is the patch size. 
With patch embedding operations, the final input is $I_{in} \in \mathbb{R}^{N \times P^{2} \times C}$, where $C$ is the embedding channel. 
To perform classification, an extra learnable embedding ``classification token" ($\mathrm{CLS}$) is added to the sequence of embedded patches. 
After the standard transformer for all patches, $I_{out} \in \mathbb{R}^{N \times P^{2} \times C}$ is obtained. 
For segmentation tasks, ViT is used as a feature extractor, meaning that $I_{out}$ is resized back to a dense map $F \in \mathbb{R}^{H\times W \times C}$.

\noindent$\bullet$
\textbf{Neck.} Feature pyramid network (FPN) has been shown effective in object detection and instance segmentation~\cite{fpn,focal_loss,tian2021fcos} for scale variation modeling. 
FPN maps the features from different stages into the same channel dimension $C$ for the decoder.
Several works~\cite{nasfpn,qiao2021detectors} design stronger FPNs via cross-scale modeling using dilation or deformable convolution. 
For example, Deformable DETR~\cite{zhu2020deformabledetr} proposes a deformable FPN to model cross-scale fusion using deformable attention. 
Lite-DETR~\cite{li2023litedetr} further refines the deformable cross-scale attention design by efficiently sampling high-level features and low-level features in an interleaved manner.
The output features are used for decoding the boxes and masks. The role of FPN is the same as previous detection-based or FCN-based segmentation methods. The FPN generates multi-scale features to handle and balance both small and large objects in the scene. For the transformer-based method, FPN architecture is often used to refine object queries from different scales, which can lead to stronger results than single-scale refinement.

\noindent$\bullet$
\textbf{Object Query.} Object query is first introduced in DETR~\cite{detr}. It plays as the dynamic anchors that are used in detectors~\cite{maskrcnn,ren2015faster}. 
In practice, it is a learnable embedding $Q_{obj} \in \mathbb{R}^{N_{ins} \times d}$. 
$N_{ins}$ represents the maximum instance number. 
The query dimension $d$ is usually the same as feature channel $c$. 
Object query is refined by the cross-attention layers. 
Each object query represents one instance of the image. During the training, each ground truth is assigned with one corresponding query for learning.
During the inference, the queries with high scores are selected as output.
Thus, object query simplifies the design of detection and segmentation models by eliminating the need for hand-crafted components such as non-maximum suppression (NMS).
The flexible design of object query has led to many research works exploring its usage in different contexts, which will be discussed in more detail in Sec.~\ref{sec:method_categorization}.

\noindent$\bullet$
\textbf{Transformer Decoder.} Transformer decoder is a crucial architecture component in transformer-based segmentation and detection models. 
Its main operation is cross-attention, which takes in the object query $Q_{obj}$ and the image/video feature $F$. 
It outputs a refined object query, denoted as $Q_{out}$. 
The cross-attention operation is derived from the vanilla transformer architecture, where $Q_{obj}$ serves as the query, and $F$ is used as the key and value in the self-attention mechanism. 
After obtaining the refined object query $Q_{out}$, it is passed through a prediction FFN, which typically consists of a 3-layer perceptron with a ReLU activation layer and a linear projection layer. 
The FFN outputs the final prediction, which depends on the specific task. 
For example, for classification, the refined query is mapped directly to class prediction via a linear layer. 
For detection, the FFN predicts the normalized center coordinates, height, and width of the object bounding box. 
For segmentation, the output embedding is used to perform dot product with feature $F$, which results in the binary mask logits. 
The transformer decoder iteratively repeats cross-attention and FFN operations to refine the object query and obtain the final prediction. 
The intermediate predictions are used for auxiliary losses during training and discarded during inference. 
The outputs from the last stage of the decoder are taken as the final detection or segmentation results. We show the detailed process in Fig.~\ref{fig:meta_architec} (b).

\noindent$\bullet$
\textbf{Mask Prediction Representation.} Transformer-based segmentation approaches adopt two formats to represent the mask prediction: pixel-wise prediction as FCNs and per-mask-wise prediction as DETR. 
The former is used in semantic-aware segmentation tasks, including SS, VSS, VOS, and \etc. 
The latter is used in instance-aware segmentation tasks, including IS, VIS, and VPS, where each query represents each instance.

\noindent$\bullet$
\textbf{Bipartite Matching and Loss Function.} Object query is usually combined with bipartite matching~\cite{kuhn1955hungarian} during training, uniquely assigning predictions with ground truth. 
This means each object query builds the one-to-one matching during training. 
Such matching is based on the {matching cost} between ground truth and predictions. The {matching cost} is defined as the distance between prediction and ground truth, including labels, boxes, and masks. 
By minimizing the cost with the Hungarian algorithm~\cite{kuhn1955hungarian}, each object query is assigned by its corresponding ground truth. 
For object detection, each object query is trained with classification and box regression loss~\cite{ren2015faster}. 
For instance-aware segmentation, each object query is trained via both mask classification loss and segmentation loss. 
The output masks are obtained via the inner product between object query and decoder features. 
The segmentation loss usually contains binary cross-entropy loss and dice loss~\cite{dice_loss}.

\noindent$\bullet$
\textbf{Discussion on Scope of Meta-Architecture.} We admit our meta-architecture may \textbf{not} cover all transformer-based segmentation methods. 
In semantic segmentation, methods such as Segformer~\cite{xie2021segformer} and SETR~\cite{SETR} employ a fully connected layer and predict each pixel as previous FCN-based methods~\cite{long2015fully,deeplabv3,deeplabv3plus}. 
These methods concentrate on enhanced feature representation.
We argue that this represents a basic form of our meta-architecture, wherein each query corresponds to a class category. 
The cascaded cross-attention layers are omitted, and bipartite matching is removed. Thus, the object query plays the same role as a fully connected layer.
In addition, meta-architecture represents the latest design philosophy.
Nearly all recent state-of-the-art methods~\cite{OMGSeg,athar2023tarvis,yuan2021polyphonicformer,wu2023uniref++} adopt this meta-architecture.
In particular, different methods may add more components to adapt their tasks and requirements.
Thus, we review recent works by modifying each component based on this meta-architecture.

\begin{table*}[t]
\centering
\small
\caption{Transformer-Based Segmentation Method Categorization. We select the representative works for reference.}
\begin{adjustbox}{width=0.90\textwidth}
\begin{tabular}{l c l c} 
  \toprule[0.15em]
    Method Categorization &  Tasks &  Reference \\
    \toprule[0.15em]
    Representation Learning (Sec.~\ref{sec:method_strong_representation})  &  \\
      \quad $\bullet$Better ViTs Design   & SS / IS  & ~\cite{VIT,DeiT,fan2021mvitv1,li2022mvitv2,lee2022mpvit,ali2021xcit,Wang_2021_ICCV_PVT,chen2021crossvit, zhang2020feature,Xu_2021_ICCV} \\
      \quad $\bullet$Hybrid CNNs / transformers / MLPs & SS / IS  &~\cite{liu2021swin,xie2021segformer,guo2021cmt,chu2021Twins,wu2021cvt,xu2021vitae,liu2022convnet,han2021connection,guo2022segnext,yu2022metaformer,dai2022demystify}  \\
      \quad $\bullet$Self-Supervised Learning   & SS / IS &~\cite{chen2020simple,he2020momentum,chen2021mocov3,bao2021beit,MaskedAutoencoders2021,wei2022maskedfeat,gandelsman2022test,hu2022exploring,gao2022convmae,radford2021learning_clip,li2022scaling}  \\
      \hline
    Cross-Attention Design in Decoder (Sec.~\ref{sec:method_interaction_design})  &  \\
       \quad $\bullet$Improved Cross-Attention Design  & SS / IS / PS &~\cite{zhu2020deformabledetr,peize2020sparse,QueryInst,hu2021ISTR,dong2021solq,he2021boundarysqueeze,zhang2021knet,cheng2021maskformer,li2021panoptic}   \\
       \quad $\bullet$Spatial-Temporal Cross-Attention Design  &  VSS / VIS / VPS
 &~\cite{VIS_TR,zhou2022transvod,yang2022tevit,cheng2021mask2formervis,IFC_21,seqformer,li2022videoknet,kim2022tubeformer}   \\
       \hline
    Optimizing Object Query (Sec.~\ref{sec:extra_cue_query_learning})  &  \\
       \quad $\bullet$Adding Position Information into Query &  IS / PS &~\cite{meng2021conditional,chen2022conditional,wang2022anchor,liu2022dabdetr}  \\
        \quad $\bullet$Adding Extra Supervision into Query. & IS / PS &~\cite{li2022dn,zhang2022dino,li2022maskdino,Instance_Unique_Querying,jia2022detrs,chen2022group,zong2022detrs}   \\
    \hline
    Using Query For Association (Sec.~\ref{sec:query_association})\\
        \quad $\bullet$Query for Instance Association  & VIS / VPS  &~\cite{meinhardt2021trackformer,transtrack,zeng2021motr,huang2022minvis,li2022videoknet,IDOL}     \\
         \quad $\bullet$Query for Linking Multi-Tasks  & VPS / DVPS / PS / PPS / IS &~\cite{panopticpartformer,yuan2021polyphonicformer,gao2022panopticdepth,xu2022fashionformer,xu2022multi,invpt2022}     \\
    \hline
    Conditional Query Generation (Sec.~\ref{sec:conditional_query_generation}) \\
       \quad $\bullet$Conditional Query Fusion on Language Features &  RIS / RVOS  &~\cite{VLT_iccv2021,MeViS,LAVT_22cvpr,ReSTR_2022_CVPR,CRIS_2022_CVPR,botach2022end,ding2022language,GRES,wu2022towards_robust_ris,wu2022language}   \\ 
       \quad $\bullet$Conditional Query Fusion on Cross Image Features &  SS/ VOS / SS / Few Shot SS &~\cite{zhang2021few,yang2021associating,park2022matteformer,shi2022transformer,lin2022structtoken,yu2022batman,jiao2022mask}   \\
    \bottomrule[0.1em]
    
\end{tabular}
\end{adjustbox}
\label{tab:method_categorization}
\end{table*}

\subsection{Method Categorization}
\label{sec:method_categorization}

In this section, we review five aspects of transformer-based segmentation methods. 
Rather than classifying the literature by the task settings, our goal is to extract the {essential and common techniques} used in the literature. We summarize the methods, techniques, related tasks, and corresponding references in Tab.~\ref{tab:method_categorization}. 
Most approaches are based on the meta-architecture described in Sec.~\ref{sec:method_meta}. 
We list the comparison of representative works in Tab.~\ref{tab:method_classification}.

\subsubsection{Strong Representations}
\label{sec:method_strong_representation}
Learning a strong feature representation always leads to better segmentation results. 
Taking the SS task as an example, SETR~\cite{SETR} is the first to replace CNN backbone with the ViT backbone. 
It achieves state-of-the-art results on the ADE20k dataset without bells and whistles. 
After ViTs, researchers start to design better vision transformers. 
We categorize the related works into three aspects: better vision transformer design, hybrid CNNs/transformers/MLPs, and self-supervised learning.

\noindent$\bullet$
\textbf{Better ViTs Design.} Rather than introducing local bias, these works follow the original ViTs design and process feature using the original MHSA for token mixing. 
DeiT~\cite{DeiT} proposes knowledge distillation and provides strong data augmentation to train ViT efficiently. 
Starting from DeiT, nearly all ViTs adopt the stronger training procedure.
MViT-V1~\cite{fan2021mvitv1} introduces the multiscale feature representation and pooling strategies to reduce the computation cost in MHSA. 
MViT-V2~\cite{li2022mvitv2} further incorporates decomposed relative positional embeddings and residual pooling design in MViT-V1, which leads to better representation. 
Motivated by MViT, from the architecture level, MPViT~\cite{lee2022mpvit} introduces multiscale patch embedding and multi-path structure to explore tokens of different scales jointly.
Meanwhile, from the operator level, XCiT~\cite{ali2021xcit} operates across feature channels rather than token inputs and proposes cross-covariance attention, which has linear complexity in the number of tokens. This design makes it easy to adapt to segmentation tasks, which always have high-resolution inputs. 
Pyramid ViT~\cite{Wang_2021_ICCV_PVT} is the first work to build multiscale features for detection and segmentation tasks. 
There are also several works~\cite{chen2021crossvit,zhang2020feature,Xu_2021_ICCV} exploring cross-scale modeling via MHSA, which exchange long-range information on different feature pyramids.

\noindent$\bullet$
\textbf{Hybrid CNNs/Transformers/MLPs.} Rather than modifying the ViTs, many works focus on introducing local bias into ViT or using CNNs with large kernels directly. 
To build a multi-stage pipeline, Swin~\cite{liu2021swin,liu2022swin} adopts shift-window attention in a CNN style. They also scale up the models to large sizes and achieve significant improvements on many vision tasks. 
From an efficient perspective, Segformer~\cite{xie2021segformer} designs a light-weight transformer encoder. It contains a sequence reduction during MHSA and a light-weight MLP decoder.
Segformer achieves better speed and accuracy trade-off for SS. 
Meanwhile, several works~\cite{guo2021cmt,chu2021Twins,wu2021cvt,xu2021vitae} directly add CNN layers to a transformer to explore the local context. 
Several works~\cite{chen2021cyclemlp,tolstikhin2021mlp} explore the pure MLPs design to replace the transformer.
With specific designs such as shifting and fusion~\cite{chen2021cyclemlp}, MLP models can also achieve comparable results with ViTs.
Later, several works~\cite{liu2022convnet,han2021connection} point out that CNNs can achieve stronger results than ViTs if using the same data augmentation pipeline. 
In particular, DWNet~\cite{han2021connection} re-visits the training pipeline of ViTs and proposes dynamic depth-wise convolution. 
Then, ConvNeXt~\cite{liu2022convnet} uses the larger kernel depth-wise convolution and a stronger data training pipeline. It achieves stronger results than Swin~\cite{liu2021swin}. 
Motivated by ConvNeXt, SegNext~\cite{guo2022segnext} designs a CNN-like backbone with linear self-attention and performs strongly on multiple SS benchmarks. 
Meanwhile, Meta-Former~\cite{yu2022metaformer} shows that the meta-architecture of ViT is the key to achieving stronger results. 
Such meta-architecture contains a token mixer, a MLP, and residual connections. 
The token mixer is a simple MHSA layer. Meta-Former shows that the token mixer is not as important as meta-architecture. Using simple pooling as a token mixer can achieve stronger results. 
Following the Meta-Former, recent work~\cite{dai2022demystify} re-benchmarks several previous works using a unified architecture to eliminate unfair engineering techniques. 
However, under stronger settings, the authors find the spatial token mixer design still matters. 
Meanwhile, several works~\cite{chen2022cyclemlp,guo2021hire} explore the MLP-like architecture for dense prediction. 

\noindent$\bullet$
\textbf{Self-Supervised Learning (SSL).} SSL has achieved huge progress in recent years~\cite{chen2020simple,he2020momentum,chen2023context}. 
Compared with supervised learning, SSL exploits unlabeled data via specially designed pseudo tasks and can be easily scaled up.
MoCo-v3~\cite{chen2021mocov3} is the first study that trains ViTs in SSL. 
It freezes the patch projection layer to stabilize the training process. 
Motivated by BERT, BEiT~\cite{bao2021beit} proposes the BERT-like per-training (Mask Image Modeling, MIM) of vision transformers. 
After BEiT, MAE~\cite{MaskedAutoencoders2021} shows that ViTs can be trained with the simplest MIM style. 
By masking a portion of input tokens and reconstructing the RGB images, MAE achieves better results than supervised training.
As a concurrent work, MaskFeat~\cite{wei2022maskedfeat} mainly studies reconstructing targets of the MIM framework, such as the histogram of oriented gradient (HOG) features. 
The following works focus on improving the MIM framework~\cite{gandelsman2022test,hu2022exploring} or replacing the backbone of ViTs with CNN architecture~\cite{gao2022convmae,tian2023designing}. 
DINO series~\cite{caron2021emergingDINO} find the self-supervised learned feature itself has grouping effects, which is always used in unsupervised learning contexts. (Sec.~\ref{sec:label_efficient})
Recently, several works~\cite{radford2021learning_clip,jia2021scaling_align} on VLM also adopt SSL by utilizing easily obtained text-image pairs. 
Recent work~\cite{li2022scaling} demonstrates the effectiveness of VLM in downstream tasks, including IS and SS. 
Moreover, several recent works~\cite{li2022uniperceiver_v2} adopt multi-modal SSL pre-training and design a unified model for many vision tasks.
For video representation learning, most current works~\cite{videomae,MaskedAutoencodersSpatiotemporal2022,liu2022video} verify such representation learning on action or motion learning, such as action recognition. 
Several works~\cite{wu2022language,ding2022vlt} adopt a video backbone for video segmentation. 
However, for video segmentation, from the method design perspective, most works focus on matching and association of entities or pixels, which is discussed in Sec.~\ref{sec:method_interaction_design} and Sec.~\ref{sec:query_association}.

\subsubsection{Cross-Attention Design in Decoder}
\label{sec:method_interaction_design}


In this section, we review the new transformer decoder designs. 
We categorize the decoder design into two groups: one for improved cross-attention design in image segmentation and the other for spatial-temporal cross-attention design in video segmentation. 
The former focuses on designing a better decoder to refine the original decoder in the original DETR. 
The latter extends the query-based object detector and segmenter into the video domain for VOD, VIS, and VPS, focusing on modeling temporal consistency and association. 

\noindent$\bullet$
\textbf{Improved Cross-Attention Design.} Cross-attention is the core operation of meta-architecture for segmentation and detection. 
Current solutions for improved cross-attention mainly focus on designing new or enhanced cross-attention operators and improved decoder architectures. 
Following DETR, Deformable DETR~\cite{zhu2020deformabledetr} proposes deformable attention to efficiently sample point features and perform cross-attention with object query jointly. 
%
%
Meanwhile, several works bring object queries into previous RCNN frameworks.
Sparse-RCNN~\cite{peize2020sparse} uses RoI pooled features to refine the object query for object detection. 
They also propose a new dynamic convolution and self-attention to enhance object query without extra cross-attention. 
In particular, the pooled query features reweight the object query, and then self-attention is applied to the object query to obtain the global view. 
After that, several works~\cite{QueryInst,hu2021ISTR} add the extra mask heads for IS.
QueryInst~\cite{QueryInst} adds mask heads and refines mask query with dynamic convolution. Meanwhile, several works~\cite{yu2022soit,dong2021solq} extend Deformable DETR by directly applying MLP on the shared query.
Inspired by MEInst~\cite{zhang2020MEInst}, SOLQ~\cite{dong2021solq} utilizes mask encodings on object query via MLP. 
%
By applying the strong Deformable DETR detector and Swin transformer~\cite{liu2021swin} backbone, it achieves remarkable results on IS.
However, these works still need extra box supervision, which makes the system complex. 
Moreover, most RoI-based approaches for IS have low mask quality issues since the mask resolution is limited within the boxes~\cite{kirillov2020pointrend}.

To fix the issues of extra box heads, several works remove the box prediction and adopt pure mask-based approaches. 
Earlier work, OCRNet~\cite{ocrnet} characterizes a pixel by exploiting the representation of the corresponding object class that forms a category query.
Then, Segmenter~\cite{strudel2021_segmenter} adopts a strong ViT backbone with the class query to directly decode class-wise masks.
Pure mask-based approaches directly generate segmentation masks from high-resolution features and naturally have better mask quality. 
Max-Deeplab~\cite{wang2020maxDeeplab} is the first to remove the box head and design a pure-mask-based segmenter for PS. 
It also achieves stronger performance than box-based PS method~\cite{qiao2021detectors}.
It combines a CNN-transformer hybrid encoder~\cite{axialDeeplab} and a transformer decoder as an extra path. 
Max-Deeplab still needs extra auxiliary loss functions, such as semantic segmentation loss, and instance discriminative loss.
%
K-Net~\cite{zhang2021knet} uses mask pooling to group the mask features and designs a gated dynamic convolution to update the corresponding query. 
By viewing the segmentation tasks as convolution with different kernels, K-Net is the first to unify all three image segmentation tasks, including SS, IS, and PS. %
Meanwhile, MaskFormer~\cite{cheng2021maskformer} extends the original DETR by removing the box head and transferring the object query into the mask query via MLPs.
It proves simple mask classification can work well enough for all three segmentation tasks. 
Compared to MaskFormer, K-Net is good at training data efficiency.
This is because K-Net adopts mask pooling to localize object features and then update object queries accordingly. 
Motivated by this, Mask2Former~\cite{cheng2021mask2former} proposes masked cross-attention and replaces the cross-attention in MaskFormer. 
Masked cross-attention makes object query only attend to the object area, guided by the mask outputs from previous stages. 
Mask2Former also adopts a stronger Deformable FPN backbone~\cite{zhu2020deformabledetr}, stronger data augmentation~\cite{detectron2}, and multiscale mask decoding.
The above works only consider updating object queries. To handle this, CMT-Deeplab~\cite{yu2022cmt} proposes an alternating procedure for object query and decoder features. 
It jointly updates object queries and pixel features. 
After that, inspired by the k-means clustering algorithm, kMaX-DeepLab~\cite{kmax_deeplab_2022} proposes k-means cross-attention by introducing cluster-wise argmax operation in the cross-attention operation. 
Meanwhile, PanopticSegformer~\cite{li2021panoptic} proposes a decoupling query strategy and deeply supervised mask decoder to speed up the training process. 
For real-time segmentation setting, SparseInst~\cite{Cheng2022SparseInst} proposes a sparse set of instance activation maps highlighting informative regions for each foreground object. 

Besides segmentation tasks, several works speed up the convergence of DETR by introducing new decoder designs, and most approaches can be extended into IS.
Several works bring such semantic priors in the DETR decoder. SAM-DETR~\cite{zhang2022_SAMDETR} projects object queries into semantic space and searches salient points with the most discriminative features. 
SMAC~\cite{gao2021fast} conducts location-aware co-attention by sampling features of high near estimated bounding box locations. 
Several works adopt dynamic feature re-weights. 
From the multiscale feature perspective, AdaMixer~\cite{gao2022adamixer} samples feature over space and scales using the estimated offsets. It dynamically decodes sampled features with an MLP, which builds a fast-converging query-based detector. 
ACT-DETR~\cite{zheng2020end} clusters the query features adaptively using a locality-sensitive hashing and replaces the query-key interaction with the prototype-key interaction to reduce cross-attention cost. 
From the feature re-weighting view, Dynamic-DETR~\cite{dai2021dynamic} introduces dynamic attention to both the encoder and decoder parts of DETR using RoI-wise dynamic convolution. 
Motivated by the sparsity of the decoder feature, Sparse-DETR~\cite{roh2022sparse} selectively updates the referenced tokens from the decoder and proposes an auxiliary detection loss on the selected tokens in the encoder to keep the sparsity. 
In summary, dynamically assigning features into query learning speeds up the convergence of DETR.

\begin{table*}[ht]
    \centering
    \small
    \caption{Representative works summarization and comparison in Sec.~\ref{sec:method_survey}.}
    \setlength{\tabcolsep}{3.0pt}
    \scalebox{0.75}{
\begin{tabular}{p{0.14\textwidth}p{0.10\textwidth}p{0.20\textwidth}p{0.20\textwidth}p{0.60\textwidth}}
    \toprule
    \belowrulesepcolor{gray!30!}
\rowcolor{gray!30!} Method & Task & Input/Output & Transformer Architecture  & \ \ ~~~~~~~~~~~~~~~~~~~~~~~~~~~~~~~~~~~~Highlight \\ \aboverulesepcolor{gray!30!} \midrule
\belowrulesepcolor{gray!15!}
\rowcolor{gray!15!}\multicolumn{5}{c}{\textbf{Strong Representations (Sec.~\ref{sec:method_strong_representation})}} \\ \aboverulesepcolor{gray!15!} \midrule
SETR~\cite{SETR} & SS & Image/Semantic Masks &  Pure transformer + CNN decoder & the first vision transformer to replace CNN backbone in SS. \\ 
\rowcolor{gray!10!} Segformer~\cite{xie2021segformer} & SS & Image/Semantic Masks &  Pure transformer + MLP head  & a light-weight transformer backbone with simple MLP prediction head. \\ 
MAE~\cite{MaskedAutoencoders2021} & SS/IS & Image/Semantic Masks &  Pure transformer + CNN decoder & a MIM pretraining framework for plain ViTs, which achieves better results than supervised training. \\ 
\rowcolor{gray!10!} SegNext~\cite{guo2022segnext} & SS & Image/Semantic Masks &  Transformer + CNN &  a large kernel CNN backbone with linear self-attention layer. \\
\midrule
\belowrulesepcolor{gray!15!}
\rowcolor{gray!15!}\multicolumn{5}{c}{\textbf{Cross-Attention Design in Decoder (Sec.~\ref{sec:method_interaction_design})}} \\ \aboverulesepcolor{gray!15!} \midrule
Deformable DETR~\cite{zhu2020deformabledetr} & OD & Image/Box & CNN + query decoder &  a new multi-scale deformable attention and a new encoder-decoder framework. \\ 
OCRNet~\cite{ocrnet} & SS & Image/Semantic Masks &  CNN + query decoder & 
introduces category queries and uses one cross-attention layer to model global context efficiently. \\
Segmenter~\cite{strudel2021_segmenter} & SS & Image/Semantic Masks &  ViT + query decoder & uses ViT backbone and category queries to directly output each class mask. \\
\rowcolor{gray!10!} Sparse-RCNN~\cite{peize2020sparse} & OD  &  Image/Box &  CNN + query decoder  & a new dynamic convolution layer and combine object query with RoI-based detector. \\  
AdaMixer~\cite{gao2022adamixer} & OD & Image/Box & CNN + query decoder &  a new multiscale query-based decoder and refine query with multiscale features.   \\ 
\rowcolor{gray!10!} Max-Deeplab~\cite{wang2020maxDeeplab} & PS & Image/Panoptic Masks & CNN + attention + query decoder &  the first pure mask supervised panoptic segmentation method and a two path framework (query and CNN features). \\ 
K-Net~\cite{zhang2021knet} & SS/IS/PS & Image/Panoptic Masks & CNN + query decoder & the first work using kernels to unify image segmentation tasks and a new mask-based dynamic kernel update module.  \\ 
\rowcolor{gray!10!} Mask2Former~\cite{cheng2021mask2former} & SS/IS/PS & Image/Panoptic Masks &  CNN + query decoder  &  design masked cross-attention and fully utilize the multiscale features in the decoder. \\ 
kMax-Deeplab~\cite{kmax_deeplab_2022} & SS/PS &  Image/Panoptic Masks &  CNN + query decoder  &  proposes a new k-mean style cross-attention by replacing softmax with argmax operation. \\ 
\rowcolor{gray!10!} VisTR~\cite{VIS_TR} & VIS & Video/Instance Masks & CNN + query decoder  &  the first end-to-end VIS method and each query represent a tracked object in a clip. \\ 
VITA~\cite{heo2022vita} & VIS & Video/Instance Masks & CNN + query decoder  & use the fixed object detector and process all frame queries with extra encoder-decoder and global queries. \\
\rowcolor{gray!10!} TubeFormer~\cite{kim2022tubeformer} & VSS/VIS/VPS & Video/Panoptic Masks &  CNN + query decoder   &  a tube-like decoder with a token exchanging mechanism within the tube, which unifies three video segmentation tasks in one framework. \\ 
Video K-Net~\cite{li2022videoknet} & VSS/VIS/VPS & Video/Panoptic Masks &  CNN + query decoder   &  
 unified online video segmentation and adopt object query for association and linking.  \\  
\midrule
\belowrulesepcolor{gray!15!}
\rowcolor{gray!15!}\multicolumn{5}{c}{\textbf{Optimizing Object Query (Sec.~\ref{sec:extra_cue_query_learning})}}\\ \aboverulesepcolor{gray!15!}\midrule
 Conditional DETR~\cite{meng2021conditional} & OD & Image/Box &  CNN + query decoder & add a spatial query to explore the extremity regions to speed up the DETR training.  \\ 
\rowcolor{gray!10!}DN-DETR~\cite{li2022dn} & OD & Image/Box & CNN + query decoder & add noisy boxes and de-noisy loss to stable query matching and improve the coverage of DETR. \\ 
Group-DETR~\cite{chen2022group} & OD & Image/Box &  CNN + query decoder  & introduce one-to-many assignment by extending more queries into groups. \\ 
\rowcolor{gray!10!}Mask-DINO~\cite{li2022maskdino} & IS/PS  & Image/Panoptic Masks & CNN + query decoder & boost instance/panoptic segmentation with object detection datasets. \\ 
\midrule\belowrulesepcolor{gray!15!}
\rowcolor{gray!15!}\multicolumn{5}{c}{\textbf{Using Query For Association (Sec.~\ref{sec:query_association})}} \\ \aboverulesepcolor{gray!15!} \midrule
 \rowcolor{gray!10!} MOTR~\cite{zeng2021motr} & MOT & Video/Box & CNN + query decoder &  design an extra tracking query for object association. \\ 
 MiniVIS~\cite{huang2022minvis} & VIS & Video/Instance Masks &  CNN/transformer + query decoder  &  
 perform video instance segmentation with image level pretraining and image object query for tracking. \\
 \rowcolor{gray!10!} Polyphonicformer \cite{yuan2021polyphonicformer} & D-VPS & Video/(Depth+Panoptic Masks) & CNN/transformer + query decoder & use object query and depth query to model instance-wise mask and depth prediction jointly. \\
 X-Decoder~\cite{zou2022xdecoder} & SS/PS &  Image /Panoptic Masks & CNN/transformer + query decoder & jointly pre-train image segmentation and language model and perform zero-shot inference on multiple segmentation tasks. \\
 \rowcolor{gray!10!}LMPM~\cite{MeViS} & RVOS & (Video+Text)/Instance Masks &  CNN/transformer + query decoder  & capture motion by associating frame-level object tokens from an off-the-shelf instance segmentation model. \\  
\midrule\belowrulesepcolor{gray!15!}
\rowcolor{gray!15!}\multicolumn{5}{c}{\textbf{Conditional Query Generation (Sec.~\ref{sec:conditional_query_generation})}} \\ \aboverulesepcolor{gray!15!} \midrule
VLT~\cite{VLT_iccv2021,ding2022vlt} & RIS & (Image+Text)/Instance Masks & CNN + transformer decoder &  design a query generation module to produce language conditional queries for transformer decoder. \\ 
\rowcolor{gray!10!} LAVT~\cite{LAVT_22cvpr} & RIS & (Image+Text)/Instance Masks & Transformer + CNN decoder &  design gated cross-attention between pyramid features and language features. \\ 
MTTR~\cite{botach2022end} & RVOS & (Video+Text)/Instance Masks &  Transformer + query decoder  & perform spatial-temporal cross-attention between language features and object query. \\ 
\rowcolor{gray!10!} X-DETR~\cite{cai2022x} &  OD & (Image+Text)/Box  &  Transformer + query decoder  &  perform directly alignment between language features and object query. \\ 
CyCTR~\cite{zhang2021few} & Few-Shot SS & (Image+Masks)/Instance Masks & Transformer + query decoder & design a cycle cross-attention between features in support images and query images.  \\
 \bottomrule
\end{tabular}
}
\label{tab:method_classification}
\vspacefigtext
\end{table*}

\noindent$\bullet$
\textbf{Spatial-Temporal Cross-Attention Design.}
\begin{figure}[!t]
	\centering
	\includegraphics[width=1.0\linewidth]{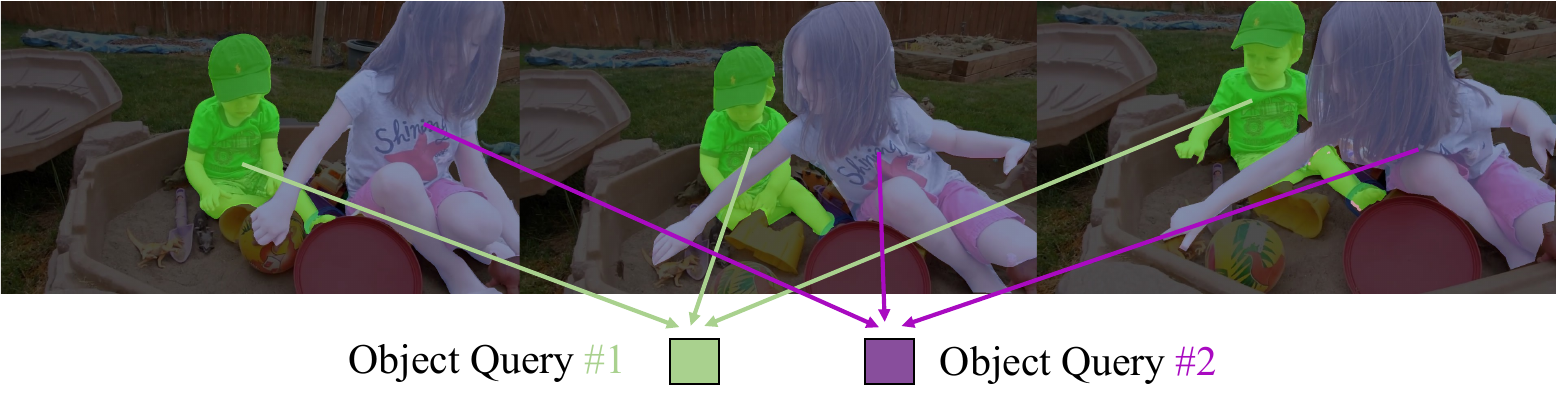}
	\caption{Illustration of object query in video segmentation.}
        \vspace{-4mm}
    \label{fig:video_query}
\end{figure}
%
After extending the object query in the video domain, each object query represents a tracked object across different frames, which is shown in Fig.~\ref{fig:video_query}.
The simplest extension is proposed by VisTR~\cite{VIS_TR} for VIS. VisTR extends the cross-attention in DETR into multiple frames by stacking all clip features into flattened spatial-temporal features. 
The spatial-temporal features also involve temporal embeddings. During inference, one object query can directly output spatial-temporal masks without extra tracking. 
Meanwhile, TransVOD~\cite{zhou2022transvod} proposes to link object query and corresponding features across the temporal domain. It splits the clips into sub-clips and performs clip-wise object detection. 
TransVOD utilizes the local temporal information and achieves better speed and accuracy trade-off. 
IFC~\cite{IFC_21} adopts message tokens to exchange temporal context among different frames. 
The message tokens are similar to learnable queries, which perform cross-attention with features in each frame and self-attention among the tokens. 
After that, TeViT~\cite{yang2022tevit} proposes a novel messenger shift mechanism for temporal fusion and a shared spatial-temporal query interaction mechanism to utilize both frame-level and instance-level temporal context information. 
Seqformer~\cite{seqformer} combines Deformable-DETR and VisTR in one framework. It also proposes to use image datasets to augment video segmentation training. 
Mask2Former-VIS~\cite{cheng2021mask2formervis} extends masked cross-attention in Mask2Former~\cite{cheng2021mask2former} into temporal masked cross-attention. 
Following VisTR, it also directly outputs spatial-temporal masks. 

In addition to VIS, several works~\cite{zhang2021knet,cheng2021mask2former,li2021panoptic} have shown that query-based methods can naturally unify different segmentation tasks. 
Following this pipeline, there are also several works~\cite{li2022videoknet,kim2022tubeformer} solving multiple video segmentation tasks in one framework. In particular, based on K-Net~\cite{zhang2021knet}, Video K-Net~\cite{li2022videoknet} proposes to unify VPS/VIS/VSS via tracking and linking kernels and works in an online manner. 
Meanwhile, TubeFormer~\cite{kim2022tubeformer} extends Max-Deeplab~\cite{wang2020maxDeeplab} into the temporal domain by obtaining the mask tubes. Cross-attention is carried out in a clip-wise manner.
During inference, the instance association is performed by mask-based matching. 
Moreover, several works~\cite{heo2022vita} propose the local temporal window to refine the global spatial-temporal cross-attention. 
For example, VITA~\cite{heo2022vita} aggregates the local temporal query on top of an off-the-shelf transformer-based image instance segmentation model~\cite{cheng2021mask2former}.
Recently, several works~\cite{shin2023video,tubelink} have explored the cross-clip association for video segmentation.
In particular, Tube-Link~\cite{tubelink} designs a universal video segmentation framework via learning cross-tube relations. It performs better than task-specific methods in VSS, VIS, and VPS.

\subsubsection{Optimizing Object Query}
\label{sec:extra_cue_query_learning}
Compared with Faster-RCNN~\cite{ren2015faster}, DETR~\cite{detr} needs a much longer schedule for convergence. 
Due to the critical role of object query, several approaches have launched studies on speeding up training schedules and improving performance. 
According to the methods for the object query, we divide the following literature into two aspects: {adding position information} and {adopting extra supervision}. 
The position information provides the cues to sample the query feature for faster training. 
The extra supervision focuses on designing specific loss functions in addition to default ones in DETR.

\noindent$\bullet$
\textbf{Adding Position Information into Query.} 
Conditional DETR~\cite{meng2021conditional} finds cross-attention in DETR relies highly on the content embeddings for localizing the four extremities. The authors introduce conditional spatial query to explore the extremity regions explicitly. 
Conditional DETR V2~\cite{chen2022conditional} introduces the box queries from the image content to improve detection results. The box queries are directly learned from image content, which is dynamic with various image inputs. The image-dependent box query helps locate the object and improve the performance. Motivated by previous anchor designs in object detectors, several works bring anchor priors in DETR.
The Efficient DETR~\cite{yao2021efficientdetr} adopts hybrid designs by including query-based and dense anchor-based predictions in one framework. 
Anchor DETR~\cite{wang2022anchor} proposes to use anchor points to replace the learnable query and also designs an efficient self-attention head for faster training.
Each object query predicts multiple objects at one position. 
DAB-DETR~\cite{liu2022dabdetr} finds the localization issues of the learnable query and proposes dynamic anchor boxes to replace the learnable query. 
Dynamic anchor boxes make the query learning more explainable and explicitly decouple the localization and content part, further improving the detection performance.

\noindent$\bullet$
\textbf{Adding Extra Supervision into Query.} DN-DETR~\cite{li2022dn} finds that the instability of bipartite graph matching causes the slow convergence of DETR and proposes a denoising loss to stabilize query learning. 
In particular, the authors feed GT bounding boxes with noises into the transformer decoder and train the model to reconstruct the original boxes. 
Motivated by DN-DETR, based on Mask2Former, MP-Former~\cite{mp_former} finds inconsistent predictions between consecutive layers. 
It further adds class embeddings of both ground truth class labels and masks to reconstruct the masks and labels.
Meanwhile, DINO~\cite{zhang2022dino} improves DN-DETR via a contrastive way of denoising training and a mixed query selection for better query initialization. 
Mask DINO~\cite{li2022maskdino} extends DINO by adding an extra query decoding head for mask prediction. 
Mask DINO~\cite{li2022maskdino} proposes a unified architecture and joint training process for both object detection and instance segmentation. 
By sharing the training data, Mask DINO can scale up and fully utilize the detection annotations to improve IS results.
Meanwhile, motivated by contrastive learning, IUQ~\cite{Instance_Unique_Querying} introduces two extra supervisions, including cross-image contrastive query loss via extra memory blocks and equivalent loss against geometric transformations. 
Both losses can be naturally adapted into query-based detectors. 
Meanwhile, there are also several works~\cite{wang2022towards,jia2022detrs,chen2022group,zong2022detrs} exploring query supervision from the target assignment perspective. 
In particular, since DETR lacks the capability of exploiting multiple positive object queries, DE-DETR~\cite{wang2022towards} first introduces one-to-many label assignment in query-based instance perception framework, to provide richer supervision for model training. Group DETR~\cite{chen2022group} proposes group-wise one-to-many assignments during training. H-DETR~\cite{jia2022detrs} adds auxiliary queries that use one-to-many matching loss during training. 
Rather than adding more queries, Co-DETR~\cite{zong2022detrs} proposes a collaborative hybrid training scheme using parallel auxiliary heads supervised by one-to-many label assignments. 
All these approaches drop the extra supervision heads during inference. These extra supervision designs can be easily extended to query-based segmentation methods~\cite{cheng2021mask2former,zhang2021knet}.

\subsubsection{Using Query For Association}
\label{sec:query_association}

Benefiting from the simplicity of query representation, several recent works have adopted it as an association tool to solve downstream tasks. There are mainly two usages: one for instance-level association and the other for task-level association. The former adopts the idea of instance discrimination, for instance-wise matching problems in video, such as joint segmentation and tracking. The latter adopts queries to link features for multitask learning.

\noindent$\bullet$
\textbf{Using Query for Instance Association.} The research in this area can be divided into two aspects: one for designing extra tracking queries and the other for using object queries directly. 
TrackFormer~\cite{meinhardt2021trackformer} is the first to treat multi-object tracking as a set prediction problem by performing joint detection and tracking-by-attention. 
TransTrack~\cite{transtrack} uses the object query from the last frame as a new track query and outputs tracking boxes from the shared decoder. 
MOTR~\cite{zeng2021motr} introduces the extra track query to model the tracked instances of the entire video. In particular, MOTR proposes a new tracklet-awared label assignment to train track queries and a temporal aggregation module to fuse temporal features. 
There are also several works~\cite{huang2022minvis,MeViS,li2022videoknet,IDOL,tubelink} adopting object query solely for tracking.  In particular, MiniVIS~\cite{huang2022minvis} directly uses object query for matching without extra tracking head modeling for VIS, where it adopts image instance segmentation training. 
Both Video K-Net~\cite{li2022videoknet} and IDOL~\cite{IDOL} learn the association embeddings directly from the object query using a temporal contrastive loss. During inference, the learned association embeddings are used to match instances across frames. 
These methods are usually verified in VIS and VPS tasks. All methods pre-train their image baseline on image datasets, including COCO and Cityscapes, and fine-tune their video architecture in the video datasets. 

\noindent$\bullet$
\textbf{Using Query for Linking Multi-Tasks.} Several works~\cite{panopticpartformer,yuan2021polyphonicformer,li2023panopticpartformer++,DsHmp} use object query to link features across different tasks to achieve mutual benefits. 
Rather than directly fusing multitask features, using object query fusion not only selects the most discriminative parts to fuse but also is more efficient than dense feature fusion. 
In particular, Panoptic-PartFormer~\cite{panopticpartformer} links part and panoptic features via different object queries into an end-to-end framework, where joint learning leads to better part segmentation results.
Several works~\cite{yuan2021polyphonicformer,gao2022panopticdepth} combine segmentation features, and depth features using the MHSA layer on corresponding depth query and segmentation query, which unify the depth prediction and panoptic segmentation prediction via shared masks. Both works find the mutual effect for both segmentation and depth prediction.
Recently, several works~\cite{xu2022multi,invpt2022} have adopted the vision transformers with multiple task-aware queries for multi-task dense prediction tasks. In particular, they treat object queries as task-specific hidden features for fusion and perform cross-task reasoning using MSHA on task queries. 
Moreover, in addition to dense prediction tasks, FashionFormer~\cite{xu2022fashionformer} unifies fashion attribute prediction and instance part segmentation in one framework. It also finds the mutual effect on instance segmentation and attribute prediction via query sharing.
Recently, X-Decoder~\cite{zou2022xdecoder} uses two different queries for segmentation and language generation tasks. 
The authors jointly pre-train two different queries using large-scale vision language datasets, where they find both queries can benefit corresponding tasks, including visual segmentation and caption generation.

\subsubsection{Conditional Query Fusion}
\label{sec:conditional_query_generation}
In addition to using object query for multitask prediction, several works adopt conditional query design for cross-modal and cross-image tasks. The query is conditional on the task inputs, and the decoder head uses such a conditional query to obtain the corresponding segmentation masks. Based on the source of different inputs, we split these works into two aspects: language features and image features.

\noindent$\bullet$
\textbf{Conditional Query Fusion From Language Feature.} Several works~\cite{VLT_iccv2021,ding2022vlt,MeViS,GRES,LAVT_22cvpr,GRES,ReSTR_2022_CVPR,D2Zero,DsHmp,botach2022end,RIE,ding2022language,wu2022towards_robust_ris} adopt conditional query fusion according to input language feature for both referring image segmentation (RIS)~\cite{GRES} and referring video object segmentation (RVOS)~\cite{MeViS} tasks. 
In particular, VLT~\cite{VLT_iccv2021,ding2022vlt} firstly adopts the vision transformer for the RIS task and proposes a query generation module to produce multiple sets of language-conditional queries, which enhances the diversified comprehensions of the language. 
Then, it adaptively selects the output features of these queries via the proposed query balance module. 
Following the same idea, LAVT~\cite{LAVT_22cvpr} designs a new gated cross-attention fusion where the image features are the query inputs of a self-attention layer in the encoder part. Compared with previous CNN approaches~\cite{ISFP,MCN}, using a vision transformer significantly improves the language-driven segmentation quality. 
With the help of CLIP's knowledge, CRIS~\cite{CRIS_2022_CVPR} proposes vision-language decoding and contrastive learning for achieving text-to-pixel alignment. 
Meanwhile, several works~\cite{wu2022multilevel_ref_video_seg,botach2022end,MeViS,wu2022language} adopt video detection transformer in Sec.~\ref{sec:method_interaction_design} for the RVOS task. MTTR~\cite{botach2022end} models the RVOS task as a sequence prediction problem and proposes both language and video features jointly. 
Recently, several works~\cite{DsHmp,MeViS} explore referring VOS under fast motion condition settings. 
Each object query in each frame combines the language features before sending it into the decoder. 
To speed up the query learning, ReferFormer~\cite{wu2022language} designs a small set of object queries conditioned on the language as the input to the transformer. 
The conditional queries are transformed into dynamic kernels to generate tracked object masks in the decoder. With the same design as VisTR, ReferFormer can segment and track object masks with given language inputs. 
In this way, each object tracklet is controlled by a given language input. 
In addition to referring segmentation tasks, MDETR~\cite{MDETR} presents an end-to-end modulated detector that detects objects in an image conditioned on a raw text query. 
In particular, they fuse the text embedding directly into visual features and jointly train the fused feature and object query. 
X-DETR~\cite{cai2022x} proposes an effective architecture for instance-wise vision-language tasks via using dot-product to align vision and language. 
In summary, these works fully utilize the interaction of language features and query features.



\noindent$\bullet$
\textbf{Condition Query Fusion From Image Feature.} Several tasks take multiple images as references and refine corresponding object masks of the main image. 
The multiple images can be support images in few shot segmentation~\cite{cao2022prototype,jiao2022mask,zhang2021few} or the same input image in matting~\cite{park2022matteformer,DIIM} and semantic 
segmentation~\cite{shi2022transformer,lin2022structtoken}. 
These works aim to model the correspondences between the main image and other images via condition query fusion. 
For SS, StructToken~\cite{lin2022structtoken} presents a new framework by doing interactions between a set of learnable structure tokens and the image features, where the image features are the spatial priors. 
In the video, BATMAN~\cite{yu2022batman} fuses optical flow features and previous frame features into mixed features and uses such features as a query to decode the current frame outputs. 
For few-shot segmentation, CyCTR~\cite{zhang2021few} aggregates pixel-wise support features into query features. In particular, CyCTR performs cross-attention between features from different images in a cycle manner, where support image features and query image features are the query inputs of the transformer jointly. 
Meanwhile, MM-Former~\cite{jiao2022mask} adopts a class-agnostic method~\cite{cheng2021mask2former} to decompose the query image into multiple segment proposals. 
Then, the support and query image features are used to select the correct masks via a transformer module. 
Then, for few-shot instance segmentation, RefTwice~\cite{han2023referencetwice} proposes an object query enhanced framework to weight query image features via object queries from support queries.
In image matting, MatteFormer~\cite{park2022matteformer} designs a new attention layer called prior-attentive window self-attention based on Swin~\cite{liu2021swin}. The prior token represents the global context feature of each trimap region, which is the query input of window self-attention. The prior token introduces spatial cues and achieves thinner matting results. 
In summary, according to the different tasks, the image features play as the decoder features in previous Sec.~\ref{sec:method_interaction_design}, which enhance the features in the main images.

\section{Specific Subfields}
\label{sec:method_downstream_beyond}

In this section, we revisit several related subfields that adopt vision transformers for segmentation tasks. The subfields include point cloud segmentation, domain-aware segmentation, label and model efficient segmentation, class agnostic segmentation, tracking, and medical segmentation.


\subsection{Segmentation}
\label{sec:point_cloud}

\noindent$\bullet$
\textbf{Semantic Level Point Cloud Segmentation.} Like image segmentation and video semantic segmentation, adopting transformers for semantic level processing mainly focuses on learning a strong representation (Sec.~\ref{sec:method_strong_representation}). 
The works~\cite{guo2021pct,point_transformer} focus on transferring the success in image/video representation learning into the point cloud. 
Early works~\cite{point_transformer} directly use modified self-attention as backbone networks and design U-Net-like architectures for segmentation. 
In particular, Point-Transformer~\cite{point_transformer} proposes vector self-attention and subtraction relation to aggregate local features progressively. The concurrent work PCT~\cite{guo2021pct} also adopts a self-attention operation and enhances input embedding with the support of farthest point sampling and nearest neighbor searching. However, the ability to model long-range context and cross-scale interaction is still limited. 
Stratified-Transformer~\cite{lai2022stratified} extends the idea of Swin Transformer~\cite{liu2021swin} into the point cloud and dived 3D inputs into cubes. It proposes a mixed key sampling method for attention input and enlarges the effective receptive field via merging different cube outputs. Meanwhile, several works also focus on better pre-training or distilling the knowledge of 2D pre-trained models. 
PointBert~\cite{yu2022pointBERT} designs the first Masked Point Modeling (MPM) task to pre-train point cloud transformers. It divides a point cloud into several local point patches as the input of a standard transformer. Moreover, it also pre-trains a point cloud Tokenizer with a discrete variational autoEncoder to encode the semantic contents and train an extra decoder using the reconstruction loss. 
Following MAE~\cite{MaskedAutoencoders2021}, several works~\cite{pang2022masked,zhang2022point_m2ae} simply the MIM pretraining process. Point-MAE~\cite{pang2022masked} divides the input point cloud into irregular point patches and randomly masks them at a high ratio. 
Then, it uses a standard transformer-based autoencoder to reconstruct the masked points. Point-M2AE~\cite{zhang2022point_m2ae} designs a multiscale MIM pretraining by making the encoder and decoder into pyramid architectures to model spatial geometries and multilevel semantics progressively. 
Meanwhile, benefiting from the same transformer architecture for point cloud and image, several works adopt image pre-trained standard transformer by distilling the knowledge from large-scale image dataset pre-trained models.

\noindent$\bullet$
\textbf{Instance Level Point Cloud Segmentation.} As shown in Sec.~\ref{sec:background}, previous PCIS / PCPS approaches are based on manually-tuned components, including a voting mechanism that predicts hand-selected geometric features for top-down approaches and heuristics for clustering the votes for bottom-up approaches. Both approaches involve many hand-crafted components and post-processing,
The usage of transformers in instance-level point cloud segmentation is similar to the image or video domain, and most works use bipartite matching for instance-level masks for indoor and outdoor scenes.
For example, Mask3D~\cite{Schult23ICRAMask3D} proposes the first Transformer-based approach for 3D semantic instance segmentation. It models each object instance as an instance query and uses the transformer decoder to refine each instance query by attending to point cloud features at different scales. 
Meanwhile, SPFormer~\cite{sun2022superpoint} learns to group the potential features from point clouds into super-points~\cite{landrieu2018large_supper_points}, and directly predicts instances through instance query with a masked-based transformer decoder. 
The super-points utilize geometric regularities to represent homogeneous neighboring points, which is more efficient than all point features. 
The transformer decoder works similarly to Mask2Former, where the cross-attention between instance query and super-point features is guided by the attention mask from the previous stage. 
PUPS~\cite{su2023pups} proposes a unified PPS system for outdoor scenes. It presents two types of learnable queries named semantic score and grouping score. The former predicts the class label for each point, while the latter indicates the probability of grouping ID for each point. 
Then, both queries are refined via grouped point features, which share the same ideas from previous Sparse-RCNN~\cite{peize2020sparse} and K-Net~\cite{zhang2021knet}. 
Moreover, PUPS also presents a context-aware mixing to balance the training instance samples, which achieves the new state-of-the-art results~\cite{semantic_kitti}.

\subsection{Tuning Foundation Models}
\label{sec:joint_learnig_with_VLM}

We divide this section into two aspects: vision adapter design and open vocabulary learning. The former introduces new ways to adapt the pre-trained large-scale foundation models for downstream tasks. The latter tries to detect and segment unknown objects with the help of the pre-trained vision language model and zero-shot knowledge transfer on unseen segmentation datasets. 
The core idea for vision adapter design is to extract the knowledge of foundation models and design better ways to fit the downstream settings. For open vocabulary learning, the core idea is to align pre-trained VLM features into current detectors to achieve novel class classification. 

\noindent$\bullet$
\textbf{Vision Adapter and Prompting Modeling.} Following the idea of prompt tuning in NLP, early works~\cite{zhou2022cocoop,zhang2021tip_adapter} adopt learnable parameters with the frozen foundation models to better transfer the downstream datasets. These works use small image classification datasets for verification and achieve better results than original zero-shot results~\cite{jia2021scaling_align}.
Meanwhile, there are several works~\cite{lin2022frozen} designing adapter and frozen foundation models for video recognition tasks. In particular, the pre-trained parameters are frozen, and only a few learnable parameters or layers are tuned. Following the idea of learnable tuning, recent works~\cite{chen2022vitadapter, rao2022denseclip} extend the vision adapter into dense prediction tasks, including segmentation and detection. 
In particular, ViT-Adapter~\cite{chen2022vitadapter} proposes a spatial prior module to solve the issue of the location prior assumptions in ViTs. 
The authors design a two-stream adaption framework using deformable attention and achieve comparable results in downstream tasks. 
From the CLIP knowledge usage view, DenseCLIP~\cite{rao2022denseclip} converts the original image-text in CLIP to a pixel-text matching problem and uses the pixel-text score maps to guide the learning of dense prediction models. 
From the task prompt view, CLIPSeg~\cite{lueddecke22_clip_seg_prompt} builds a system
to generate image segmentations based on arbitrary prompts at test time. 
A prompt can be a text or an image where the CLIP visual model is frozen during training. In this way, the segmentation model can be turned into a different task driven by the task prompt. 
Previous works only focus on a single task. OneFormer~\cite{jain2022oneformer} extends the Mask2Former with multiple target training setting and perform segmentation driven by the task prompt. 
Moreover, using a vision adapter and text prompt can easily reduce the taxonomy problems of each dataset and learn a more general representation for different segmentation tasks.
Recently, SAM~\cite{kirillov2023segment} proposes more generalized prompting methods, including mask, points, box, and text. The authors build a larger dataset with 1 billion masks. 
SAM achieves good zero-shot performance in various segmentation datasets.

\noindent$\bullet$
\textbf{Open Vocabulary Learning.} Recent studies~\cite{zareian2021open, ViLD, D2Zero, detic, OV-DETR,PAD, zhou2023rethinking} focus on the open vocabulary and open world setting, where their goal is to detect and segment novel classes, which are not seen during the training.
Different from zero-shot learning, an open vocabulary setting assumes that large vocabulary data or knowledge can provide cues for final classification.
Most models are trained by leveraging pre-trained language-text pairs, including captions and text prompts, or with the help of VLM. 
Then, trained models can detect and segment the novel classes with the help of weakly annotated captions or existing publicly available VLM. 
In particular, VilD~\cite{ViLD} distills the knowledge from a trained open vocabulary image classification model CLIP into a two-stage detector. 
However, VilD still needs an extra visual CLIP encoder for visual distillation. 
To handle this, Forzen-VLM~\cite{kuo2022fvlm} adopts the frozen visual clip model and combines the scores of both learned visual embedding and CLIP embedding for novel class detection. 
From the data augmentation view, MViT~\cite{Maaz2022Multimodal} combines the Deformable DETR and CLIP text encoder for the open world class-agnostic detection, where the authors build a large dataset by mixing existing detection datasets. 
Motivated by the more balanced samples from image classification datasets, Detic~\cite{detic} improves the performance of the novel classes with existing image classification datasets by supervising the max-size proposal with all image labels.
OV-DETR~\cite{OV-DETR} designs the first query-based open vocabulary framework by learning conditional matching between class text embedding and query features.
Besides these open vocabulary detection settings, recent works~\cite{OpenSeg, LSeg} perform open vocabulary segmentation. 
%
%
In particular, L-Seg~\cite{LSeg} presents a new setting for language-driven semantic image segmentation and proposes a transformer-based image encoder that computes dense per-pixel embeddings according to the language inputs. 
OpenSeg~\cite{OpenSeg} learns to generate segmentation masks for possible candidates using a DETR-like transformer. 
Then it performs visual-semantic alignments by aligning each word in a caption to one or a few predicted masks. 
BetrayedCaption~\cite{wu2023betrayed} presents a unified transformer framework by joint segmentation and caption learning, where the caption part contains both caption generation and caption grounding. The novel class information is encoded into the network during training. 
With the goal of unifying different segmentation with text prompts, FreeSeg~\cite{qin2023freeseg} adopts a similar pipeline as OpenSeg to crop frozen CLIP features for novel class classification.
Meanwhile, open set segmentation~\cite{wang2021unidentified} requires the model to output class agnostic masks and enhance the generality of segmentation models. 
Recently, ODISE~\cite{xu2023odise} uses a frozen diffusion model as the feature extractor, a Mask2Former head, and joint training with caption data to perform open vocabulary panoptic segmentation.
There are also several works~\cite{gupta2021ow} focusing on open-world object detection, where the task detects a known set of object categories while simultaneously identifying unknown objects.
In particular, OW-DETR~\cite{gupta2021ow} adopts the DETR as the base detector and proposes several improvements, including attention-driven pseudo-labeling, novelty classification, and objectness scoring. 
In summary, most approaches~\cite{qin2023freeseg,xu2023side} adopt the idea of region proposal network~\cite{ren2015faster} to generate class-agnostic mask proposals via different approaches, including anchor-based and query-based decoders in Sec.~\ref{sec:method_meta}. 
Then, the open vocabulary problem turns into a region-level matching problem to match the visual region features with pre-trained VLM language embedding.

\subsection{Domain-aware Segmentation}
\label{sec:domain_aware}

\noindent$\bullet$
\textbf{Domain Adaption.} Unsupervised Domain Adaptation (UDA) aims at adapting the network trained with source (synthetic) domain into target (real) domain~\cite{cordts2016cityscapes,liu2020open} without access to target labels. 
UDA has two different settings, including semantic segmentation and object detection. 
Before ViTs, the previous works~\cite{yang2020fda} mainly design domain-invariant representation learning strategies. 
DAFormer~\cite{hoyer2022daformer} replaces the outdated backbone with the advanced transformer backbone~\cite{xie2021segformer} and proposes three training strategies, including rare class sampling, thing-class ImageNet feature loss, and a learning rate warm-up method. 
It achieves new state-of-the-art results and is a strong baseline for UDA segmentation. 
Then, HRDA~\cite{hoyer2022hrda} improves DAFormer via a multi-resolution training approach and uses various crops to preserve fine segmentation details and long-range contexts. 
Motivated by MIM~\cite{MaskedAutoencoders2021}, MIC~\cite{hoyer2022mic} proposes a masked image consistency to learn spatial context relations of the target domain as additional clues. 
MIC enforces the consistency between predictions of masked target images and pseudo-labels via a teacher-student framework. It is a plug-in module that is verified among various UDA settings. 
For detection transformers on UDA, SFA~\cite{wang2021exploring} finds feature distribution alignment on CNN brings limited improvements. Instead, it proposes a domain query-based feature alignment and a token-wise feature alignment module to enhance. 
In particular, the alignment is achieved by introducing a domain query and performing the domain classification on the decoder. 
DA-DETR~\cite{zhang2021da_detr} proposes a hybrid attention module (HAM), which contains a coordinate attention module and a level attention module along with the transformer encoder. 
A single domain-aware discriminator supervises the output of HAM. MTTrans~\cite{yu2022mttrans} presents a teacher-student framework and a shared object query strategy. 
Meanwhile, SePiCo~\cite{xie2023sepico} introduces a new framework that extracts the semantic meaning of individual pixels to learn class-discriminative and class-balanced pixel representations. It supports both CNN and Transformer architecture.
The image and object features between source and target domains are aligned at local, global, and instance levels.

\noindent$\bullet$
\textbf{Multi-Dataset Segmentation.} The goal of multi-dataset segmentation is to learn a universal segmentation model on various domains. 
MSeg~\cite{lambert2020mseg} re-defines the taxonomies and aligns the pixel-level annotations by relabeling several existing semantic segmentation benchmarks. Then, the following works try to avoid taxonomy conflicts via various approaches. For example, Sentence-Seg~\cite{yin2022devil} replaces each class label with a vector-valued embedding. 
The embedding is generated by a language model~\cite{BERT}. To further handle inflexible one-hot common taxonomy, LMSeg~\cite{lmseg} extends such embedding with learnable tokens~\cite{zhou2022cocoop} and proposes a dataset-specific augmentation for each dataset. 
It dynamically aligns the segment queries in MaskFormer~\cite{cheng2021maskformer} with the category embeddings for both SS and PS tasks. 
Meanwhile, there are several works on multi-dataset object detection~\cite{Zhou_2022_CVPR_mutl_data_det,meng2022detectionhub}.
In particular, Detection-Hub~\cite{meng2022detectionhub} proposes to adapt object queries on language embedding of categories per dataset. 
Rather than previously shared embedding for all datasets, it learns semantic bias for each dataset based on the common language embedding to avoid the domain gap. 
Meanwhile, several works~\cite{hoyer2023domain,zhao2023style} focus on segmentation domain generation, which directly transfers learned knowledge from one domain to the remaining domains.
TarVIS~\cite{athar2023tarvis} jointly pre-trains one video segmentation model for different tasks spanning multiple benchmarks, where it extends Mask2Former into the video domain and adopts the unified image datasets pretraining and video fine-tuning. 
Recently, OMG-Seg~\cite{OMGSeg} has unified multi-dataset segmentation, image/video segmentation, and open-vocabulary segmentation in one shared model and achieved using one model to segment all entities.

\subsection{Label and Model Efficient Segmentation}
\label{sec:label_efficient}

\noindent$\bullet$
\textbf{Weakly Supervised Segmentation.} Weakly supervised segmentation methods learn segmentation with weaker annotations, such as image labels and object boxes. 
For weakly supervised semantic segmentation, previous works~\cite{xu2022multitoke_wsss,wang2020self} improve the typical CNN pipeline with class activation maps (CAM) and use refined CAM as training labels, which requires an extra model for training. 
ViT-PCM~\cite{rossetti2022max} shows the self-supervised transformers~\cite{chen2021mocov3} with a global max pooling can leverage patch features to negotiate pixel-label probability and achieve end-to-end training and test with one model. 
MCTformer~\cite{xu2022multitoke_wsss} adopts the idea that the attended regions of the one-class token in the vision transformer can be leveraged to form a class-agnostic localization map. 
It extends to multiple classes by using multiple class tokens to learn interactions between the class tokens and the patch tokens to generate the segmentation labels. 
For weakly supervised instance segmentation, previous works~\cite{hsu2019bbtp,lan2021discobox,tian2021boxinst} mainly leverage the box priors to supervise mask heads. 
Recently, MAL~\cite{hsu2019bbtp} shows that vision transformers are good mask auto-labelers. It takes the box-cropped images as inputs and adopts a teacher-student framework, where the two vision transformers are trained with multiple instances loss~\cite{hsu2019bbtp}. 
MAL proves the zero-shot segmentation ability and achieves nearly mask-supervised performance on various baselines. 
Meanwhile, several works~\cite{xu2022groupvit, yi2023simple} explore the text-only supervision for semantic segmentation. One representative work, GroupViT~\cite{xu2022groupvit} adopts ViT to group image regions into progressively larger shaped segments.

\noindent$\bullet$
\textbf{Unsupervised Segmentation.}
Unsupervised segmentation performs segmentation without any labels~\cite{maskfreevis}. Before ViTs, recent progress~\cite{van2021unsupervised} leverages the ideas from self-supervised learning. 
DINO~\cite{caron2021emergingDINO} finds that the self-supervised ViT features naturally contain explicit information on the segmentation of input image. 
It finds that the attention maps between the $\mathrm{CLS}$ token and feature to describe the segmentation of objects. 
Instead of using the CLS token, LOST~\cite{simeoni2021localizingLOST} solves unsupervised object discovery by using the key component of the last attention layer for computing the similarities between the different patches. 
%
%
Several works are aiming at finding the semantic correspondence of multiple images.
Then, by utilizing the correspondence maps as guidance, they achieve better performance than DINO.
Given a pair of images, SETGO~\cite{hamilton2022unsupervisedSETGO} finds the self-supervised learned features of DINO have semantically consistent correlations. 
It proposes to distill unsupervised features into high-quality discrete semantic labels. 
Motivated by the success of VLM, ReCo~\cite{shin2022reco} adopts the language-image pre-trained model, CLIP, to retrieve large unlabeled images by leveraging the correspondences in deep representation.
Then, it performs co-segmentation among both input and retrieved images. There are also several works adopting sequential pipelines. 
MaskDistill~\cite{van2022discoveringMaskDistill} firstly identifies groups of pixels that likely belong to the same object with a bottom-up model. 
Then, it clusters the object masks and uses the result as pseudo ground truths to train an extra model. Finally, the output masks are selected from the offline model according to the object score. 
FreeSOLO~\cite{wang2022freesolo} firstly adopts an extra self-supervised trained model to obtain the coarse masks. Then, it trains a SOLO-based instance segmentation model via weak supervision. 
%
%
CutLER~\cite{wang2023cut} proposes a new framework for multiple object mask generation. It first designs the MaskCut to discover multiple coarse masks based on the self-supervised features (DINO). Then, it adopts a detector to recall the missing masks via a loss-dropping strategy. Finally, it further refines mask quality via self-training


\noindent$\bullet$
\textbf{Mobile Segmentation.} Most transformer-based segmentation methods have huge computational costs and memory requirements, which make these methods unsuitable for mobile devices. 
Different from previous real-time segmentation methods~\cite{BiSeNet,zhao2017icnet,sfnet}, the mobile segmentation methods need to be deployed on mobile devices with considering both power cost and latency. Several earlier works~\cite{maaz2023edgenext,mehta2021mobilevit,zhang2023rethinking,liang2022expediting,zhou2023edgesam,xu2024rapsam} focus on a more efficient transformer backbone. 
In particular, Mobile-ViT~\cite{mehta2021mobilevit} introduces the first transformer backbone for mobile devices. It reduces image patches via MLPs before performing MHSA and shows better task-level generalization properties. There have also been several works on designing mobile semantic segmentation using transformers. 
TopFormer~\cite{zhang2022topformer} proposes a token pyramid module that takes the tokens from various scales as input to produce the scale-aware semantic feature. 
SeaFormer~\cite{wan2023seaformer} proposes a squeeze-enhanced axial transformer that contains a generic attention block. The block mainly contains two branches: a squeeze axial attention layer to model efficient global context and a detail enhancement module to preserve the details. 
RAP-SAM~\cite{xu2024rapsam} proposes a new unified setting to put real-time interactive segmentation, panoptic segmentation, and video segmentation into one framework.

\vspace{-2mm}
\subsection{Class Agnostic Segmentation and Tracking} 
\label{sec:class_agnostic}

\noindent$\bullet$
\textbf{Fine-grained Object Segmentation.}
\label{sec:fine_grained}
Several applications, such as image and video editing, often need fine-grained details of object mask boundaries. 
Earlier CNN-based works focus on refining the object masks with extra convolution modules~\cite{kirillov2020pointrend}, or extra networks~\cite{cheng2020cascadepsp}. 
Most transformer-based approaches~\cite{transfiner,video_transfiner,liu2022simpleclick,SegRefiner, song2024basam} adopt vision transformers due to their fine-grained multiscale features and long-range context modeling. 
Transfiner~\cite{transfiner} refines the region of the coarse mask via a quad-tree transformer. By considering multiscale point features, it produces more natural boundaries while revealing details for the objects. 
Then, Video-Transfiner~\cite{video_transfiner} refines the spatial-temporal mask boundaries by applying Transfiner~\cite{transfiner} to the video segmentation method~\cite{VIS_TR}. 
It can refine the existing video instance segmentation datasets~\cite{vis_dataset}. 
PatchDCT~\cite{wen2023patchdct} adopts the idea of ViT by making object masks into patches. 
Then, each mask is encoded into a DCT vector~\cite{shen2021dct}, and PatchDCT designs a classifier and a regressor to refine each encoded patch. 
Entity segmentation~\cite{qi2022openentity} aims to segment all visual entities without predicting their semantic labels. 
Its goal is to obtain high-quality and generalized segmentation results. 
%

\noindent$\bullet$
\textbf{Video Object Segmentation.} \label{sec:vos} Recent approaches for VOS mainly focus on designing better memory-based matching methods~\cite{stm_vos}. 
Inspired by the Non-local network~\cite{wang2018nonlocal} in image recognition tasks, the representative work STM~\cite{stm_vos} is the first to adopt cross-frame attention, where previous features are seen as memory. 
Then, the following works~\cite{yang2021associating} design a better memory-matching process. associating objects with transformers (AOT)~\cite{yang2021associating} matches and decodes multiple objects jointly. 
The authors propose a novel hierarchical matching and propagation, named long short-term transformer, where they joint persevere an identity bank and long-short term attention. 
XMem~\cite{cheng2022xmem} proposes a mixed memory design to handle the long video inputs. 
The mixed memory design is also based on the self-attention architecture. 
Meanwhile, Clip-VOS~\cite{park2022perclip_vos} introduces per-clip memory matching for inference efficiency. 
Recently, to enhance instance-level context, Wang~\etal~\cite{Wang2022LookBY} adds an extra query from Mask2Former into memory matching for VOS.

\subsection{Medical Image Segmentation}
\label{sec:medical_image_segmentation_review} 
CNNs have achieved milestones in medical image analysis. In particular, the U-shaped architecture and skip-connections~\cite{ronneberger2015_unet,isensee2021nnu} have been widely applied in various medical image segmentation tasks. 
With the success of ViTs, recent representative works~\cite{chen2021transunet,cao2021swinunet} adopt vision transformers into the U-Net architecture and achieve better results. 
TransUNet~\cite{chen2021transunet} merges transformer and U-Net, where the transformer encodes tokenized image patches to build the global context. 
Then decoder upsamples the encoded features, which are then combined with the high-resolution CNN feature maps to enable precise localization. 
Swin-Unet~\cite{cao2021swinunet} designs a symmetric Swin-like~\cite{liu2021swin} decoder to recover fine details. 
TransFuse~\cite{zhang2021transfuse} combines transformers and CNNs in a parallel style, where global dependency and low-level spatial details can be efficiently captured jointly. 
UNETR~\cite{hatamizadeh2022unetr} focuses on 3D input medical images and designs a similar U-Net-like architecture. The encoded representations of different layers in the transformer are extracted and merged with a decoder via skip connections to get the final 3D mask outputs.

\section{Benchmark Results}
\label{sec:benchmark}
%

In this section, we report recent transformer-based visual segmentation and tabulate the performance of previously discussed algorithms. For each reviewed field, the most widely used datasets are selected for performance benchmark in Sec.~\ref{sec:main_image_results} and Sec.~\ref{sec:main_results_video_segmentation_benchmarks}. 
We further re-benchmark several representative works in Sec.~\ref{sec:our_re_becnmarking} using {the same data augmentations} and {feature extractor}. 
Note that we only list \textbf{\textit{published works}} for reference. For simplicity, we have excluded several works on representation learning and only present specific segmentation methods. For a comprehensive method comparison, please refer to the supplementary material that provides a more detailed analysis. In addition, several works~\cite{zhang2023simple,zou2023segment,HIPIE} achieve better results. However, due to the extra datasets~\cite{objects365} they used, we do not list them here.

\subsection{Main Results on Image Segmentation Datasets}
\label{sec:main_image_results}

\begin{table}[htbp]
\centering
\caption{\textbf{Benchmark results on semantic segmentation validation datasets.} The results are with mIoU metric.}\vspace{-2mm}
\label{tab:sem_seg_datasets_results}
\begin{adjustbox}{width=0.50\textwidth}
\begin{tabular}{c c c c c c }
\toprule[0.15em]
\textbf{Method}  & backbone & COCO-Stuff & Cityscapes & ADE20K & Pascal-Context  \\ 
\midrule[0.15em]
SETR~\cite{SETR} & ViT-Large & - & 82.2 & 50.3 & 55.8   \\
Segmenter~\cite{strudel2021_segmenter} & ViT-Large &- & 81.3 & 53.6 & 59.0 \\ 
SegFormer~\cite{xie2021segformer} & MiT-B5 & 46.7 &  84.0 & 51.8 & -   \\
SegNext~\cite{guo2022segnext} & MSCAN-L & 47.2 & 83.9 & 52.1 & 60.9  \\
ConvNext~\cite{liu2022convnet} & ConvNeXt-XL & - &- & 54.0 & -  \\
MAE~\cite{MaskedAutoencoders2021} &  ViT-L & - & - & 53.6 & -  \\
K-Net~\cite{zhang2021knet} & Swin-L & - & - & 54.3 & -   \\
MaskFormer~\cite{cheng2021maskformer} & Swin-L & - & - & 55.6 & - \\
Mask2Former~\cite{cheng2021mask2former} & Swin-L & - & 84.3 & 57.3 & -   \\
CLUSTSEG~\cite{liang2023clustseg} &  Swin-B & - & - & 57.4 & - \\
OneFormer~\cite{jain2022oneformer} & ConvNeXt-XL & - & 84.6 & 58.8 & - \\
\bottomrule
\end{tabular}
\end{adjustbox}
\end{table}

\begin{table}[htbp]
\centering
\caption{\textbf{Benchmark results on instance segmentation of coco validation datasets.} The results are with the mAP metric.}
\label{tab:instance_seg_results}
\begin{adjustbox}{width=0.50\textwidth}
\begin{tabular}{c c c c c c c c c }
\toprule[0.15em] 
\textbf{Method}  & backbone & AP & $AP^{50}$ & $AP^{75}$ & $AP^{S}$ & $AP^{M}$ & $AP^{L}$  \\ 
\midrule[0.15em]
SOLQ~\cite{dong2021solq}  & ResNet50 & 39.7&-&-&21.5&42.5&53.1 \\
K-Net~\cite{zhang2021knet} & ResNet50 & 
38.6&60.9&41.0&19.1&42.0&57.7\\
Mask2Former~\cite{cheng2021mask2former} & ResNet50 & 43.7 & - & - & 23.4 & 47.2 & 64.8  \\
Mask DINO~\cite{li2022maskdino} & ResNet50 & 46.3 & 69.0 & 50.7 & 26.1 & 49.3 & 66.1\\
\hline
Mask2Former~\cite{cheng2021mask2former} & Swin-L & 50.1&-&-&29.9&53.9&72.1 \\
Mask DINO~\cite{li2022maskdino} & Swin-L & 52.3 & 76.6&57.8&33.1&55.4&72.6 \\
OneFormer~\cite{jain2022oneformer} & Swin-L & 49.0 & - & - & - & - & - \\
\bottomrule
\end{tabular}
\end{adjustbox}
\end{table}

\begin{table}[htbp]
\centering
\caption{\textbf{Benchmark results on panoptic segmentation validation datasets.} The results are with PQ metric.}
\label{tab:panoptic_seg_results}
\begin{adjustbox}{width=0.50\textwidth}
\begin{tabular}{c c c c c c}
\toprule[0.15em]
\textbf{Method}  & backbone & COCO & Cityscapes & ADE20K &  Mapillary  \\ 
\midrule[0.15em]
PanopticSegFormer~\cite{li2021panoptic} & ResNet50 & 49.6 & - & 36.4 & - \\
K-Net~\cite{zhang2021knet} & ResNet50 & 47.1 & - & - & -  \\
Mask2Former~\cite{cheng2021mask2former} & ResNet50 & 51.9 & 62.1 & 39.7 & 36.3  \\
K-Max Deeplab~\cite{kmax_deeplab_2022} & ResNet50 & 53.0 & 64.3 & 42.3 & - \\
Mask DINO~\cite{li2022maskdino} & ResNet50 & 53.0 & - & - & - \\
\hline
Max-Deeplab~\cite{wang2020maxDeeplab} & MaX-L & 51.1 & - & - & -  \\
PanopticSegFormer~\cite{li2021panoptic} & Swin-L & 55.8 & - & - & - \\
Mask2Former~\cite{cheng2021mask2former} & Swin-L & 57.8 & 66.6 & 48.1 & 45.5 \\
CMT-Deeplab~\cite{yu2022cmt} & Axial-R104-RFN & 55.3 & - & - & -  \\
K-Max Deeplab~\cite{kmax_deeplab_2022} & ConvNeXt-L & 58.1 & 68.4 & 50.9 & - \\
OneFormer~\cite{jain2022oneformer} & Swin-L & 57.9 & 67.2 & 51.4 & - \\
Mask DINO~\cite{li2022maskdino} & Swin-L & 58.3 & - & - & - \\
CLUSTSEG~\cite{liang2023clustseg} & Swin-B & 59.0 & - & - & - \\
\bottomrule
\end{tabular}
\end{adjustbox}
\end{table}

\begin{table}[htbp]
\centering
\caption{\textbf{Benchmark results on video semantic segmentation of VPSW validation datasets.} The results are with mIoU and mVC (mean Video Consistency) metrics.}
\label{tab:video_semantic_seg_results}
\begin{adjustbox}{width=0.45\textwidth}
\begin{tabular}{c c c c c }
\toprule[0.15em]
\textbf{Method}  & backbone & mIoU & $mVC_{8}$ & $mVC_{16}$ \\ 
\midrule[0.15em]
TCB~\cite{miao2021vspw} & ResNet101 & 37.5 &  87.0  &  82.1 \\
Video K-Net~\cite{li2022videoknet} & ResNet101 & 38.0 & 87.2 & 82.3 \\
CFFM~\cite{sun2022vss} & MiT-B5 & 49.3 & 90.8 & 87.1\\
MRCFA~\cite{sun2022mining} & MiT-B5 & 49.9 & 90.9  & 87.4 \\
TubeFormer~\cite{kim2022tubeformer} & Axial-ResNet & 63.2 & 92.1 & 87.9
 \\
\bottomrule
\end{tabular}
\end{adjustbox}
\end{table}

\noindent$\bullet$
\noindent
\textbf{Results On Semantic Segmentation Datasets.} In Tab.~\ref{tab:sem_seg_datasets_results}, Mask2Former~\cite{cheng2021mask2former} and OneFormer~\cite{jain2022oneformer} perform the best on Cityscapes and ADE20K dataset, while SegNext~\cite{guo2022segnext} achieves the best results on COCO-Stuff and Pascal-Context datasets.

\noindent$\bullet$
\noindent
\textbf{Results on COCO Instance Segmentation.} In Tab.~\ref{tab:instance_seg_results}, Mask DINO~\cite{li2022maskdino} achieves the best results on the COCO instance segmentation with both ResNet and Swin-L backbones.

\noindent$\bullet$
\noindent
\textbf{Results on Panoptic Segmentation.} In Tab.~\ref{tab:panoptic_seg_results}, for panoptic segmentation, Mask DINO~\cite{li2022maskdino} and K-Max Deeplab~\cite{kmax_deeplab_2022} achieve the best results on the COCO dataset. K-Max Deeplab also achieves the best results on Cityscapes. OneFormer~\cite{jain2022oneformer} performs the best on ADE20K.

\subsection{Re-Benchmarking For Image Segmentation}
\label{sec:our_re_becnmarking}

\noindent$\bullet$
\textbf{Motivation.} We perform re-benchmarking on two segmentation tasks: semantic segmentation and panoptic segmentation on four public datasets, including ADE20K, COCO, Cityscapes, and COCO-Stuff datasets. In particular, we want to explore the effect of the transformer decoder. Thus, we use the same encoder~\cite{resnet} and neck architecture~\cite{zhu2020deformabledetr} for a fair comparison. 

\begin{table}[htbp]
\centering
\caption{\textbf{Experiment results on semantic segmentation datasets.} The results are with mIoU metric.}\vspace{-2mm}
\label{tab:experiments_res_sem_seg}
\begin{adjustbox}{width=0.50\textwidth}
\begin{tabular}{c c c c c c c}
\toprule[0.15em]
\textbf{Method} & backbone & COCO-Stuff & Cityscapes & ADE20K  & Param & FPS  \\ 
\midrule[0.15em]
Segformer+~\cite{xie2021segformer} & MiT-B2  & 41.6 &  81.6 & 47.5 & 30.5 & 25.5 \\
K-Net+~\cite{zhang2021knet} & MiT-B2 & 36.3 & 81.4 & 45.9 & 36.2  & 23.3 \\
K-Net+~\cite{zhang2021knet} & ResNet50 & 35.2 & 81.3 & 43.9 & 40.2  & 20.3 \\
MaskFormer+~\cite{cheng2021maskformer} & ResNet50 & 37.1 & 80.1 & 44.9 & 45.0 & 23.7 \\
Mask2Former~\cite{cheng2021mask2former} &  ResNet50 & 38.8  & 80.4  & 48.0 & 44.0 &  19.4 \\
\bottomrule
\end{tabular}
\end{adjustbox}
\end{table}

\begin{table}[htbp]
\centering
\caption{\textbf{Experiment results on instance segmentation datasets.} We report results on the validation set using the ResNet50 backbone. The results are with mAP metric.}
\label{tab:experiments_res_ins_seg}
\begin{adjustbox}{width=0.50\textwidth}
\begin{tabular}{c c c c c c c c}
\toprule[0.15em]
\textbf{Method}  & mAP & AP@50 & AP@75 & APs & APm & APl  & FPS  \\ 
\midrule[0.15em]
QueryInst~\cite{QueryInst} & 40.7 & 62.7 & 44.4 & 20.6 & 43.9 & 60.6 & 20.9 \\
SOLQ~\cite{dong2021solq} &  39.6 & 61.2 & 43.2 & 19.2 & 44.2 & 60.2 & 18.2 \\
K-Net+~\cite{zhang2021knet} & 41.5 & 64.2 & 44.4 & 20.3 & 45.2 & 62.4 & 21.2 \\
MaskFormer+~\cite{cheng2021mask2former} & 36.9 & 58.4 & 38.9 & 16.3 & 39.2 & 58.1 & 15.6 \\
Mask2Former~\cite{cheng2021mask2former} & 43.1 & 65.6 & 46.5 & 22.7 & 46.7 & 64.7 & 13.4 \\
\bottomrule
\end{tabular}
\end{adjustbox}
\end{table}

\begin{table}[htbp]
\centering
\caption{\textbf{Experiment results on panoptic segmentation datasets.} We report results on the validation set using the ResNet50 backbone. The results are with PQ metric.}
\label{tab:experiments_res_pano_seg}
\begin{adjustbox}{width=0.50\textwidth}
\begin{tabular}{c c c c c c}
\toprule[0.15em]
\textbf{Method}  & COCO & Cityscapes & ADE20K  & Param & FPS  \\ 
\midrule[0.15em]
PanopticSegFormer~\cite{li2021panoptic} & 50.1 & 60.2 & 36.4 & 51.0  & 10.3\\
K-Net+~\cite{zhang2021knet} &  49.2 &  59.7 &  35.1 & 40.2 & 21.2 \\
MaskFormer+~\cite{cheng2021mask2former} & 47.2 & 53.1 & 36.3 & 45.1 & 13.4\\
Mask2Former~\cite{cheng2021mask2former} & 52.1 & 62.3 & 39.2 & 44.0 & 15.6 \\
YOSO~\cite{hu2023you} & 49.2 &  59.3 & 38.2 & 42.2 & 35.2  \\
\bottomrule
\end{tabular}
\end{adjustbox}
\end{table}

\noindent$\bullet$
\textbf{Results on Semantic Segmentation.} As shown in Tab.~\ref{tab:experiments_res_sem_seg}, we carry out re-benchmark experiments for SS. In particular, using the same neck architecture, Segformer+~\cite{xie2021segformer} achieves the best results on COCO-Stuff and Cityscapes. Mask2Former achieves the best result on the ADE-20k dataset.


\noindent$\bullet$
\textbf{Results on Instance Segmentation.} In Tab.~\ref{tab:experiments_res_ins_seg}, we also explore the instance segmentation methods on COCO datasets. Under the same neck architecture, we observe gains on both K-Net and MaskFormer, compared with origin results in Tab.~\ref{tab:instance_seg_results}. Mask2Former achieve the best results.

\noindent$\bullet$
\textbf{Results on Panoptic Segmentation.} In Tab.~\ref{tab:experiments_res_pano_seg}, we present the re-benchmark results for PS. In particular, Mask2Former achieves the best results on all three datasets. Compared with K-Net and MaskFormer, both K-Net+ and MaskFormer+ achieve over 3-4\% improvements due to the usage of stronger neck~\cite{zhu2020deformabledetr}, which close the gaps between their original results and Mask2Former.


\subsection{Main Results for Video Segmentation Datasets}

\noindent$\bullet$
\noindent
\textbf{Results On Video Semantic Segmentation} In Tab.~\ref{tab:video_semantic_seg_results}, we report VSS results on VPSW. Among the methods, TubeFormer~\cite{kim2022tubeformer} achieves the best results.

\noindent$\bullet$
\noindent
\textbf{Results on Video Instance Segmentation} In Tab.~\ref{tab:video_instance_seg_results}, for VIS, CTVIS~\cite{ying2023ctvis} achieves the best result on YT-VIS-2019 and YT-VIS-2021 using ResNet50 backbone.
GenVIS~\cite{heo2022generalized} achieves better results on OVIS using ResNet50 backbone.
When adopting Swin-L backbone, CTVIS~\cite{ying2023ctvis} achieves the best results.

\noindent$\bullet$
\textbf{Results on Video Panoptic Segmentation}
\label{sec:main_results_video_segmentation_benchmarks} In Tab.~\ref{tab:video_panoptic_seg_results}, for VPS, SLOT-VPS~\cite{zhou2022slot} achieves the best results on Cityscapes-VPS. TubeLink~\cite{tubelink} achieves the best results on the VIP-Seg dataset. Video K-Net~\cite{li2022videoknet} achieves the best results on the KITTI-STEP dataset.

\begin{table}[htbp]
\centering
\caption{\textbf{Benchmark results on video instance segmentation validation dataset.} The results are with the mAP metric.}
\label{tab:video_instance_seg_results}
\begin{adjustbox}{width=0.50\textwidth}
\begin{tabular}{c c c c c}
\toprule[0.15em]
\textbf{Method}  & backbone & YT-VIS-2019 & YT-VIS-2021 & OVIS \\ 
\midrule[0.15em]
VISTR~\cite{VIS_TR} & ResNet50 & 36.2 & - & -  \\
IFC~\cite{IFC_21} & ResNet50 & 42.8 & 36.6 & -\\
Seqformer~\cite{seqformer} & ResNet50 & 47.4 & 40.5 & - \\
Mask2Former-VIS~\cite{cheng2021mask2formervis} & ResNet50 & 46.4 & 40.6 & -\\
IDOL~\cite{IDOL} & ResNet50 & 49.5 & 43.9 & 30.2\\
VITA~\cite{heo2022vita} & ResNet50 & 49.8 & 45.7 & 19.6 \\
Min-VIS~\cite{huang2022minvis} & ResNet50 & 47.4 & 44.2 & 25.0 \\
GenVIS~\cite{heo2022generalized} & ResNet50 & 51.3 & 46.3 & 35.8 \\
Tube-Link~\cite{tubelink} & ResNet50 & 52.8 & 47.9 &  29.5 \\
CTVIS~\cite{ying2023ctvis} & ResNet50 & 55.1 & 50.1 & 35.5 \\
\hline
SeqFormer~\cite{seqformer} & Swin-L & 59.3 & 51.8 & - \\
Mask2Former-VIS~\cite{cheng2021mask2formervis} & Swin-L & 60.4 & 52.6 & -\\
IDOL~\cite{IDOL} & Swin-L & 64.3 & 56.1 & 42.6\\
VITA~\cite{heo2022vita} & Swin-L & 63.0 & 57.5 & 27.7\\
Min-VIS~\cite{huang2022minvis} & Swin-L & 61.6 & 55.3 & 39.4 \\
GenVIS~\cite{heo2022generalized} & Swin-L & 63.8 & 60.1 & 45.4 \\
Tube-Link~\cite{tubelink} & Swin-L & 64.6 & 58.4 & - \\
CTVIS~\cite{ying2023ctvis} & Swin-L & 65.6 & 61.2 & 46.9 \\
\bottomrule
\end{tabular}
\end{adjustbox}
\end{table}

\begin{table}[htbp]
\centering
\caption{\textbf{Benchmark results on video panoptic segmentation validation datasets.} The results are with VPQ and STQ metrics.}
\label{tab:video_panoptic_seg_results}
\begin{adjustbox}{width=0.50\textwidth}
\begin{tabular}{c c c c c c}
\toprule[0.15em]
{\scriptsize \textbf{Method}}  & {\scriptsize backbone} & {\scriptsize Cityscapes-VPS (VPQ)} & {\scriptsize KITTI-STEP (STQ)} & {\scriptsize VIP-Seg (STQ)} \\ 
\midrule[0.15em]
VIP-Deeplab~\cite{vip_deeplab} & ResNet50 & 60.6 & - & -\\
PolyphonicFormer~\cite{yuan2021polyphonicformer} & ResNet50 & 65.4 & - & -\\
Tube-PanopticFCN~\cite{miao2022large} & ResNet50 & - & - & 31.5 \\ 
TubeFormer~\cite{kim2022tubeformer} & Axial-ResNet & - & 70.0 & -\\ 
Video K-Net~\cite{li2022videoknet} & Swin-B & 62.2 & 73.0 & 46.3 \\ 
TubeLink~\cite{tubelink} & Swin-B & - & 72.0  & 49.4  \\
SLOT-VPS~\cite{zhou2022slot} & Swin-L & 63.7 & - & - \\
\bottomrule
\end{tabular}
\end{adjustbox}
\end{table}

\section{Future Directions}
\label{sec:future_work}

\noindent$\bullet$
\textbf{General and Unified Image/Video Segmentation.} 
{The trend of using transformers to unify diverse segmentation tasks is gaining traction. Recent studies~\cite{zhang2021knet,li2022videoknet,wang2020maxDeeplab,kim2022tubeformer,jain2022oneformer,OMGSeg,wu2023uniref++} have employed query-based transformers for various segmentation tasks within a unified architecture. A promising research avenue is the integration of image and video segmentation tasks in a universal model across different datasets. Such models may achieve general, robust segmentation capabilities in multiple scenarios, like detecting rare classes for improved robotic decision-making. This approach holds significant practical value, particularly in applications like robot navigation and autonomous vehicles.}

\noindent$\bullet$
\textbf{Joint Learning with Multi-Modality.} 
{Transformers' inherent flexibility in handling various modalities positions them as ideal for unifying vision and language tasks. Segmentation tasks, which offer pixel-level information, can enhance associated vision-language tasks such as text-image retrieval and caption generation~\cite{phraseclick}. Recent studies~\cite{lu2022unified,zou2022xdecoder,qi2024generalizable,yuan2024open} demonstrate the potential of a universal transformer architecture that concurrently learns segmentation alongside visual language tasks, paving the way for integrated multi-modal segmentation learning.}

\noindent$\bullet$
\noindent
\textbf{Life-Long Learning for Segmentation.} Existing segmentation methods are usually benchmarked on closed-world datasets with a set of predefined categories, \ie, assuming that the training and testing samples have the same categories and feature spaces that are known beforehand. However, realistic scenarios are usually open-world and non-stationary, where novel classes may occur continuously~\cite{zhang2021prototypical}. 
For example, unseen situations can occur unexpectedly in self-driving vehicles and medical diagnoses. There is a distinct gap between the performance and capabilities of existing methods in realistic and open-world settings. 
Thus, it is desired to gradually and continuously incorporate novel concepts into the existing knowledge base of segmentation models, making the model capable of lifelong learning.

\noindent$\bullet$
\textbf{Long Video Segmentation in Dynamic Scenes.} Long videos introduce several challenges~\cite{MOSE,MeViS,zhou2024dvis}. First, existing video segmentation methods are designed to work with short video inputs and may struggle to associate instances over longer periods. Thus, new methods must incorporate long-term memory design and consider the association of instances over a more extended period. Second, maintaining segmentation mask consistency over long periods can be difficult, especially when instances move in and out of the scene. This requires new methods to incorporate temporal consistency constraints and update the segmentation masks over time. Third, heavy occlusion can occur in long videos, making it challenging to segment all instances accurately. New methods should incorporate occlusion reasoning and detection to improve segmentation accuracy. Finally, long video inputs often involve various scene inputs, which can bring domain robustness challenges for video segmentation models. New methods must incorporate domain adaptation techniques to ensure the model can handle diverse scene inputs. In short, addressing these challenges requires the development of new long video segmentation models that incorporate advanced memory design, temporal consistency constraints, occlusion reasoning, and detection techniques.

\noindent$\bullet$
\textbf{Generative Segmentation.} With the rise of stronger generative models, recent works~\cite{chen2022generalist, wang2024explore, xie2023mosaicfusion} solve image segmentation problems via generative modeling, inspired by a stronger transformer decoder and high-resolution representation in the diffusion model~\cite{rombach2021highresolution}. Adopting a generative design avoids the transformer decoder and object query design, which makes the entire framework simpler. However, these generative models typically introduce a complicated training pipeline. A simpler training pipeline is needed for further research.

\noindent$\bullet$
\textbf{Segmentation with Visual Reasoning.}
Visual reasoning~\cite{johnson2015image,pvsg,yang_psg,psg4d,wang2023pair} requires the robot to understand the connections between objects in the scene, and this understanding plays a crucial role in motion planning. Previous research has explored using segmentation results as input to visual reasoning models for various applications, such as object tracking and scene understanding. Joint segmentation and visual reasoning can be a promising direction, with the potential for mutual benefits for both segmentation and relation classification. By incorporating visual reasoning into the segmentation process, researchers can leverage the power of reasoning to improve the segmentation accuracy, while segmentation can provide better input for visual reasoning.

\vspace{-4mm}
\section{Conclusion}
\label{sec:conclusion}
This survey provides a comprehensive review of recent advancements in transformer-based visual segmentation, which, to our knowledge, is the first of its kind. 
The paper covers essential background knowledge and an overview of previous works before transformers and summarizes more than 120 deep-learning models for various segmentation tasks. 
The recent works are grouped into six categories based on the meta-architecture of the segmenter. 
Additionally, the paper reviews five specific subfields and reports the results of several representative segmentation methods on widely-used datasets. 
To ensure fair comparisons, we also re-benchmark several representative works under the same settings. 
Finally, we conclude by pointing out future research directions for transformer-based visual segmentation.

\noindent
\textbf{Acknowledgement.} This work is supported by The Alan Turing Institute (UK) through the project 'Turing-DSO Labs Singapore Collaboration' (SDCfP2\textbackslash100009). This study is also supported under the RIE2020 Industry Alignment Fund Industry Collaboration Projects (IAF-ICP) Funding Initiative and Singapore MOE AcRF Tier 1 (RG16/21). , as well as cash and in-kind contributions from the industry partner(s).



\appendix
\noindent
\textbf{Overview.} In this appendix, we provide more details as a supplementary adjunct to the main paper.

\begin{enumerate}
\setlength{\leftmargin}{-1em}
\setlength{\parsep}{0ex} 
\setlength{\topsep}{0ex}
\setlength{\itemsep}{0.5ex}  
\setlength{\labelsep}{0.5em} 
\setlength{\itemindent}{-0.5em} 
\setlength{\listparindent}{0em} 
\item More descriptions on task metrics. (Sec.~\ref{sec:task_metric})
\item Representative works in the section "Specific Subfields" in the main paper. (Sec.~\ref{sec:representativework_in_related_domain_beyond})
\item More detailed benchmark and re-benchmark results on remaining segmentation tasks. (Sec.~\ref{sec:more_benchmark_result})
\item Detailed experiment settings and implementation details for re-benchmarking. (Sec.~\ref{sec:re_benchmark_details})
\item More future directions. (Sec.~\ref{sec:more_future_directions})
\end{enumerate}


\begin{table*}[h]
    \centering
    \small
    \caption{Representative works summarization and comparison in the specific subfields section.}\vspace{-2mm}
    \setlength{\tabcolsep}{1.2pt}
\scalebox{0.80}{
\begin{tabular}{p{0.14\textwidth}p{0.10\textwidth}p{0.15\textwidth}p{0.20\textwidth}p{0.60\textwidth}}
    \toprule
    \belowrulesepcolor{gray!30!}
\rowcolor{gray!30!} Method & Task & Input/Output & Transformer Architecture  & \ \ ~~~~~~~~~~~~~~~~~~~~~~~~~~~~~~~~~~~~Highlight \\ \aboverulesepcolor{gray!30!} \midrule
\belowrulesepcolor{gray!15!}
\rowcolor{gray!15!}\multicolumn{5}{c}{\textbf{Point Cloud Segmentation}} \\ \aboverulesepcolor{gray!15!} \midrule
Pointformer~\cite{point_transformer} & PCSS & Point Cloud / Semantic Masks &  Pure transformer  & the first pure transformer backbone for point cloud and a vector self-attention operation. \\ 
\rowcolor{gray!10!} Stratified Transformer~\cite{lai2022stratified} & PCSS & Point Cloud / Semantic Masks & Pure transformer  & cube-wised point cloud transformer with mixed local and long-range key point sampling. \\
PointBert~\cite{yu2022pointBERT} & PCSS & Point Cloud / Semantic Masks & Pure transformer  & the first MIM-like pre-training on point cloud analysis. \\ 
\rowcolor{gray!10!}SPFormer~\cite{sun2022superpoint} & PCIS & Point Cloud / Instance Masks & Query-based decoder & using object query to perform masked cross-attention with super points. \\ 
 PUPS~\cite{su2023pups} & PCPS & Point Cloud / Panoptic Masks & Query-based decoder & a query decoder with two queries including semantic score query and grouping score query. \\ 
  \midrule\belowrulesepcolor{gray!15!}
\rowcolor{gray!15!}\multicolumn{5}{c}{\textbf{Tuning Foundation Models}} \\ \aboverulesepcolor{gray!15!} \midrule
\rowcolor{gray!10!} ViT-Adapter~\cite{chen2022vitadapter} & SS/IS/PS  &  Image/Panoptic Masks  & Transformer + CNN   &  design an extra spatial attention module to finetune the large vision models. \\ 
OneFormer~\cite{jain2022oneformer} &  SS/IS/PS & Image/Panoptic Masks  &  CNN/Transformer + query decoder  &  use task text prompt to build one model for three different segmentation tasks. \\
SAM~\cite{kirillov2023segment} & class agnostic segmentation & Image + prompts / Instance Masks & Transformer + query decoder & propose a simple yet universal segmentation framework with different prompts for class agnostic segmentation. \\

\rowcolor{gray!10!} L-Seg~\cite{LSeg} &  SS & (Image + Text) / Semantic Masks  &  Transformer + CNN decoder  & propose the first language-driven segmentation framework and use the CLIP text to segment specific classes. \\
OpenSeg~\cite{OpenSeg} & SS & (Image + Text) / Semantic Masks & Transformer + query decoder & proposes first open vocabulary segmentation model by combining class-agnostic detector and ALIGN test features. \\

 \midrule
\belowrulesepcolor{gray!15!}
\rowcolor{gray!15!}\multicolumn{5}{c}{\textbf{Domain-aware Segmentation}} \\ \aboverulesepcolor{gray!15!} \midrule
DAFormer~\cite{hoyer2022daformer} & SS & Image / Semantic Masks & Pure transformer encoder + MLP decoder & the first segmentation transformer in domain adaption and build stronger baselines than CNNs. \\ 
\rowcolor{gray!10!}SFA~\cite{wang2021exploring}  & OD & Image / Boxes & CNN + query-based transformer decoder & the first object detection and introduce a domain query for domain alignment. \\ 
LMSeg~\cite{lmseg} & SS / PS & Image / Panoptic Masks & CNN + query-based transformer decoder & introduce text prompt for multi-dataset segmentation and dataset-specific augmentation during training. \\ \midrule
\belowrulesepcolor{gray!15!}
\rowcolor{gray!15!}\multicolumn{5}{c}{\textbf{Label and Model Efficient Segmentation}}\\ \aboverulesepcolor{gray!15!}\midrule
 MCTformer~\cite{xu2022multitoke_wsss} & weakly supervised SS & Image / Semantic Masks & Pure transformer  & leverage the multiple class tokens with patch tokens to generate multi-class attention maps. \\ 
\rowcolor{gray!10!} MAL~\cite{hsu2019bbtp} &  weakly supervised SS & Image / Semantic Masks & Pure transformer & a teacher-student framework based on ViTs to learn binary masks with box supervision. \\ 
SETGO~\cite{hamilton2022unsupervisedSETGO} & unsupervised SS & Image / Semantic Masks & Pure transformer & use unsupervised features correspondences of DINO to distill semantic segmentation heads. \\
\rowcolor{gray!10!}TopFormer~\cite{zhang2022topformer} & mobile SS & Image / Semantic Masks & Pure transformer &  the first transformer-based mobile segmentation framework and a token pyramid module.  \\
\midrule\belowrulesepcolor{gray!15!}
\rowcolor{gray!15!}\multicolumn{5}{c}{\textbf{Class Agnostic Segmentation and Tracking}} \\ \aboverulesepcolor{gray!15!} \midrule
Transfiner~\cite{transfiner} & IS & Image / Instance Masks &  CNN detector + transformer &  a transformer encoder uses multiscale point features to refine coarse masks.  \\ 
\rowcolor{gray!10!}PatchDCT~\cite{wen2023patchdct}  & IS & Image / Instance Masks  &  CNN detector + transformer & a token based DCT to persevere more local details.  \\ 
STM~\cite{stm_vos} & VOS &  Video / Instance Masks &  CNN + transformer & the first attention-based VOS framework for VOS and use mask matching for propagation. \\
\rowcolor{gray!15!} AOT~\cite{yang2021associating} & VOS & Video / Instance Masks &  CNN + transformer &  a memory-based transformer to perform multiple instances mask matching jointly.  \\ 
\midrule\belowrulesepcolor{gray!15!}
\rowcolor{gray!15!}\multicolumn{5}{c}{\textbf{Medical Image Segmentation}} \\ \aboverulesepcolor{gray!15!} \midrule
 TransUNet~\cite{chen2021transunet} &  SS  & Image / Semantic Masks & CNN + transformer & the first U-Net-like architecture with transformer encoder for medical images. \\ 
 \rowcolor{gray!10!} TransFuse~\cite{zhang2021transfuse} & SS  & Image / Semantic Masks & CNN + transformer & combine transformer and CNN in a dual framework to balance the global context and local details. \\ \bottomrule
\end{tabular}
}
\label{tab:related_domain_beyond_summary}
\vspacefigtext
\end{table*}

\subsection{Task Metrics}
\label{sec:task_metric}

In this section, we present detailed descriptions for different segmentation task metrics. 

\noindent
\textbf{Mean Intersection over Union (mIoU).} It is a metric used to evaluate the performance of a semantic segmentation model. The predicted segmentation mask is compared to the ground truth segmentation mask for each class. The IoU score measures the overlap between the predicted mask and the ground truth mask for each class. The IoU score is calculated as the ratio of the area of intersection between the predicted and ground truth masks to the area of union between the two masks. The mean IoU score is then calculated as the average of the IoU scores across all the classes. A higher mean IoU score indicates better performance of the model in accurately segmenting the image into different classes. A mean IoU score of 1 indicates perfect segmentation, where the predicted and ground truth masks completely overlap for all classes. The mean IoU score indicates how well the model can separate the different objects or regions in the image only based on the semantic class. It is a commonly used metric to evaluate the performance of semantic segmentation models as well as video or point cloud semantic segmentation.

\noindent
\textbf{Mean Average Precision (mAP).} This is calculated by comparing the predicted segmentation masks with the ground truth masks for each object in the image. The mAP score is calculated by averaging the Average Precision (AP) scores across all the object categories in the image. The AP score for each object category is calculated based on the intersection-over-union (IoU) between the predicted segmentation mask and the ground truth mask. IoU measures the overlap between the two masks, and a higher IoU indicates a better match between the predicted and ground truth masks. The AP scores for all object categories are averaged to calculate the mAP score. The mAP score ranges from 0 to 1, with a higher score indicating better performance of the model in detecting and localizing objects in the image. For the COCO dataset, the mAP metric is usually reported at different IoU thresholds (typically, 0.5, 0.75, and 0.95). This measures the performance of the model at different levels. Then, the mAP score is reported as the average of these three IoU thresholds.

Similar to instance segmentation in images, the mAP score for VIS~\cite{vis_dataset} is calculated by comparing the predicted segmentation masks with the ground truth masks for each object in each frame of the video. The mAP score is then calculated by averaging the AP scores across all the object categories in all the frames of the video.

\noindent
\textbf{Panoptic Quality (PQ).} This is a default evaluation metric for panoptic segmentation. PQ is computed by comparing the predicted segmentation masks with the ground truth masks for both objects and stuff in an image. In particular, the PQ is measured at segment level. Given a semantic class $c$, by matching predictions $p$ to ground-truth segments $g$ based on the IoU scores, these segments can be divided into true positives (TP), false positives (FP), and false negatives (FN). Only a threshold of greater than 0.5 IoU is chosen to guarantee the unique matching. Then, the PQ is calculated as follows:

\begin{equation}
    \textrm{PQ}_{c} = \frac{\sum_{(p,g) \in \mathrm{TP}}\textrm{IoU}(p,g)}{|\mathrm{TP}| + \frac{1}{2}|\mathrm{FP}|+ \frac{1}{2}|\mathrm{FN}|},
 \label{equ:metric_pq}
\end{equation}

The final $\mathrm{PQ}$ is obtained by the average results of all classes as defined in Equ.~\ref{equ:metric_pq}.

\noindent
\textbf{Video Panoptic Quality (VPQ).} This metric extends the PQ into video by calculating the spatial-temporal mask IoU along different temporal window sizes. It is designed for VPS tasks~\cite{kim2020vps}. When the temporal window size is 1, the VPQ is the same as the PQ. Following PQ, VPQ performs matching by a threshold of greater than 0.5 IoU for all temporal segment predictions. In particular, for the thing classes, only tracked objects with the same semantic classes are considered as TP. There are only two differences: one for temporal mask prediction and the other for the different temporal window sizes. For the former, any cross-frame inconsistency of semantic or instance label prediction will result in a low tube IoU, and may drop the match out of the TP set. For the latter, short window sizes measure the consistency for short clip inputs, while long window sizes focus on long clip inputs. The final VPQ is obtained by averaging the results of different window sizes.

\noindent
\textbf{Segmentation and Tracking Quality (STQ).} Both PQ and VPQ have different thresholds and extra parameters for VPS. To solve this and give a decoupled view for segmentation and tracking, STQ~\cite{STEP} is proposed to evaluate the entire video clip at the pixel level. STQ combines association quality (AQ) and segmentation quality (SQ), which measure the tracking and segmentation quality, respectively. The proposed AQ is designed to work at the pixel level of a full video. All correct and incorrect associations influence the score, independent of whether a segment is above or below an IoU threshold. Motivated by HOTA~\cite{luiten2021hota}, AQ is calculated by jointly considering the localization and association ability for each pixel. SQ has the same definition as mIoU. The overall STQ score is the geometric mean of AQ and SQ.

\noindent
\textbf{Region Similarity, J.} This is used for region-based segmentation
similarity for VOS. The previous VOS methods adopt the Jaccard index, $J$ defined as the intersection over the union of estimated segmentation and the ground truth mask.

\noindent
\textbf{Contour Accuracy, F.} This is used for contour-based segmentation
similarity for VOS. Previous works adopt the F-measure between the contour points from both the predicted segmentation mask and the ground truth mask.

\subsection{Representative Works in Specific Subfields}
\label{sec:representativework_in_related_domain_beyond}
Due to the limited space of the main paper, in Tab.~\ref{tab:related_domain_beyond_summary}, we list several representative works to augment the Specific Subfields section. Moreover, we also list the detailed improvements for different techniques to improve K-Net using COCO-panoptic datasets, and we adopt the default settings for re-benchmarking.

\subsection{More Benchmark Results}
\label{sec:more_benchmark_result}

\noindent$\bullet$
\noindent
\textbf{Results on Point Cloud Segmentation.} In Tab.~\ref{tab:point_cloud_seg_bench}, we list the transformer-based point cloud segmentation methods on ScanNet and S3DIS datasets. The OneFormer3D~\cite{kolodiazhnyi2023oneformer3d} achieves the best results on three subtasks.

\begin{table}[htbp]
    \centering
    \caption{\textbf{Benchmark results on point cloud segmentation.}}
    \vspace{-2mm}
    \label{tab:point_cloud_seg_bench}
    \begin{adjustbox}{width=0.50\textwidth}
    \begin{tabular}{c c c c c}
    \toprule[0.15em]
    \textbf{Method} & \textbf{Dataset} & mIoU~(sem.) & AP50~(ins.)  & PQ~(pan.)  \\ 
    \midrule[0.15em]
    PointTransformer~\cite{zhao2021point}& \multirow{5}{*}{ScanNet} & 70.6 & - & - \\
    PointTransformerV2~\cite{wu2022point}& & 75.4 & - & - \\
    Mask3D~\cite{schult2023mask3d} &  & - & 73.7 & - \\
    SPFormer~\cite{sun2023superpoint}& & - & 73.9 & - \\
    OneFormer3D~\cite{kolodiazhnyi2023oneformer3d}& & 76.6 & 78.1 & 71.2 \\
    \midrule
    PointTransformer~\cite{zhao2021point}& \multirow{5}{*}{S3DIS} & 70.4 & - & - \\
    PointTransformerV2~\cite{wu2022point}& & 71.6 & - & - \\
    Mask3D~\cite{schult2023mask3d} &  & - & 71.9 & - \\
    SPFormer~\cite{sun2023superpoint}& & - & 66.8 & - \\
    OneFormer3D~\cite{kolodiazhnyi2023oneformer3d}& & 72.4 & 72.0 & 62.2 \\
    \bottomrule
    \end{tabular}
    \end{adjustbox}
\end{table}

\noindent$\bullet$
\noindent
\textbf{Results on Open Vocabulary Semantic Segmentation.} In Tab.~\ref{table:results-ovss-self}, we report methods on open vocabulary semantic segmentation in the self-evaluation setting. The self-evaluation setting splits the classes into base classes for training and treats the novel classes as background. It uses both base and novel classes for testing. From that table, FreeSeg~\cite{qin2023freeseg} achieves the best results. 
In Tab.~\ref{table:results-ovss-cross}, we list several representative works under the cross-evaluation setting. Since different methods adopt different datasets and supervision types for pre-training and co-training, we also show the detailed extra data for reference. Among these methods, X-Decoder~\cite{zou2022xdecoder} achieves the best results. We refer the reader to the work~\cite{wu2023open} for a more comprehensive comparison.

\begin{table*}[t!]
\centering
\caption{\textbf{Open vocabulary semantic segmentation performances} under \textbf{the self-evaluation setting.} The metric is mIoU. ``Harmonic'' means the harmonic mean of the mIoU for base classes and the mIoU for novel classes.}
  \scalebox{0.90}{
  \begin{tabular}{lcc|ccccccccc}
    \toprule[0.15em]
    \multirow{2}{*}{Method} & \multirow{2}{*}{Backbone} & \multirow{2}{*}{VLM} & \multicolumn{3}{c}{COCO-Stuff} & \multicolumn{3}{c}{PASCAL-VOC} & \multicolumn{3}{c}{PASCAL-Context} \\
    & & & Base & Novel & Harmonic & Base & Novel & Harmonic & Base & Novel & Harmonic \\
    \midrule
    ZegFormer~\cite{ding2022decoupling} (\textit{CVPR'22}) & ResNet101 & CLIP-ViT-B/16 & 36.6 & 33.2 & 34.8 & 86.4 & 63.6 & 73.3 & - & - & - \\
    Xu et al.~\cite{xu2022_simple_baseline_open_voc} (\textit{ECCV'22}) & ResNet101 & CLIP-ViT-B/16 & 39.6 & 43.6 & 41.5 & 79.2 & 78.1 & 79.3 & - & - & - \\
    PADing~\cite{PAD} (\textit{CVPR'23}) & ResNet101 & CLIP-ViT-B/16 & 39.9   & 44.9 & 42.2 & - & - & - & - & - & - \\
    MaskCLIP+~\cite{zhou2021denseclip} (\textit{ECCV'22}) & ResNet101 & CLIP-ResNet50 & 39.6 & {54.7} & 45.0 & 88.1 & {86.1} & {87.4} & {48.1} & {66.7} & {53.3} \\
    FreeSeg~\cite{qin2023freeseg} (\textit{CVPR'23}) & ResNet101 & CLIP-ViT-B/16 & {42.2} & 49.1 & {45.3} & {91.8} & 82.6 & 86.9 & - & - & - \\
    \bottomrule[0.15em]
  \end{tabular}}
\label{table:results-ovss-self}
\end{table*}

\begin{table*}[t!]
\centering
\caption{\textbf{Open vocabulary semantic segmentation performances} under \textbf{the cross-evaluation setting.} The metric is mIoU. For COCO, different methods use different supervision to train their models, including mask, classification (cls), and caption (cap). PAS-20$^b$ is an evaluation setting using the PASCAL-VOC dataset proposed by OpenSeg~\cite{OpenSeg}. It only assigns the background class in PASCAL-VOC to the pixels having a PC-59 category, which is harder than PAS-20 in general. ``*'' means the result is only tested on the novel classes in the class split. A-150 means ADE20k~\cite{ADE20K} with 150 classes.}
  \scalebox{0.79}{
  \begin{tabular}{lcc|cccc|cccccc}
    \toprule[0.15em]
    \multirow{2}{*}{Method} & \multirow{2}{*}{Backbone} & \multirow{2}{*}{VLM} & \multicolumn{3}{c}{COCO} & \multirow{2}{*}{Extra data} & \multirow{2}{*}{A-847} & \multirow{2}{*}{PC-459} & \multirow{2}{*}{A-150} & \multirow{2}{*}{PC-59} & \multirow{2}{*}{PAS-20$^b$} & \multirow{2}{*}{PAS-20} \\
    & & & mask & cls & cap & & & & & & & \\
    
    \midrule

    MaskCLIP~\cite{panoptic-MaskCLIP} (\textit{ICML'22}) & ResNet50 & CLIP-ViT-L/14 & \cmark & \cmark & \xmark & \textit{None} & 8.2 & 10.0 & 23.7 & 45.9 & - & - \\
    
    \midrule
    ZegFormer~\cite{ding2022decoupling} (\textit{CVPR'22}) & RN-101 & CLIP-ViT-B/16 & \cmark & \cmark & \xmark & \textit{None} & 5.6 & 10.4 & 18.0 & 45.5 & 65.5 & 89.5 \\
    LSeg+~\cite{OpenSeg} (\textit{ECCV'22}) & RN-101 & ALIGN & \cmark & \cmark & \xmark & \textit{None} & 2.5 & 5.2 & 13.0 & 36.0 & 59.0 & - \\
    OpenSeg~\cite{OpenSeg} (\textit{ECCV'22}) & RN-101 & ALIGN & \cmark & \xmark & \cmark & Localized Narrative & 4.4 & 7.9 & 17.5 & 40.1 & 63.8 & - \\
    Xu et al.~\cite{xu2022_simple_baseline_open_voc} (\textit{ECCV'22}) & RN-101 & CLIP-ViT-B/16 & \cmark & \cmark & \xmark & \textit{None} & 7.0 & - & 20.5 & 47.7 & & 88.4 \\
    OVSeg~\cite{Mask-adapted-clip} (\textit{CVPR'23}) & RN-101c & CLIP-ViT-B/16 & \cmark & \cmark & \cmark & \textit{None} & 7.1 & 11.0 & 24.8 & 53.3 & - & 92.6 \\
    Han et al.~\cite{global-knowledge-calibration} (\textit{ICCV'23}) & RN-101 & CLIP-ResNet50 & \cmark & \cmark & \cmark & \textit{None} & 3.5 & 7.1 & 18.8 & 45.2 & 83.2 & - \\
    \midrule

    OpenSeg~\cite{OpenSeg} (\textit{ECCV'22}) & Eff-B7 & ALIGN & \cmark & \xmark & \cmark & Localized Narrative & 8.1 & 11.5 & 26.4 & 44.8 & 70.2 & - \\
    ODISE~\cite{xu2023odise} (\textit{CVPR'23}) & Stable diffusion & CLIP-ViT-L/14 & \cmark & \cmark & \xmark & \textit{None} & 11.1 & 14.5 & 29.9 & 57.3 & {84.6} & - \\
    OVSeg~\cite{Mask-adapted-clip} (\textit{CVPR'23}) & Swin-B & CLIP-ViT-L/14 & \cmark & \cmark & \cmark & \textit{None} & 9.0 & 12.4 & 29.6 & 55.7 & - & 94.5 \\
    SAN~\cite{side-adapter} (\textit{CVPR'23}) & ViT-L/14 & CLIP-ViT-L/14 & \cmark & \cmark & \xmark & \textit{None} & 12.4 & 15.7 & {32.1} & 57.7 & - & 94.6 \\
    \midrule

    X-Decoder~\cite{zou2022xdecoder} (\textit{CVPR'23}) & DaViT-L & \textit{None} & \cmark & \cmark & \cmark & \cite{conceptual-captions, SBU, visual-genome, coco-captions} & 9.2 & 16.1 & 29.6 & {64.0} & - & {97.7} \\
    OpenSeed~\cite{zhang2023simple} (\textit{ICCV'23}) & Swin-L & \textit{None} & \cmark & \cmark & \xmark & Objects365~\cite{objects365} & - & - & 23.4 & - & - & - \\
    HIPIE~\cite{HIPIE} (\textit{NeurIPS'23}) & ViT-H & \textit{None} & \cmark & \cmark & \xmark & Objects365~\cite{objects365} & - & - & 29.0 & - & - & - \\

    \bottomrule[0.15em]
  \end{tabular}}
\label{table:results-ovss-cross}
\end{table*}

\begin{table*}[t!]
   \centering
    \caption{\textbf{Open vocabulary instance segmentation} performances on the COCO-Instances dataset.} \vspace{-2mm}
   \scalebox{0.9}{
   \begin{tabular}{lcc|cccc|cc|ccc}
      \toprule[0.15em]
      \multirow{2}{*}{Method} & \multirow{2}{*}{Backbone} & \multirow{2}{*}{VLM} & \multicolumn{3}{c}{COCO} & \multirow{2}{*}{Extra data} & \multicolumn{2}{c|}{Constrained}  & \multicolumn{3}{c}{Generalized} \\
        & & & mask & cls & cap & & $AP_{base}$ & $AP_{novel}$ & $AP_{base}$ & $AP_{novel}$ & $AP_{all}$ \\
        \hline
        D$^2$Zero~\cite{D2Zero} (\textit{CVPR'23}) & ResNet50 & CLIP-ViT-B/16 & \cmark & \cmark & \xmark & \textit{None} & - & 23.7 & 40.9 & 21.9 & -\\
        XPM~\cite{XPM} (\textit{CVPR'22}) & ResNet50 & \textit{None} & \cmark & \cmark & \cmark & CC3M & 42.4 & 24.0 & 41.5 & 21.6 & 36.3 \\
        Mask-free OVIS~\cite{mask-free-OVIS} (\textit{CVPR'23}) & ResNet50 & ALBEF & \xmark & \xmark & \cmark & \textit{None} & - & 27.4 & - & 25.0 & - \\
        CGG~\cite{wu2023betrayed} (\textit{ICCV'23}) & ResNet50 & \textit{None} & \cmark & \cmark & \cmark & \textit{None} & {46.8} & {29.5} & {46.0} & {28.4} & {41.4} \\
      \bottomrule[0.10em]
   \end{tabular}}
   \label{tab:result_OVIS}
\end{table*}

\begin{table*}[t!]
   \centering
    \caption{\textbf{Open vocabulary panoptic segmentation} performances on the COCO dataset. No methods use extra data for training.}
   \scalebox{1.0}{
   \begin{tabular}{lcc|ccc|cccccc}
      \toprule[0.15em]
      \multirow{2}{*}{Method} & \multirow{2}{*}{Backbone} & \multirow{2}{*}{VLM} & \multicolumn{3}{c|}{COCO} & \multirow{2}{*}{PQ$^{s}$} & \multirow{2}{*}{SQ$^{s}$} & \multirow{2}{*}{RQ$^{s}$} & \multirow{2}{*}{PQ$^{u}$} & \multirow{2}{*}{SQ$^{u}$} & \multirow{2}{*}{RQ$^{u}$}\\
        & & & mask & cls & cap & & & & & & \\
        \hline
        ZSSeg~\cite{xu2022simple} (\textit{ECCV'22}) & ResNe50 &  CLIP-ViT-B/16 & \cmark & \cmark & \xmark & 27.2  & 76.1 & 34.7 & 9.7  & 71.7  & 12.2 \\
        PADing~\cite{PAD} (\textit{CVPR'23}) & ResNet50 & CLIP-ViT-B/16 & \cmark & \cmark & \xmark & {41.5} & {80.6} & {49.7} & 15.3 &72.8 &18.4 \\
        FreeSeg~\cite{qin2023freeseg} (\textit{CVPR'23}) & ResNet101 & CLIP-ViT-B/16 & \cmark & \cmark & \xmark & 31.4 & 78.3 & 38.9 & {29.8} & {79.2} & {37.6} \\
      \bottomrule[0.10em]
   \end{tabular}}
   \label{tab:result_OVPS_coco}
\end{table*}

\noindent$\bullet$
\noindent
\textbf{Results on Open Vocabulary Instance Segmentation.} In Tab.~\ref{tab:result_OVIS}, we report open vocabulary instance segmentation on COCO datasets. Among these methods, CGG~\cite{wu2023betrayed} achieves the best performance.

\noindent$\bullet$
\noindent
\textbf{Results on Open Vocabulary Panoptic Segmentation.} In Tab.~\ref{tab:result_OVPS_coco}, we report open vocabulary panoptic segmentation on COCO datasets. Among these methods, FreeSeg~\cite{qin2023freeseg} achieves the best performance.

\begin{table}[htbp]
    \centering
    \caption{\textbf{Benchmark on weakly supervised semantic segmentation.}}
    \vspace{-2mm}
    \label{tab:wsss_bench}
    \begin{adjustbox}{width=0.50\textwidth}
    \begin{tabular}{c c c c c c}
    \toprule[0.15em]
    \textbf{Method} & \textbf{Backbone} & VOC (val) & VOC (test) & COCO (val) \\
    \midrule
    AFA~\cite{ru2022weakly} & MiT-B1 & 66.0 &66.3 &38.9\\
    MCTformer~\cite{xu2022multitoke_wsss} & WR38 & 71.9 & 71.6 &42.0\\
    MCTformer+~\cite{xu2023mctformer_p} &WR38 & 74.0  & 73.6 & 45.2\\
    ToCo~\cite{ru2023token} & ViT-B & 71.1 & 72.2 & 42.3\\
    WeakTr~\cite{zhu2023weaktr} & ViT-S & 78.4 & 79.0 &50.3\\
    \bottomrule
    \end{tabular}
    \end{adjustbox}
\end{table}

\begin{table}[htbp]
    \centering
    \caption{\textbf{Benchmark on unsupervised semantic segmentation.}}
    \vspace{-2mm}
    \label{tab:unsup_ss_bench}
    \begin{adjustbox}{width=0.50\textwidth}
    \begin{tabular}{c c c c c c}
    \toprule[0.15em]
    \multirow{2}{*}{\textbf{Method}}& \multirow{2}{*}{\textbf{Backbone}} & \multicolumn{2}{c}{\textbf{Unsupervised}} & \multicolumn{2}{c}{\textbf{Linear Probing}} \\
    & & Accuracy & mIoU &  Accuracy & mIoU\\
    \midrule
    DINO~\cite{caron2021emergingDINO} & ViT-S/8 & 28.7 & 11.3 & 66.8 & 29.4\\
    STEGO~\cite{hamilton2022unsupervisedSETGO} & ViT-S/8 & 48.3 & 24.5 & 74.4 &38.3\\
    CLAUSE-TR~\cite{kim2023causal} & ViT-S/8 & 69.6 & 32.4 & 78.8 & 47.2 \\
    \midrule
    DINO~\cite{caron2021emergingDINO} & ViT-B/8 & 42.4 & 13.0 & 66.8 & 29.4\\
    STEGO~\cite{hamilton2022unsupervisedSETGO} & ViT-B/8 &56.9& 28.2& 76.1& 41.0\\
    CLAUSE-TR~\cite{kim2023causal} & ViT-B/8 & 74.9 & 41.9 & 80.1 & 52.3 \\
    \bottomrule
    \end{tabular}
    \end{adjustbox}
\end{table}

\noindent$\bullet$
\noindent
\textbf{Results on Weakly-Supervised Semantic Segmentation.} In Tab.~\ref{tab:wsss_bench}, we list the transformer-based weakly-supervised semantic segmentation methods. The WeakTr~\cite{zhu2023weaktr} achieves the best performance on VOC and COCO datasets.

\noindent$\bullet$
\noindent
\textbf{Results on Unsupervised Semantic Segmentation.} We also list several unsupervised semantic segmentation methods in Tab.~\ref{tab:unsup_ss_bench} using ViT-S and ViT-B as backbone. The CLAUSE~\cite{kim2023causal} achieves the best results on both regular unsupervised setting and linear probing setting.


\begin{table}[t]
\centering
\caption{\textbf{Ablation Study on Improved Techniques Used in Re-benchmarking using K-Net for Instance Segmentation and Panoptic Segmentation on COCO.} We adopt ResNet50 as backbone.}\vspace{-2mm}
\label{tab:ablation_tech_COCO}
\begin{adjustbox}{width=0.45\textwidth}
\begin{tabular}{c c c c c c}
\toprule[0.15em]
{\scriptsize \textbf{Method}}  & PQ & mAP & epoch  \\ 
\midrule[0.15em]
K-Net~\cite{zhang2021knet} & 47.1 & 39.1 & 36\\
\hline
+ lsj augmentation during training~\cite{ghiasi2021simple} & 47.5 & 40.2 & 36 \\
+ deformable FPN~\cite{zhu2020deformabledetr} & 47.9 & 42.1 &  36 \\
+ both & 48.5 & 42.5 & 36\\
+ both (K-Net++, final setting by default) & 49.2  & 43.9 & 50 \\
\bottomrule
\end{tabular}
\end{adjustbox}
\end{table}

\begin{table}[t]
\centering
\caption{\textbf{Ablation Study on Improved Techniques Used in Re-benchmarking using Mask2Former For Semantic Segmentation on ADE-20k.} We adopt ResNet50 as backbone and training epoch is set to 120.}\vspace{-2mm}
\label{tab:ablation_tech_ade20k}
\begin{adjustbox}{width=0.45\textwidth}
\begin{tabular}{c c c c c c}
\toprule[0.15em]
{\scriptsize \textbf{Method}}  &  mIoU  \\ 
\midrule[0.15em]
Mask2Former~\cite{cheng2021mask2former} &  47.0  \\
\hline
+ lsj augmentation during training~\cite{ghiasi2021simple} &  47.5 \\
w/o deformable FPN~\cite{zhu2020deformabledetr} & 44.2 \\
\bottomrule
\end{tabular}
\end{adjustbox}
\end{table}

\begin{table}[h]
\centering
\caption{\textbf{Ablation Study on Improved Techniques Used in Re-benchmarking using Mask2Former For Semantic Segmentation on VSPW.} We adopt ResNet50 as backbone.}\vspace{-2mm}
\label{tab:ablation_tech_VSPW}
\begin{adjustbox}{width=0.45\textwidth}
\begin{tabular}{c c c c c c}
\toprule[0.15em]
{\scriptsize \textbf{Method}}  &  mIoU  \\ 
\midrule[0.15em]
Mask2Former~\cite{cheng2021mask2former} & 39.6  \\
\hline
+ lsj augmentation during training~\cite{ghiasi2021simple} & 39.5 \\
w/o deformable FPN~\cite{zhu2020deformabledetr} & 37.5 \\
\bottomrule
\end{tabular}
\end{adjustbox}
\end{table}

\begin{table}[h]
\centering
\caption{\textbf{Ablation Study on Improved Techniques Used in Re-benchmarking using Mask2Former-VIS For Video Instance Segmentation on Youtube-VIS-2019.}}\vspace{-2mm}
\label{tab:ablation_tech_youtube_vis}
\begin{adjustbox}{width=0.45\textwidth}
\begin{tabular}{c c c c c c}
\toprule[0.15em]
{\scriptsize \textbf{Method}}  &  mAP  \\ 
\midrule[0.15em]
Mask2Former-VIS~\cite{cheng2021mask2formervis} &  46.4 \\
\hline
+ lsj augmentation during training~\cite{ghiasi2021simple} & 46.2 \\
w/o deformable FPN~\cite{zhu2020deformabledetr} & 42.3 \\
\bottomrule
\end{tabular}
\end{adjustbox}
\end{table}

\subsection{Details of Benchmark Experiment}
\label{sec:re_benchmark_details}

\noindent
\textbf{Implementation Details on Semantic Segmentation Benchmarks.} We adopt the MMSegmentation~\cite{mmseg2020} codebase with the same setting to carry out the re-benchmarking experiments for semantic segmentation. In particular, we train the models with the AdamW optimizer for 160K iterations on ADE20K, Cityscapes, and 80K iterations on COCO-Stuff. For Cityscapes, we adopt a random crop size of 1024. For the other datasets, we adopt the crop size of 512.

\noindent
\textbf{Implementation Details on Instance and Panoptic Segmentation Benchmarks.} We adopt the MMDetection~\cite{chen2019mmdetection} codebase with the same setting to carry out the re-benchmarking experiments for panoptic segmentation. We strictly follow the Mask2Former settings for all models and all datasets. In particular, a learning rate multiplier of 0.1 is applied to the backbone. For data augmentation, we use the default large-scale jittering augmentation with a random scale sampled from the range 0.1 to 2.0 with the crop size of 1024 $\times$ 1024 in the COCO dataset. For the ADE20k dataset, the crop size is set to 640. The training iteration is set to 160k. For Cityscapes, we set the crop size to $512 \times 1024$ with 90k training iterations. We refer the readers to our code for more details.

\noindent
\textbf{Detailed ablation studies on improved technical tricks.} We present more ablation studies on various baselines in more segmentation tasks, including semantic segmentation, instance segmentation, video semantic segmentation, and video instance segmentation. 

From Tab.~\ref{tab:ablation_tech_COCO}, we find both techniques work effectively on COCO datasets, whether on instance segmentation or panoptic segmentation. We use the K-Net~\cite{zhang2021knet} trained on COCO-panoptic dataset for reference. Adding lsj augmentation~\cite{ghiasi2021simple} can improve the performance by 0.5\%. Adding deformable FPN~\cite{zhu2020deformabledetr} can lead to  1.0\% gains. Adding both leads to better results. Finally, we follow the default Mask2Former design by extending the training epoch to 50 for re-benchmarking, which leads to the best performance. 
These results indicate both stronger augmentation and stronger feature pyramid networks are important in the COCO dataset. From Tab.~\ref{tab:ablation_tech_ade20k}, we carry out the semantic segmentation experiment on the ADE-20k dataset using Mask2Former as a baseline. As shown in that table, both techniques still work effectively. 

For video segmentation, we explore video semantic segmentation and video instance segmentation in Tab.~\ref{tab:ablation_tech_VSPW} and Tab.~\ref{tab:ablation_tech_youtube_vis}. However, we find adding the lsj augmentation does not bring extra gains. This is because both diversity and small objects in VSPW and Youtube-VIS-2019 are less than in the COCO dataset. From both tables, we find the deformable FPN works nearly for all different video segmentation tasks while the lsj augmentation works less effectively.

In conclusion, we find that adding deformable FPN works in all settings since it leads to stronger multi-scale feature representation, while the LSJ training is more effective in complex scenes such as COCO datasets.

\subsection{More Future Directions}
\label{sec:more_future_directions}

In this section, we present more discussion on potential future directions, including mobile segmentation, using synthetic datasets for joint training, efficient modeling in segmentation, domain generation, and 4D point cloud segmentation. 

\noindent$\bullet$
\textbf{Mobile Segmentation.} Several works~\cite{zhang2022topformer,dong2023afformer} design mobile segmentation methods using Transformer. However, these designs mainly focus on the semantic level with pure image inputs. With the rise of short videos in mobile applications, instance-level mobile video segmentation may need more research efforts. Thus, efficiently segmenting and tracking each instance on mobile devices in the video clip may require more research for potential applications.

\noindent$\bullet$
\textbf{Using Synthetic Datasets for Joint Training.}
Segmentation models always need huge pixel-wised annotations. While methods like collecting image-text pairs~\cite{jia2021scaling_align} and using open vocabulary approaches~\cite{detic} can reduce annotation costs, they still require significant manual effort in data collection. One alternative solution is to use a generated synthetic dataset. Recently, diffusion-based generation models~\cite{rombach2021highresolution} have emerged as a promising option for high-quality image and mask generation. These models can create synthetic images and masks with fewer domain gaps than natural images, making them ideal for training segmentation models without requiring access to real data.
Additionally, synthetic datasets can be tailored to specific application scenarios, such as few-shot segmentation tasks or long-tail segmentation tasks. Using synthetic datasets for joint training has the potential to significantly reduce annotation costs and accelerate the development of segmentation models for real-world applications.

\noindent$\bullet$
\textbf{Weakly Supervised or Unsupervised Segmentation.} Most segmentation approaches need lots of mask annotations for training, which leads to huge manual annotation costs. Thus, developing annotation-efficient segmentation algorithms is needed. Weakly supervised annotations, including image labels or boxes, can be used to replace fine-grained mask annotations. Recent work~\cite{lan2023vision} shows the transformer itself can learn a good mask classifier with only box supervision.
This finding makes the training instance segmentation model easier. Moreover, with the recent progress of contrastive pre-training, exploring the vision transformer itself as a mask generator for unsupervised segmentation is also a promising direction.

\noindent$\bullet$
\textbf{Domain Generation.} This task aims to adapt a segmenter from the seen domains to the new unseen domains without re-training or accessing new unseen images during the training. Previous works~\cite{zhao2022shade,choi2021robustnet} adopt specific designs, including data augmentation, feature distillation, and feature whitening. Only a few works~\cite{zhao2022shadevdg,zhao2022shade} explore the effectiveness of segmentation transformer in cases of data augmentation for domain generation. However, there are no works using stronger foundation models to build stronger baselines. Moreover, current works only consider similar domains in the driving scene, which are not generalizable in real applications. A robot should adapt themselves to various scene inputs, including outdoor scenes and indoor scenes. Thus, methods that analyze the generation ability in various domains are needed in the future.

\noindent$\bullet$
\textbf{4D Point Cloud Panoptic Segmentation.} This task requires the model to segment and track each point in the video. Current methods~\cite{hong20224ddsnet,Hong_2021_CVPR,aygun20214d,Zhu_2021_CVPR} for 4D point cloud panoptic segmentation usually adopt specific pipelines, including point segmentation, point clustering, and point association. A unified solution using a transformer may facilitate this direction by simplifying the pipeline and performing task association.

\ifCLASSOPTIONcaptionsoff
  \newpage
\fi



{
\bibliographystyle{IEEEtran}
\bibliography{IEEEabrv,egbib}

\begin{thebibliography}{100}
\providecommand{\url}[1]{#1}
\csname url@samestyle\endcsname
\providecommand{\newblock}{\relax}
\providecommand{\bibinfo}[2]{#2}
\providecommand{\BIBentrySTDinterwordspacing}{\spaceskip=0pt\relax}
\providecommand{\BIBentryALTinterwordstretchfactor}{4}
\providecommand{\BIBentryALTinterwordspacing}{\spaceskip=\fontdimen2\font plus
\BIBentryALTinterwordstretchfactor\fontdimen3\font minus
  \fontdimen4\font\relax}
\providecommand{\BIBforeignlanguage}[2]{{%
\expandafter\ifx\csname l@#1\endcsname\relax
\typeout{** WARNING: IEEEtran.bst: No hyphenation pattern has been}%
\typeout{** loaded for the language `#1'. Using the pattern for}%
\typeout{** the default language instead.}%
\else
\language=\csname l@#1\endcsname
\fi
#2}}
\providecommand{\BIBdecl}{\relax}
\BIBdecl

\bibitem{malik2001contour}
J.~Malik, S.~Belongie, T.~Leung, and J.~Shi, ``Contour and texture analysis for
  image segmentation,'' \emph{IJCV}, 2001.

\bibitem{shi2000normalized}
J.~Shi and J.~Malik, ``Normalized cuts and image segmentation,'' \emph{TPAMI},
  2000.

\bibitem{stella2003multiclass}
X.~Y. Stella and J.~Shi, ``Multiclass spectral clustering,'' in \emph{ICCV},
  2003.

\bibitem{schroff2008object}
F.~Schroff, A.~Criminisi, and A.~Zisserman, ``Object class segmentation using
  random forests.'' in \emph{BMVC}, 2008.

\bibitem{kass1988snakes}
M.~Kass, A.~Witkin, and D.~Terzopoulos, ``Snakes: Active contour models,''
  \emph{IJCV}, 1988.

\bibitem{russakovsky2015imagenet}
O.~Russakovsky, J.~Deng, H.~Su, J.~Krause, S.~Satheesh, S.~Ma, Z.~Huang,
  A.~Karpathy, A.~Khosla, M.~Bernstein \emph{et~al.}, ``Imagenet large scale
  visual recognition challenge,'' \emph{IJCV}, 2015.

\bibitem{resnet}
K.~{He}, X.~{Zhang}, S.~{Ren}, and J.~{Sun}, ``Deep residual learning for image
  recognition,'' in \emph{CVPR}, 2016.

\bibitem{simonyan2014very}
K.~Simonyan and A.~Zisserman, ``Very deep convolutional networks for
  large-scale image recognition,'' in \emph{ICLR}, 2015.

\bibitem{long2015fully}
J.~Long, E.~Shelhamer, and T.~Darrell, ``Fully convolutional networks for
  semantic segmentation,'' in \emph{CVPR}, 2015.

\bibitem{chen2017deeplab}
L.-C. Chen, G.~Papandreou, I.~Kokkinos, K.~Murphy, and A.~L. Yuille,
  ``{DeepLab}: Semantic image segmentation with deep convolutional nets, atrous
  convolution, and fully connected {CRFs},'' \emph{IEEE TPAMI}, vol.~40, no.~4,
  pp. 834--848, 2017.

\bibitem{zhao2017pyramid}
H.~Zhao, J.~Shi, X.~Qi, X.~Wang, and J.~Jia, ``Pyramid scene parsing network,''
  in \emph{CVPR}, 2017.

\bibitem{ding2018context}
H.~Ding, X.~Jiang, B.~Shuai, A.~Q. Liu, and G.~Wang, ``Context contrasted
  feature and gated multi-scale aggregation for scene segmentation,'' in
  \emph{CVPR}, 2018.

\bibitem{vaswani2017attention}
A.~Vaswani, N.~Shazeer, N.~Parmar, J.~Uszkoreit, L.~Jones, A.~N. Gomez,
  L.~Kaiser, and I.~Polosukhin, ``Attention is all you need,'' in \emph{NIPS},
  2017.

\bibitem{LSTM}
S.~Hochreiter and J.~Schmidhuber, ``Long short-term memory,'' \emph{Neural
  computation}, 1997.

\bibitem{BERT}
J.~Devlin, M.-W. Chang, K.~Lee, and K.~Toutanova, ``{BERT}: Pre-training of
  deep bidirectional transformers for language understanding,'' in
  \emph{NAACL}, 2019.

\bibitem{GPT3}
T.~Brown, B.~Mann, N.~Ryder, M.~Subbiah, J.~D. Kaplan, P.~Dhariwal,
  A.~Neelakantan, P.~Shyam, G.~Sastry, A.~Askell \emph{et~al.}, ``Language
  models are few-shot learners,'' in \emph{NeurIPS}, 2020.

\bibitem{zhao2018psanet}
H.~Zhao, Y.~Zhang, S.~Liu, J.~Shi, C.~Change~Loy, D.~Lin, and J.~Jia,
  ``{PSANet}: Point-wise spatial attention network for scene parsing,'' in
  \emph{ECCV}, 2018.

\bibitem{wang2018nonlocal}
X.~Wang, R.~Girshick, A.~Gupta, and K.~He, ``Non-local neural networks,'' in
  \emph{CVPR}, 2018.

\bibitem{zhao2020exploring}
H.~Zhao, J.~Jia, and V.~Koltun, ``Exploring self-attention for image
  recognition,'' in \emph{CVPR}, 2020.

\bibitem{hu2019local}
H.~Hu, Z.~Zhang, Z.~Xie, and S.~Lin, ``Local relation networks for image
  recognition,'' in \emph{ICCV}, 2019.

\bibitem{VIT}
A.~Dosovitskiy, L.~Beyer, A.~Kolesnikov, D.~Weissenborn, X.~Zhai,
  T.~Unterthiner, M.~Dehghani, M.~Minderer, G.~Heigold, S.~Gelly, J.~Uszkoreit,
  and N.~Houlsby, ``An image is worth 16x16 words: Transformers for image
  recognition at scale,'' in \emph{ICLR}, 2021.

\bibitem{detr}
N.~Carion, F.~Massa, G.~Synnaeve, N.~Usunier, A.~Kirillov, and S.~Zagoruyko,
  ``End-to-end object detection with transformers,'' in \emph{ECCV}, 2020.

\bibitem{liu2021swin}
Z.~Liu, Y.~Lin, Y.~Cao, H.~Hu, Y.~Wei, Z.~Zhang, S.~Lin, and B.~Guo, ``Swin
  transformer: Hierarchical vision transformer using shifted windows,''
  \emph{ICCV}, 2021.

\bibitem{MaskedAutoencoders2021}
K.~He, X.~Chen, S.~Xie, Y.~Li, P.~Doll{\'a}r, and R.~Girshick, ``Masked
  autoencoders are scalable vision learners,'' \emph{CVPR}, 2022.

\bibitem{zhu2020deformabledetr}
X.~Zhu, W.~Su, L.~Lu, B.~Li, X.~Wang, and J.~Dai, ``Deformable {DETR}:
  Deformable transformers for end-to-end object detection,'' in \emph{ICLR},
  2021.

\bibitem{wang2020maxDeeplab}
H.~Wang, Y.~Zhu, H.~Adam, A.~Yuille, and L.-C. Chen, ``{MaX-DeepLab}:
  End-to-end panoptic segmentation with mask transformers,'' in \emph{CVPR},
  2021.

\bibitem{chen2021pre}
H.~Chen, Y.~Wang, T.~Guo, C.~Xu, Y.~Deng, Z.~Liu, S.~Ma, C.~Xu, C.~Xu, and
  W.~Gao, ``Pre-trained image processing transformer,'' in \emph{CVPR}, 2021,
  pp. 12\,299--12\,310.

\bibitem{bertasius2021space}
G.~Bertasius, H.~Wang, and L.~Torresani, ``Is space-time attention all you need
  for video understanding?'' in \emph{ICML}, 2021.

\bibitem{point_transformer}
H.~Zhao, L.~Jiang, J.~Jia, P.~H. Torr, and V.~Koltun, ``Point transformer,'' in
  \emph{ICCV}, 2021.

\bibitem{pan20213d}
X.~Pan, Z.~Xia, S.~Song, L.~E. Li, and G.~Huang, ``{3D} object detection with
  pointformer,'' in \emph{CVPR}, 2021, pp. 7463--7472.

\bibitem{han2022survey}
K.~Han, Y.~Wang, H.~Chen, X.~Chen, J.~Guo, Z.~Liu, Y.~Tang, A.~Xiao, C.~Xu,
  Y.~Xu \emph{et~al.}, ``A survey on vision transformer,'' \emph{TPAMI}, 2022.

\bibitem{khan2022transformers}
S.~Khan, M.~Naseer, M.~Hayat, S.~W. Zamir, F.~S. Khan, and M.~Shah,
  ``Transformers in vision: A survey,'' \emph{ACM computing surveys (CSUR)},
  2022.

\bibitem{lin2022survey}
T.~Lin, Y.~Wang, X.~Liu, and X.~Qiu, ``A survey of transformers,'' \emph{AI
  Open}, 2022.

\bibitem{selva2022videotransformer_survey}
J.~Selva, A.~S. Johansen, S.~Escalera, K.~Nasrollahi, T.~B. Moeslund, and
  A.~Clap{\'e}s, ``Video transformers: A survey,'' \emph{arXiv preprint
  arXiv:2201.05991}, 2022.

\bibitem{lahoud20223d_transformer_survey}
J.~Lahoud, J.~Cao, F.~S. Khan, H.~Cholakkal, R.~M. Anwer, S.~Khan, and M.-H.
  Yang, ``{3D} vision with transformers: A survey,'' \emph{arXiv preprint
  arXiv:2208.04309}, 2022.

\bibitem{xu2022multimodal}
P.~Xu, X.~Zhu, and D.~A. Clifton, ``Multimodal learning with transformers: A
  survey,'' \emph{arXiv preprint arXiv:2206.06488}, 2022.

\bibitem{minaee2021image}
S.~Minaee, Y.~Y. Boykov, F.~Porikli, A.~J. Plaza, N.~Kehtarnavaz, and
  D.~Terzopoulos, ``Image segmentation using deep learning: A survey,''
  \emph{TPAMI}, 2021.

\bibitem{hao2020brief}
S.~Hao, Y.~Zhou, and Y.~Guo, ``A brief survey on semantic segmentation with
  deep learning,'' \emph{Neurocomputing}, 2020.

\bibitem{zhou2023survey}
T.~Zhou, F.~Porikli, D.~J. Crandall, L.~Van~Gool, and W.~Wang, ``A survey on
  deep learning technique for video segmentation,'' \emph{TPAMI}, 2023.

\bibitem{miao2022large}
J.~Miao, X.~Wang, Y.~Wu, W.~Li, X.~Zhang, Y.~Wei, and Y.~Yang, ``Large-scale
  video panoptic segmentation in the wild: A benchmark,'' in \emph{CVPR}, 2022.

\bibitem{pascalvoc}
M.~{Everingham}, L.~J.~V. {Gool}, C.~K.~I. {Williams}, J.~M. {Winn}, and
  A.~{Zisserman}, ``The {PASCAL} visual object classes (voc) challenge,''
  \emph{IJCV}, 2010.

\bibitem{pascalcontext}
R.~Mottaghi, X.~Chen, X.~Liu, N.-G. Cho, S.-W. Lee, S.~Fidler, R.~Urtasun, and
  A.~Yuille, ``The role of context for object detection and semantic
  segmentation in the wild,'' in \emph{CVPR}, 2014.

\bibitem{coco_dataset}
T.-Y. Lin, M.~Maire, S.~Belongie, J.~Hays, P.~Perona, D.~Ramanan,
  P.~Doll{\'a}r, and C.~L. Zitnick, ``Microsoft {COCO}: Common objects in
  context,'' in \emph{ECCV}, 2014.

\bibitem{ADE20K}
B.~Zhou, H.~Zhao, X.~Puig, S.~Fidler, A.~Barriuso, and A.~Torralba, ``{Semantic
  understanding of scenes through the ADE20K dataset},'' \emph{CVPR}, 2017.

\bibitem{cordts2016cityscapes}
M.~Cordts, M.~Omran, S.~Ramos, T.~Rehfeld, M.~Enzweiler, R.~Benenson,
  U.~Franke, S.~Roth, and B.~Schiele, ``The cityscapes dataset for semantic
  urban scene understanding,'' in \emph{CVPR}, 2016.

\bibitem{neuhold2017mapillary}
G.~Neuhold, T.~Ollmann, S.~Rota~Bulo, and P.~Kontschieder, ``The mapillary
  vistas dataset for semantic understanding of street scenes,'' in \emph{ICCV},
  2017.

\bibitem{RefCOCO}
L.~Yu, P.~Poirson, S.~Yang, A.~C. Berg, and T.~L. Berg, ``Modeling context in
  referring expressions,'' in \emph{ECCV}, 2016.

\bibitem{GRES}
C.~Liu, H.~Ding, and X.~Jiang, ``{GRES}: Generalized referring expression
  segmentation,'' in \emph{CVPR}, 2023.

\bibitem{miao2021vspw}
J.~Miao, Y.~Wei, Y.~Wu, C.~Liang, G.~Li, and Y.~Yang, ``{VSPW}: A large-scale
  dataset for video scene parsing in the wild,'' in \emph{CVPR}, 2021.

\bibitem{vis_dataset}
L.~Yang, Y.~Fan, and N.~Xu, ``Video instance segmentation,'' in \emph{ICCV},
  2019.

\bibitem{OVIS}
J.~Qi, Y.~Gao, Y.~Hu, X.~Wang, X.~Liu, X.~Bai, S.~Belongie, A.~Yuille, P.~H.
  Torr, and S.~Bai, ``Occluded video instance segmentation: A benchmark,''
  \emph{IJCV}, vol. 130, no.~8, pp. 2022--2039, 2022.

\bibitem{kim2020vps}
D.~Kim, S.~Woo, J.-Y. Lee, and I.~S. Kweon, ``Video panoptic segmentation,'' in
  \emph{CVPR}, 2020.

\bibitem{STEP}
M.~Weber, J.~Xie, M.~Collins, Y.~Zhu, P.~Voigtlaender, H.~Adam, B.~Green,
  A.~Geiger, B.~Leibe, D.~Cremers, A.~Osep, L.~Leal-Taix{\'e}, and L.-C. Chen,
  ``{STEP}: Segmenting and tracking every pixel,'' \emph{NeurIPS}, 2021.

\bibitem{davis2017}
J.~Pont-Tuset, F.~Perazzi, S.~Caelles, P.~Arbel{\'a}ez, A.~Sorkine-Hornung, and
  L.~Van~Gool, ``The 2017 davis challenge on video object segmentation,''
  \emph{arXiv preprint arXiv:1704.00675}, 2017.

\bibitem{vos2018}
\BIBentryALTinterwordspacing
N.~Xu, L.~Yang, Y.~Fan, D.~Yue, Y.~Liang, J.~Yang, and T.~S. Huang,
  ``Youtube-vos: {A} large-scale video object segmentation benchmark,''
  \emph{CoRR}, vol. abs/1809.03327, 2018. [Online]. Available:
  \url{http://arxiv.org/abs/1809.03327}
\BIBentrySTDinterwordspacing

\bibitem{MOSE}
H.~Ding, C.~Liu, S.~He, X.~Jiang, P.~H. Torr, and S.~Bai, ``{MOSE}: A new
  dataset for video object segmentation in complex scenes,'' \emph{ICCV}, 2023.

\bibitem{MeViS}
H.~Ding, C.~Liu, S.~He, X.~Jiang, and C.~C. Loy, ``{MeViS}: A large-scale
  benchmark for video segmentation with motion expressions,'' in \emph{ICCV},
  2023.

\bibitem{yu2018learning}
C.~Yu, J.~Wang, C.~Peng, C.~Gao, G.~Yu, and N.~Sang, ``Learning a
  discriminative feature network for semantic segmentation,'' in \emph{CVPR},
  2018.

\bibitem{ding2020semantic}
H.~Ding, X.~Jiang, B.~Shuai, A.~Q. Liu, and G.~Wang, ``Semantic segmentation
  with context encoding and multi-path decoding,'' \emph{TIP}, 2020.

\bibitem{peng2017large}
C.~Peng, X.~Zhang, G.~Yu, G.~Luo, and J.~Sun, ``Large kernel matters--improve
  semantic segmentation by global convolutional network,'' in \emph{CVPR},
  2017.

\bibitem{SVCNet}
H.~Ding, X.~Jiang, B.~Shuai, A.~Q. Liu, and G.~Wang, ``Semantic correlation
  promoted shape-variant context for segmentation,'' in \emph{CVPR}, 2019.

\bibitem{deeplabv3}
L.-C. Chen, G.~Papandreou, F.~Schroff, and H.~Adam, ``Rethinking atrous
  convolution for semantic image segmentation,'' \emph{arXiv:1706.05587}, 2017.

\bibitem{shuai2018toward}
B.~Shuai, H.~Ding, T.~Liu, G.~Wang, and X.~Jiang, ``Toward achieving robust
  low-level and high-level scene parsing,'' \emph{IEEE TIP}, vol.~28, no.~3,
  pp. 1378--1390, 2018.

\bibitem{li2020gated}
X.~Li, H.~Zhao, L.~Han, Y.~Tong, S.~Tan, and K.~Yang, ``Gated fully fusion for
  semantic segmentation,'' in \emph{AAAI}, 2020.

\bibitem{li2021global}
X.~Li, L.~Zhang, G.~Cheng, K.~Yang, Y.~Tong, X.~Zhu, and T.~Xiang, ``Global
  aggregation then local distribution for scene parsing,'' \emph{IEEE TIP},
  2021.

\bibitem{ocrnet}
Y.~Yuan, X.~Chen, and J.~Wang, ``Object-contextual representations for semantic
  segmentation,'' \emph{ECCV}, 2020.

\bibitem{zhangli_dgcn}
L.~Zhang, X.~Li, A.~Arnab, K.~Yang, Y.~Tong, and P.~H. Torr, ``Dual graph
  convolutional network for semantic segmentation,'' in \emph{BMVC}, 2019.

\bibitem{sfnet}
X.~Li, A.~You, Z.~Zhu, H.~Zhao, M.~Yang, K.~Yang, and Y.~Tong, ``Semantic flow
  for fast and accurate scene parsing,'' in \emph{ECCV}, 2020.

\bibitem{Li2022SFNetFA}
X.~Li, J.~Zhang, Y.~Yang, G.~Cheng, K.~Yang, Y.~Tong, and D.~Tao, ``Sfnet:
  Faster, accurate, and domain agnostic semantic segmentation via semantic
  flow,'' \emph{ArXiv}, vol. abs/2207.04415, 2022.

\bibitem{BiSeNet}
C.~Yu, J.~Wang, C.~Peng, C.~Gao, G.~Yu, and N.~Sang, ``{BiSeNet}: Bilateral
  segmentation network for real-time semantic segmentation,'' in \emph{ECCV},
  2018.

\bibitem{kirillov2020pointrend}
A.~Kirillov, Y.~Wu, K.~He, and R.~Girshick, ``Pointrend: Image segmentation as
  rendering,'' in \emph{CVPR}, 2020.

\bibitem{BoundaryAware}
H.~Ding, X.~Jiang, A.~Q. Liu, N.~M. Thalmann, and G.~Wang, ``Boundary-aware
  feature propagation for scene segmentation,'' in \emph{ICCV}, 2019.

\bibitem{li2020improving}
X.~Li, X.~Li, L.~Zhang, G.~Cheng, J.~Shi, Z.~Lin, S.~Tan, and Y.~Tong,
  ``Improving semantic segmentation via decoupled body and edge supervision,''
  in \emph{ECCV}, 2020.

\bibitem{ebl_he_iccv}
H.~He, X.~Li, G.~Cheng, J.~Shi, Y.~Tong, G.~Meng, V.~Prinet, and L.~Weng,
  ``Enhanced boundary learning for glass-like object segmentation,'' in
  \emph{ICCV}, 2021.

\bibitem{DAnet}
J.~Fu, J.~Liu, H.~Tian, Z.~Fang, and H.~Lu, ``Dual attention network for scene
  segmentation,'' in \emph{CVPR}, 2019.

\bibitem{maskrcnn}
K.~He, G.~Gkioxari, P.~Doll{\'a}r, and R.~Girshick, ``{Mask R-CNN},'' in
  \emph{ICCV}, 2017.

\bibitem{tian2020conditional}
Z.~Tian, C.~Shen, and H.~Chen, ``Conditional convolutions for instance
  segmentation,'' in \emph{ECCV}, 2020.

\bibitem{neven2019instanceSeg}
D.~Neven, B.~D. Brabandere, M.~Proesmans, and L.~V. Gool, ``Instance
  segmentation by jointly optimizing spatial embeddings and clustering
  bandwidth,'' in \emph{CVPR}, 2019.

\bibitem{de2017semanticInstanceLoss}
B.~De~Brabandere, D.~Neven, and L.~Van~Gool, ``Semantic instance segmentation
  with a discriminative loss function,'' \emph{arXiv preprint
  arXiv:1708.02551}, 2017.

\bibitem{htc}
K.~Chen, J.~Pang, J.~Wang, Y.~Xiong, X.~Li, S.~Sun, W.~Feng, Z.~Liu, J.~Shi,
  W.~Ouyang, C.~C. Loy, and D.~Lin, ``Hybrid task cascade for instance
  segmentation,'' in \emph{CVPR}, 2019.

\bibitem{zhang2020MEInst}
R.~Zhang, Z.~Tian, C.~Shen, M.~You, and Y.~Yan, ``Mask encoding for single shot
  instance segmentation,'' in \emph{CVPR}, 2020.

\bibitem{bolya2019yolact}
D.~Bolya, C.~Zhou, F.~Xiao, and Y.~J. Lee, ``{YOLACT}: Real-time instance
  segmentation,'' in \emph{ICCV}, 2019.

\bibitem{qiao2021detectors}
S.~Qiao, L.-C. Chen, and A.~Yuille, ``Detectors: Detecting objects with
  recursive feature pyramid and switchable atrous convolution,'' in
  \emph{CVPR}, 2021.

\bibitem{cheng2020panoptic}
B.~Cheng, M.~D. Collins, Y.~Zhu, T.~Liu, T.~S. Huang, H.~Adam, and L.-C. Chen,
  ``{Panoptic-DeepLab}: A simple, strong, and fast baseline for bottom-up
  panoptic segmentation,'' in \emph{CVPR}, 2020.

\bibitem{chen2019tensormask}
X.~Chen, R.~Girshick, K.~He, and P.~Doll{\'a}r, ``Tensormask: A foundation for
  dense object segmentation,'' in \emph{ICCV}, 2019.

\bibitem{wang2020solov2}
X.~Wang, R.~Zhang, T.~Kong, L.~Li, and C.~Shen, ``{SOLOv2}: Dynamic and fast
  instance segmentation,'' in \emph{NeurIPS}, 2020.

\bibitem{xiong2019upsnet}
Y.~Xiong, R.~Liao, H.~Zhao, R.~Hu, M.~Bai, E.~Yumer, and R.~Urtasun,
  ``{UPSNet}: A unified panoptic segmentation network,'' in \emph{CVPR}, 2019.

\bibitem{li2020panopticFCN}
Y.~Li, H.~Zhao, X.~Qi, L.~Wang, Z.~Li, J.~Sun, and J.~Jia, ``Fully
  convolutional networks for panoptic segmentation,'' \emph{CVPR}, 2021.

\bibitem{axialDeeplab}
H.~Wang, Y.~Zhu, B.~Green, H.~Adam, A.~Yuille, and L.-C. Chen,
  ``{Axial-DeepLab}: Stand-alone axial-attention for panoptic segmentation,''
  in \emph{ECCV}, 2020.

\bibitem{gadde2017semantic}
R.~Gadde, V.~Jampani, and P.~V. Gehler, ``Semantic video cnns through
  representation warping,'' in \emph{ICCV}, 2017.

\bibitem{shelhamer2016clockwork}
E.~Shelhamer, K.~Rakelly, J.~Hoffman, and T.~Darrell, ``Clockwork convnets for
  video semantic segmentation,'' in \emph{ECCV}, 2016.

\bibitem{DFF}
X.~Zhu, Y.~Xiong, J.~Dai, L.~Yuan, and Y.~Wei, ``Deep feature flow for video
  recognition,'' in \emph{CVPR}, 2017.

\bibitem{mask_pro_vis}
G.~Bertasius and L.~Torresani, ``Classifying, segmenting, and tracking object
  instances in video with mask propagation,'' in \emph{CVPR}, 2020.

\bibitem{fu2021compfeat}
Y.~Fu, L.~Yang, D.~Liu, T.~S. Huang, and H.~Shi, ``{CompFeat}: Comprehensive
  feature aggregation for video instance segmentation,'' \emph{AAAI}, 2021.

\bibitem{li2022improving}
X.~Li, H.~He, Y.~Yang, H.~Ding, K.~Yang, G.~Cheng, Y.~Tong, and D.~Tao,
  ``Improving video instance segmentation via temporal pyramid routing,''
  \emph{IEEE TPAMI}, 2022.

\bibitem{vip_deeplab}
S.~Qiao, Y.~Zhu, H.~Adam, A.~Yuille, and L.-C. Chen, ``Vip-deeplab: Learning
  visual perception with depth-aware video panoptic segmentation,'' in
  \emph{CVPR}, 2021.

\bibitem{deformablev2}
X.~Zhu, H.~Hu, S.~Lin, and J.~Dai, ``Deformable convnets v2: More deformable,
  better results,'' in \emph{CVPR}, 2019.

\bibitem{VOS_data}
F.~Perazzi, J.~Pont-Tuset, B.~McWilliams, L.~Van~Gool, M.~Gross, and
  A.~Sorkine-Hornung, ``A benchmark dataset and evaluation methodology for
  video object segmentation,'' in \emph{CVPR}, 2016.

\bibitem{voigtlaender2019mots}
P.~Voigtlaender, M.~Krause, A.~Osep, J.~Luiten, B.~B.~G. Sekar, A.~Geiger, and
  B.~Leibe, ``{MOTS}: Multi-object tracking and segmentation,'' in \emph{CVPR},
  2019.

\bibitem{qi2017pointnet}
C.~R. Qi, H.~Su, K.~Mo, and L.~J. Guibas, ``{PointNet}: Deep learning on point
  sets for 3d classification and segmentation,'' in \emph{CVPR}, 2017, pp.
  652--660.

\bibitem{qi2017pointnet++}
C.~R. Qi, L.~Yi, H.~Su, and L.~J. Guibas, ``{Pointnet++}: Deep hierarchical
  feature learning on point sets in a metric space,'' \emph{NeurIPS}, 2017.

\bibitem{yang2019learning}
B.~Yang, J.~Wang, R.~Clark, Q.~Hu, S.~Wang, A.~Markham, and N.~Trigoni,
  ``Learning object bounding boxes for 3d instance segmentation on point
  clouds,'' in \emph{NeurIPS}, 2019.

\bibitem{yi2019gspn}
L.~Yi, W.~Zhao, H.~Wang, M.~Sung, and L.~J. Guibas, ``{GSPN}: Generative shape
  proposal network for 3d instance segmentation in point cloud,'' in
  \emph{CVPR}, 2019.

\bibitem{wang2018sgpn}
W.~Wang, R.~Yu, Q.~Huang, and U.~Neumann, ``{SGPN}: Similarity group proposal
  network for 3d point cloud instance segmentation,'' in \emph{CVPR}, 2018.

\bibitem{jiang2020pointgroup}
L.~Jiang, H.~Zhao, S.~Shi, S.~Liu, C.-W. Fu, and J.~Jia, ``{PointGroup}:
  Dual-set point grouping for 3d instance segmentation,'' in \emph{CVPR}, 2020.

\bibitem{mao2019interpolated}
J.~Mao, X.~Wang, and H.~Li, ``Interpolated convolutional networks for 3d point
  cloud understanding,'' in \emph{ICCV}, 2019.

\bibitem{hu2020randla}
Q.~Hu, B.~Yang, L.~Xie, S.~Rosa, Y.~Guo, Z.~Wang, N.~Trigoni, and A.~Markham,
  ``{RandLA-Net}: Efficient semantic segmentation of large-scale point
  clouds,'' in \emph{CVPR}, 2020.

\bibitem{cheng20212}
R.~Cheng, R.~Razani, E.~Taghavi, E.~Li, and B.~Liu, ``{(AF)2-S3Net}: Attentive
  feature fusion with adaptive feature selection for sparse semantic
  segmentation network,'' in \emph{CVPR}, 2021.

\bibitem{zhou2021panoptic}
Z.~Zhou, Y.~Zhang, and H.~Foroosh, ``{Panoptic-PolarNet}: Proposal-free lidar
  point cloud panoptic segmentation,'' in \emph{CVPR}, 2021.

\bibitem{xu2022sparse}
S.~Xu, R.~Wan, M.~Ye, X.~Zou, and T.~Cao, ``Sparse cross-scale attention
  network for efficient lidar panoptic segmentation,'' \emph{AAAI}, 2022.

\bibitem{Hong_2021_CVPR}
F.~Hong, H.~Zhou, X.~Zhu, H.~Li, and Z.~Liu, ``{LiDAR}-based panoptic
  segmentation via dynamic shifting network,'' in \emph{CVPR}, 2021.

\bibitem{aygun20214d}
M.~Aygun, A.~Osep, M.~Weber, M.~Maximov, C.~Stachniss, J.~Behley, and
  L.~Leal-Taix{\'e}, ``{4D} panoptic lidar segmentation,'' in \emph{CVPR},
  2021.

\bibitem{Zhu_2021_CVPR}
X.~Zhu, H.~Zhou, T.~Wang, F.~Hong, Y.~Ma, W.~Li, H.~Li, and D.~Lin,
  ``Cylindrical and asymmetrical 3d convolution networks for lidar
  segmentation,'' in \emph{CVPR}, 2021.

\bibitem{chen2018a2net}
Y.~Chen, Y.~Kalantidis, J.~Li, S.~Yan, and J.~Feng, ``{A2-Nets}: Double
  attention networks,'' \emph{NeurIPS}, 2018.

\bibitem{ren2015faster}
S.~Ren, K.~He, R.~Girshick, and J.~Sun, ``{Faster R-CNN}: Towards real-time
  object detection with region proposal networks,'' in \emph{NeurIPS}, 2015.

\bibitem{fpn}
T.-Y. {Lin}, P.~{Dollár}, R.~B. {Girshick}, K.~{He}, B.~{Hariharan}, and S.~J.
  {Belongie}, ``Feature pyramid networks for object detection,'' in
  \emph{CVPR}, 2017.

\bibitem{focal_loss}
T.-Y. Lin, P.~Goyal, R.~Girshick, K.~He, and P.~Doll{\'a}r, ``Focal loss for
  dense object detection,'' in \emph{ICCV}, 2017.

\bibitem{tian2021fcos}
Z.~Tian, C.~Shen, H.~Chen, and T.~He, ``{FCOS}: A simple and strong anchor-free
  object detector,'' \emph{TPAMI}, 2021.

\bibitem{nasfpn}
G.~Ghiasi, T.-Y. Lin, and Q.~V. Le, ``{NAS-FPN}: Learning scalable feature
  pyramid architecture for object detection,'' in \emph{CVPR}, 2019.

\bibitem{li2023litedetr}
F.~Li, A.~Zeng, S.~Liu, H.~Zhang, H.~Li, L.~Zhang, and L.~M. Ni, ``{Lite DETR}:
  An interleaved multi-scale encoder for efficient detr,'' \emph{CVPR}, 2023.

\bibitem{kuhn1955hungarian}
H.~W. Kuhn, ``The hungarian method for the assignment problem,'' \emph{Naval
  research logistics quarterly}, 1955.

\bibitem{dice_loss}
F.~Milletari, N.~Navab, and S.~Ahmadi, ``{V-Net: Fully} convolutional neural
  networks for volumetric medical image segmentation,'' in \emph{3DV}, 2016.

\bibitem{xie2021segformer}
E.~Xie, W.~Wang, Z.~Yu, A.~Anandkumar, J.~M. Alvarez, and P.~Luo,
  ``{SegFormer}: Simple and efficient design for semantic segmentation with
  transformers,'' in \emph{NeurIPS}, 2021.

\bibitem{SETR}
S.~Zheng, J.~Lu, H.~Zhao, X.~Zhu, Z.~Luo, Y.~Wang, Y.~Fu, J.~Feng, T.~Xiang,
  P.~H. Torr, and L.~Zhang, ``Rethinking semantic segmentation from a
  sequence-to-sequence perspective with transformers,'' in \emph{CVPR}, 2021.

\bibitem{deeplabv3plus}
L.-C. Chen, Y.~Zhu, G.~Papandreou, F.~Schroff, and H.~Adam, ``Encoder-decoder
  with atrous separable convolution for semantic image segmentation,'' in
  \emph{ECCV}, 2018.

\bibitem{OMGSeg}
X.~Li, H.~Yuan, W.~Li, H.~Ding, S.~Wu, W.~Zhang, Y.~Li, K.~Chen, and C.~C. Loy,
  ``Omg-seg: Is one model good enough for all segmentation?'' in \emph{CVPR},
  2024.

\bibitem{athar2023tarvis}
A.~Athar, A.~Hermans, J.~Luiten, D.~Ramanan, and B.~Leibe, ``{TarViS}: A
  unified approach for target-based video segmentation,'' \emph{CVPR}, 2023.

\bibitem{yuan2021polyphonicformer}
H.~Yuan, X.~Li, Y.~Yang, G.~Cheng, J.~Zhang, Y.~Tong, L.~Zhang, and D.~Tao,
  ``Polyphonicformer: Unified query learning for depth-aware video panoptic
  segmentation,'' \emph{ECCV}, 2022.

\bibitem{wu2023uniref++}
J.~Wu, Y.~Jiang, B.~Yan, H.~Lu, Z.~Yuan, and P.~Luo, ``Uniref++: Segment every
  reference object in spatial and temporal spaces,'' \emph{arXiv preprint
  arXiv:2312.15715}, 2023.

\bibitem{DeiT}
H.~Touvron, M.~Cord, M.~Douze, F.~Massa, A.~Sablayrolles, and H.~J{\'e}gou,
  ``Training data-efficient image transformers \& distillation through
  attention,'' in \emph{ICML}, 2021.

\bibitem{fan2021mvitv1}
H.~Fan, B.~Xiong, K.~Mangalam, Y.~Li, Z.~Yan, J.~Malik, and C.~Feichtenhofer,
  ``Multiscale vision transformers,'' in \emph{ICCV}, 2021.

\bibitem{li2022mvitv2}
Y.~Li, C.-Y. Wu, H.~Fan, K.~Mangalam, B.~Xiong, J.~Malik, and C.~Feichtenhofer,
  ``{MViTv2}: Improved multiscale vision transformers for classification and
  detection,'' in \emph{CVPR}, 2022.

\bibitem{lee2022mpvit}
Y.~Lee, J.~Kim, J.~Willette, and S.~J. Hwang, ``{MPViT}: Multi-path vision
  transformer for dense prediction,'' in \emph{CVPR}, 2022.

\bibitem{ali2021xcit}
A.~Ali, H.~Touvron, M.~Caron, P.~Bojanowski, M.~Douze, A.~Joulin, I.~Laptev,
  N.~Neverova, G.~Synnaeve, J.~Verbeek \emph{et~al.}, ``{XCiT}:
  Cross-covariance image transformers,'' \emph{NeurIPS}, 2021.

\bibitem{Wang_2021_ICCV_PVT}
W.~Wang, E.~Xie, X.~Li, D.-P. Fan, K.~Song, D.~Liang, T.~Lu, P.~Luo, and
  L.~Shao, ``Pyramid vision transformer: A versatile backbone for dense
  prediction without convolutions,'' in \emph{ICCV}, 2021.

\bibitem{chen2021crossvit}
C.-F.~R. Chen, Q.~Fan, and R.~Panda, ``Crossvit: Cross-attention multi-scale
  vision transformer for image classification,'' in \emph{ICCV}, 2021.

\bibitem{zhang2020feature}
D.~Zhang, H.~Zhang, J.~Tang, M.~Wang, X.~Hua, and Q.~Sun, ``Feature pyramid
  transformer,'' in \emph{ECCV}, 2020.

\bibitem{Xu_2021_ICCV}
W.~Xu, Y.~Xu, T.~Chang, and Z.~Tu, ``Co-scale conv-attentional image
  transformers,'' in \emph{ICCV}, 2021.

\bibitem{guo2021cmt}
J.~Guo, K.~Han, H.~Wu, C.~Xu, Y.~Tang, C.~Xu, and Y.~Wang, ``{CMT}:
  Convolutional neural networks meet vision transformers,'' \emph{CVPR}, 2022.

\bibitem{chu2021Twins}
X.~Chu, Z.~Tian, Y.~Wang, B.~Zhang, H.~Ren, X.~Wei, H.~Xia, and C.~Shen,
  ``Twins: Revisiting the design of spatial attention in vision transformers,''
  in \emph{NeurIPS}, 2021.

\bibitem{wu2021cvt}
H.~Wu, B.~Xiao, N.~Codella, M.~Liu, X.~Dai, L.~Yuan, and L.~Zhang, ``{CvT}:
  Introducing convolutions to vision transformers,'' \emph{ICCV}, 2021.

\bibitem{xu2021vitae}
Y.~Xu, Q.~Zhang, J.~Zhang, and D.~Tao, ``{ViTAE}: Vision transformer advanced
  by exploring intrinsic inductive bias,'' \emph{NeurIPS}, 2021.

\bibitem{liu2022convnet}
Z.~Liu, H.~Mao, C.-Y. Wu, C.~Feichtenhofer, T.~Darrell, and S.~Xie, ``A
  {ConvNet} for the 2020s,'' \emph{CVPR}, 2022.

\bibitem{han2021connection}
Q.~Han, Z.~Fan, Q.~Dai, L.~Sun, M.-M. Cheng, J.~Liu, and J.~Wang, ``On the
  connection between local attention and dynamic depth-wise convolution,'' in
  \emph{ICLR}, 2022.

\bibitem{guo2022segnext}
M.-H. Guo, C.-Z. Lu, Q.~Hou, Z.~Liu, M.-M. Cheng, and S.-M. Hu, ``{SegNeXt}:
  Rethinking convolutional attention design for semantic segmentation,''
  \emph{NeurIPS}, 2022.

\bibitem{yu2022metaformer}
W.~Yu, M.~Luo, P.~Zhou, C.~Si, Y.~Zhou, X.~Wang, J.~Feng, and S.~Yan,
  ``{MetaFormer} is actually what you need for vision,'' in \emph{CVPR}, 2022.

\bibitem{dai2022demystify}
J.~Dai, M.~Shi, W.~Wang, S.~Wu, L.~Xing, W.~Wang, X.~Zhu, L.~Lu, J.~Zhou,
  X.~Wang \emph{et~al.}, ``Demystify transformers \& convolutions in modern
  image deep networks,'' \emph{arXiv preprint arXiv:2211.05781}, 2022.

\bibitem{chen2020simple}
T.~Chen, S.~Kornblith, M.~Norouzi, and G.~Hinton, ``A simple framework for
  contrastive learning of visual representations,'' \emph{ICML}, 2020.

\bibitem{he2020momentum}
K.~He, H.~Fan, Y.~Wu, S.~Xie, and R.~Girshick, ``Momentum contrast for
  unsupervised visual representation learning,'' in \emph{CVPR}, 2020.

\bibitem{chen2021mocov3}
X.~Chen*, S.~Xie*, and K.~He, ``An empirical study of training self-supervised
  vision transformers,'' \emph{ICCV}, 2021.

\bibitem{bao2021beit}
H.~Bao, L.~Dong, and F.~Wei, ``{BEiT: BERT} pre-training of image
  transformers,'' \emph{ICLR}, 2022.

\bibitem{wei2022maskedfeat}
C.~Wei, H.~Fan, S.~Xie, C.-Y. Wu, A.~Yuille, and C.~Feichtenhofer, ``Masked
  feature prediction for self-supervised visual pre-training,'' in \emph{CVPR},
  2022.

\bibitem{gandelsman2022test}
Y.~Gandelsman, Y.~Sun, X.~Chen, and A.~A. Efros, ``Test-time training with
  masked autoencoders,'' in \emph{NeurIPS}, 2022.

\bibitem{hu2022exploring}
R.~Hu, S.~Debnath, S.~Xie, and X.~Chen, ``Exploring long-sequence masked
  autoencoders,'' \emph{arXiv preprint arXiv:2210.07224}, 2022.

\bibitem{gao2022convmae}
P.~Gao, T.~Ma, H.~Li, J.~Dai, and Y.~Qiao, ``{ConvMAE}: Masked convolution
  meets masked autoencoders,'' \emph{NeurIPS}, 2022.

\bibitem{radford2021learning_clip}
A.~Radford, J.~W. Kim, C.~Hallacy, A.~Ramesh, G.~Goh, S.~Agarwal, G.~Sastry,
  A.~Askell, P.~Mishkin, J.~Clark \emph{et~al.}, ``Learning transferable visual
  models from natural language supervision,'' in \emph{ICML}, 2021.

\bibitem{li2022scaling}
Y.~Li, H.~Fan, R.~Hu, C.~Feichtenhofer, and K.~He, ``Scaling language-image
  pre-training via masking,'' \emph{arXiv preprint arXiv:2212.00794}, 2022.

\bibitem{peize2020sparse}
P.~Sun, R.~Zhang, Y.~Jiang, T.~Kong, C.~Xu, W.~Zhan, M.~Tomizuka, L.~Li,
  Z.~Yuan, C.~Wang, and P.~Luo, ``{SparseR-CNN}: End-to-end object detection
  with learnable proposals,'' \emph{CVPR}, 2021.

\bibitem{QueryInst}
Y.~Fang, S.~Yang, X.~Wang, Y.~Li, C.~Fang, Y.~Shan, B.~Feng, and W.~Liu,
  ``Instances as queries,'' in \emph{ICCV}, 2021.

\bibitem{hu2021ISTR}
J.~Hu, L.~Cao, Y.~Lu, S.~Zhang, K.~Li, F.~Huang, L.~Shao, and R.~Ji, ``{ISTR}:
  End-to-end instance segmentation via transformers,'' \emph{arXiv preprint
  arXiv:2105.00637}, 2021.

\bibitem{dong2021solq}
B.~Dong, F.~Zeng, T.~Wang, X.~Zhang, and Y.~Wei, ``{SOLQ}: Segmenting objects
  by learning queries,'' \emph{NeurIPS}, 2021.

\bibitem{he2021boundarysqueeze}
H.~He, X.~Li, Y.~Yang, G.~Cheng, Y.~Tong, L.~Weng, Z.~Lin, and S.~Xiang,
  ``Boundarysqueeze: Image segmentation as boundary squeezing,'' \emph{arXiv
  preprint arXiv:2105.11668}, 2021.

\bibitem{zhang2021knet}
W.~Zhang, J.~Pang, K.~Chen, and C.~C. Loy, ``{K-Net}: Towards unified image
  segmentation,'' in \emph{NeurIPS}, 2021.

\bibitem{cheng2021maskformer}
B.~Cheng, A.~G. Schwing, and A.~Kirillov, ``Per-pixel classification is not all
  you need for semantic segmentation,'' in \emph{NeurIPS}, 2021.

\bibitem{li2021panoptic}
Z.~Li, W.~Wang, E.~Xie, Z.~Yu, A.~Anandkumar, J.~M. Alvarez, T.~Lu, and P.~Luo,
  ``Panoptic segformer: Delving deeper into panoptic segmentation with
  transformers,'' \emph{CVPR}, 2022.

\bibitem{VIS_TR}
Y.~Wang, Z.~Xu, X.~Wang, C.~Shen, B.~Cheng, H.~Shen, and H.~Xia, ``End-to-end
  video instance segmentation with transformers,'' in \emph{CVPR}, 2021.

\bibitem{zhou2022transvod}
Q.~Zhou, X.~Li, L.~He, Y.~Yang, G.~Cheng, Y.~Tong, L.~Ma, and D.~Tao,
  ``{TransVOD}: End-to-end video object detection with spatial-temporal
  transformers,'' \emph{PAMI}, 2022.

\bibitem{yang2022tevit}
S.~Yang, X.~Wang, Y.~Li, Y.~Fang, J.~Fang, Liu, X.~Zhao, and Y.~Shan,
  ``Temporally efficient vision transformer for video instance segmentation,''
  in \emph{CVPR}, 2022.

\bibitem{cheng2021mask2formervis}
B.~Cheng, A.~Choudhuri, I.~Misra, A.~Kirillov, R.~Girdhar, and A.~G. Schwing,
  ``Mask2former for video instance segmentation,'' \emph{arXiv preprint
  arXiv:2112.10764}, 2021.

\bibitem{IFC_21}
S.~Hwang, M.~Heo, S.~W. Oh, and S.~J. Kim, ``Video instance segmentation using
  inter-frame communication transformers,'' \emph{NeurIPS}, 2021.

\bibitem{seqformer}
J.~Wu, Y.~Jiang, S.~Bai, W.~Zhang, and X.~Bai, ``{SeqFormer}: Sequential
  transformer for video instance segmentation,'' in \emph{ECCV}, 2022.

\bibitem{li2022videoknet}
X.~Li, W.~Zhang, J.~Pang, K.~Chen, G.~Cheng, Y.~Tong, and C.~C. Loy, ``Video
  {K-Net}: A simple, strong, and unified baseline for video segmentation,'' in
  \emph{CVPR}, 2022.

\bibitem{kim2022tubeformer}
D.~Kim, J.~Xie, H.~Wang, S.~Qiao, Q.~Yu, H.-S. Kim, H.~Adam, I.~S. Kweon, and
  L.-C. Chen, ``{TubeFormer-DeepLab}: Video mask transformer,'' in \emph{CVPR},
  2022.

\bibitem{meng2021conditional}
D.~Meng, X.~Chen, Z.~Fan, G.~Zeng, H.~Li, Y.~Yuan, L.~Sun, and J.~Wang,
  ``Conditional detr for fast training convergence,'' in \emph{CVPR}, 2021.

\bibitem{chen2022conditional}
X.~Chen, F.~Wei, G.~Zeng, and J.~Wang, ``Conditional detr v2: Efficient
  detection transformer with box queries,'' \emph{arXiv preprint
  arXiv:2207.08914}, 2022.

\bibitem{wang2022anchor}
Y.~Wang, X.~Zhang, T.~Yang, and J.~Sun, ``Anchor {DETR}: Query design for
  transformer-based detector,'' in \emph{AAAI}, 2022.

\bibitem{liu2022dabdetr}
S.~Liu, F.~Li, H.~Zhang, X.~Yang, X.~Qi, H.~Su, J.~Zhu, and L.~Zhang,
  ``{DAB}-{DETR}: Dynamic anchor boxes are better queries for {DETR},'' in
  \emph{ICLR}, 2022.

\bibitem{li2022dn}
F.~Li, H.~Zhang, S.~Liu, J.~Guo, L.~M. Ni, and L.~Zhang, ``{DN}-{DETR}:
  Accelerate detr training by introducing query denoising,'' in \emph{CVPR},
  2022.

\bibitem{zhang2022dino}
H.~Zhang, F.~Li, S.~Liu, L.~Zhang, H.~Su, J.~Zhu, L.~Ni, and H.-Y. Shum,
  ``{DINO}: {DETR} with improved denoising anchor boxes for end-to-end object
  detection,'' in \emph{ICLR}, 2023.

\bibitem{li2022maskdino}
F.~Li, H.~Zhang, H.~xu, S.~Liu, L.~Zhang, L.~M. Ni, and H.-Y. Shum, ``{Mask}
  {DINO}: Towards a unified transformer-based framework for object detection
  and segmentation,'' in \emph{CVPR}, 2023.

\bibitem{Instance_Unique_Querying}
W.~Wang, J.~Liang, and D.~Liu, ``Learning equivariant segmentation with
  instance-unique querying,'' \emph{NeurIPS}, 2022.

\bibitem{jia2022detrs}
D.~Jia, Y.~Yuan, H.~He, X.~Wu, H.~Yu, W.~Lin, L.~Sun, C.~Zhang, and H.~Hu,
  ``{DETRs} with hybrid matching,'' in \emph{CVPR}, 2023.

\bibitem{chen2022group}
Q.~Chen, X.~Chen, J.~Wang, H.~Feng, J.~Han, E.~Ding, G.~Zeng, and J.~Wang,
  ``{Group} {DETR}: Fast detr training with group-wise one-to-many
  assignment,'' \emph{arXiv preprint arXiv:2207.13085}, 2022.

\bibitem{zong2022detrs}
Z.~Zong, G.~Song, and Y.~Liu, ``Detrs with collaborative hybrid assignments
  training,'' \emph{arXiv preprint arXiv:2211.12860}, 2022.

\bibitem{meinhardt2021trackformer}
T.~Meinhardt, A.~Kirillov, L.~Leal-Taixe, and C.~Feichtenhofer,
  ``{TrackFormer}: Multi-object tracking with transformers,'' in \emph{CVPR},
  2022.

\bibitem{transtrack}
P.~Sun, J.~Cao, Y.~Jiang, R.~Zhang, E.~Xie, Z.~Yuan, C.~Wang, and P.~Luo,
  ``{TransTrack}: Multiple-object tracking with transformer,'' \emph{arXiv
  preprint arXiv: 2012.15460}, 2020.

\bibitem{zeng2021motr}
F.~Zeng, B.~Dong, T.~Wang, C.~Chen, X.~Zhang, and Y.~Wei, ``{MOTR}: End-to-end
  multiple-object tracking with transformer,'' in \emph{ECCV}, 2022.

\bibitem{huang2022minvis}
D.-A. Huang, Z.~Yu, and A.~Anandkumar, ``{MinVIS}: A minimal video instance
  segmentation framework without video-based training,'' \emph{NeurIPS}, 2022.

\bibitem{IDOL}
J.~Wu, Q.~Liu, Y.~Jiang, S.~Bai, A.~Yuille, and X.~Bai, ``In defense of online
  models for video instance segmentation,'' in \emph{ECCV}, 2022.

\bibitem{panopticpartformer}
X.~Li, S.~Xu, Y.~Yang, G.~Cheng, Y.~Tong, and D.~Tao, ``Panoptic-partformer:
  Learning a unified model for panoptic part segmentation,'' in \emph{ECCV},
  2022.

\bibitem{gao2022panopticdepth}
N.~Gao, F.~He, J.~Jia, Y.~Shan, H.~Zhang, X.~Zhao, and K.~Huang,
  ``{PanopticDepth}: A unified framework for depth-aware panoptic
  segmentation,'' in \emph{CVPR}, 2022.

\bibitem{xu2022fashionformer}
S.~Xu, X.~Li, J.~Wang, G.~Cheng, Y.~Tong, and D.~Tao, ``Fashionformer: A
  simple, effective and unified baseline for human fashion segmentation and
  recognition,'' \emph{ECCV}, 2022.

\bibitem{xu2022multi}
Y.~Xu, X.~Li, H.~Yuan, Y.~Yang, J.~Zhang, Y.~Tong, L.~Zhang, and D.~Tao,
  ``Multi-task learning with multi-query transformer for dense prediction,''
  \emph{arXiv preprint arXiv:2205.14354}, 2022.

\bibitem{invpt2022}
H.~Ye and D.~Xu, ``Inverted pyramid multi-task transformer for dense scene
  understanding,'' in \emph{ECCV}, 2022.

\bibitem{VLT_iccv2021}
H.~Ding, C.~Liu, S.~Wang, and X.~Jiang, ``Vision-language transformer and query
  generation for referring segmentation,'' in \emph{ICCV}, 2021.

\bibitem{LAVT_22cvpr}
Z.~Yang, J.~Wang, Y.~Tang, K.~Chen, H.~Zhao, and P.~H. Torr, ``{LAVT}:
  Language-aware vision transformer for referring image segmentation,'' in
  \emph{CVPR}, 2022.

\bibitem{ReSTR_2022_CVPR}
N.~Kim, D.~Kim, C.~Lan, W.~Zeng, and S.~Kwak, ``{ReSTR}: Convolution-free
  referring image segmentation using transformers,'' in \emph{CVPR}, 2022.

\bibitem{CRIS_2022_CVPR}
Z.~Wang, Y.~Lu, Q.~Li, X.~Tao, Y.~Guo, M.~Gong, and T.~Liu, ``{CRIS}:
  {CLIP}-driven referring image segmentation,'' in \emph{CVPR}, 2022.

\bibitem{botach2022end}
A.~Botach, E.~Zheltonozhskii, and C.~Baskin, ``End-to-end referring video
  object segmentation with multimodal transformers,'' in \emph{CVPR}, 2022.

\bibitem{ding2022language}
Z.~Ding, T.~Hui, J.~Huang, X.~Wei, J.~Han, and S.~Liu, ``Language-bridged
  spatial-temporal interaction for referring video object segmentation,'' in
  \emph{CVPR}, 2022.

\bibitem{wu2022towards_robust_ris}
J.~Wu, X.~Li, X.~Li, H.~Ding, Y.~Tong, and D.~Tao, ``Towards robust referring
  image segmentation,'' \emph{arXiv preprint arXiv:2209.09554}, 2022.

\bibitem{wu2022language}
J.~Wu, Y.~Jiang, P.~Sun, Z.~Yuan, and P.~Luo, ``Language as queries for
  referring video object segmentation,'' in \emph{CVPR}, 2022.

\bibitem{zhang2021few}
G.~Zhang, G.~Kang, Y.~Yang, and Y.~Wei, ``Few-shot segmentation via
  cycle-consistent transformer,'' in \emph{NeurIPS}, 2021.

\bibitem{yang2021associating}
Z.~Yang, Y.~Wei, and Y.~Yang, ``Associating objects with transformers for video
  object segmentation,'' in \emph{NeurIPS}, 2021.

\bibitem{park2022matteformer}
G.~Park, S.~Son, J.~Yoo, S.~Kim, and N.~Kwak, ``Matteformer: Transformer-based
  image matting via prior-tokens,'' in \emph{CVPR}, 2022.

\bibitem{shi2022transformer}
B.~Shi, D.~Jiang, X.~Zhang, H.~Li, W.~Dai, J.~Zou, H.~Xiong, and Q.~Tian, ``A
  transformer-based decoder for semantic segmentation with multi-level context
  mining,'' in \emph{ECCV}, 2022.

\bibitem{lin2022structtoken}
F.~Lin, Z.~Liang, J.~He, M.~Zheng, S.~Tian, and K.~Chen, ``Structtoken:
  Rethinking semantic segmentation with structural prior,'' \emph{arXiv
  preprint arXiv:2203.12612}, 2022.

\bibitem{yu2022batman}
Y.~Yu, J.~Yuan, G.~Mittal, L.~Fuxin, and M.~Chen, ``{BATMAN}: Bilateral
  attention transformer in motion-appearance neighboring space for video object
  segmentation,'' in \emph{ECCV}, 2022.

\bibitem{jiao2022mask}
S.~Jiao, G.~Zhang, S.~Navasardyan, L.~Chen, Y.~Zhao, Y.~Wei, and H.~Shi, ``Mask
  matching transformer for few-shot segmentation,'' \emph{arXiv preprint
  arXiv:2301.01208}, 2022.

\bibitem{liu2022swin}
Z.~Liu, H.~Hu, Y.~Lin, Z.~Yao, Z.~Xie, Y.~Wei, J.~Ning, Y.~Cao, Z.~Zhang,
  L.~Dong \emph{et~al.}, ``Swin transformer v2: Scaling up capacity and
  resolution,'' in \emph{CVPR}, 2022.

\bibitem{chen2021cyclemlp}
S.~Chen, E.~Xie, C.~Ge, R.~Chen, D.~Liang, and P.~Luo, ``{CycleMLP}: A mlp-like
  architecture for dense prediction,'' \emph{ICLR}, 2022.

\bibitem{tolstikhin2021mlp}
I.~O. Tolstikhin, N.~Houlsby, A.~Kolesnikov, L.~Beyer, X.~Zhai, T.~Unterthiner,
  J.~Yung, A.~Steiner, D.~Keysers, J.~Uszkoreit \emph{et~al.}, ``{MLP}-mixer:
  An all-{MLP} architecture for vision,'' \emph{NeurIPS}, 2021.

\bibitem{chen2022cyclemlp}
S.~Chen, E.~Xie, C.~GE, R.~Chen, D.~Liang, and P.~Luo, ``Cycle{MLP}: A
  {MLP}-like architecture for dense prediction,'' in \emph{ICLR}, 2022.

\bibitem{guo2021hire}
J.~Guo, Y.~Tang, K.~Han, X.~Chen, H.~Wu, C.~Xu, C.~Xu, and Y.~Wang,
  ``Hire-{MLP}: Vision {MLP} via hierarchical rearrangement,'' in \emph{CVPR},
  2022.

\bibitem{chen2023context}
X.~Chen, M.~Ding, X.~Wang, Y.~Xin, S.~Mo, Y.~Wang, S.~Han, P.~Luo, G.~Zeng, and
  J.~Wang, ``Context autoencoder for self-supervised representation learning,''
  \emph{IJCV}, 2023.

\bibitem{tian2023designing}
K.~Tian, Y.~Jiang, Q.~Diao, C.~Lin, L.~Wang, and Z.~Yuan, ``Designing bert for
  convolutional networks: Sparse and hierarchical masked modeling,''
  \emph{ICLR}, 2023.

\bibitem{caron2021emergingDINO}
M.~Caron, H.~Touvron, I.~Misra, H.~J{\'e}gou, J.~Mairal, P.~Bojanowski, and
  A.~Joulin, ``Emerging properties in self-supervised vision transformers,'' in
  \emph{ICCV}, 2021.

\bibitem{jia2021scaling_align}
C.~Jia, Y.~Yang, Y.~Xia, Y.-T. Chen, Z.~Parekh, H.~Pham, Q.~Le, Y.-H. Sung,
  Z.~Li, and T.~Duerig, ``Scaling up visual and vision-language representation
  learning with noisy text supervision,'' in \emph{ICML}, 2021.

\bibitem{li2022uniperceiver_v2}
H.~Li, J.~Zhu, X.~Jiang, X.~Zhu, H.~Li, C.~Yuan, X.~Wang, Y.~Qiao, X.~Wang,
  W.~Wang \emph{et~al.}, ``{Uni-Perceiver v2}: A generalist model for
  large-scale vision and vision-language tasks,'' \emph{arXiv preprint
  arXiv:2211.09808}, 2022.

\bibitem{videomae}
Z.~Tong, Y.~Song, J.~Wang, and L.~Wang, ``Videomae: Masked autoencoders are
  data-efficient learners for self-supervised video pre-training,'' \emph{arXiv
  preprint arXiv:2203.12602}, 2022.

\bibitem{MaskedAutoencodersSpatiotemporal2022}
C.~Feichtenhofer, H.~Fan, Y.~Li, and K.~He, ``Masked autoencoders as
  spatiotemporal learners,'' \emph{arXiv:2205.09113}, 2022.

\bibitem{liu2022video}
Z.~Liu, J.~Ning, Y.~Cao, Y.~Wei, Z.~Zhang, S.~Lin, and H.~Hu, ``Video swin
  transformer,'' in \emph{CVPR}, 2022.

\bibitem{ding2022vlt}
H.~Ding, C.~Liu, S.~Wang, and X.~Jiang, ``{VLT}: Vision-language transformer
  and query generation for referring segmentation,'' \emph{TPAMI}, 2022.

\bibitem{yu2022soit}
X.~Yu, D.~Shi, X.~Wei, Y.~Ren, T.~Ye, and W.~Tan, ``{SOIT}: Segmenting objects
  with instance-aware transformers,'' in \emph{AAAI}, 2022.

\bibitem{strudel2021_segmenter}
R.~Strudel, R.~Garcia, I.~Laptev, and C.~Schmid, ``Segmenter: Transformer for
  semantic segmentation,'' \emph{ICCV}, 2021.

\bibitem{cheng2021mask2former}
B.~Cheng, I.~Misra, A.~G. Schwing, A.~Kirillov, and R.~Girdhar,
  ``Masked-attention mask transformer for universal image segmentation,'' in
  \emph{CVPR}, 2022.

\bibitem{detectron2}
Y.~Wu, A.~Kirillov, F.~Massa, W.-Y. Lo, and R.~Girshick, ``Detectron2,''
  \url{https://github.com/facebookresearch/detectron2}, 2019.

\bibitem{yu2022cmt}
Q.~Yu, H.~Wang, D.~Kim, S.~Qiao, M.~Collins, Y.~Zhu, H.~Adam, A.~Yuille, and
  L.-C. Chen, ``{CMT-DeepLab}: Clustering mask transformers for panoptic
  segmentation,'' in \emph{CVPR}, 2022.

\bibitem{kmax_deeplab_2022}
Q.~Yu, H.~Wang, S.~Qiao, M.~Collins, Y.~Zhu, H.~Adam, A.~Yuille, and L.-C.
  Chen, ``k-means mask transformer,'' in \emph{ECCV}, 2022.

\bibitem{Cheng2022SparseInst}
T.~Cheng, X.~Wang, S.~Chen, W.~Zhang, Q.~Zhang, C.~Huang, Z.~Zhang, and W.~Liu,
  ``Sparse instance activation for real-time instance segmentation,'' in
  \emph{CVPR}, 2022.

\bibitem{zhang2022_SAMDETR}
G.~Zhang, Z.~Luo, Y.~Yu, K.~Cui, and S.~Lu, ``Accelerating {DETR} convergence
  via semantic-aligned matching,'' in \emph{CVPR}, 2022.

\bibitem{gao2021fast}
P.~Gao, M.~Zheng, X.~Wang, J.~Dai, and H.~Li, ``Fast convergence of detr with
  spatially modulated co-attention,'' \emph{ICCV}, 2021.

\bibitem{gao2022adamixer}
Z.~Gao, L.~Wang, B.~Han, and S.~Guo, ``{AdaMixer}: A fast-converging
  query-based object detector,'' in \emph{CVPR}, 2022.

\bibitem{zheng2020end}
M.~Zheng, P.~Gao, R.~Zhang, K.~Li, X.~Wang, H.~Li, and H.~Dong, ``End-to-end
  object detection with adaptive clustering transformer,'' \emph{BMVC}, 2021.

\bibitem{dai2021dynamic}
X.~Dai, Y.~Chen, J.~Yang, P.~Zhang, L.~Yuan, and L.~Zhang, ``Dynamic {DETR}:
  End-to-end object detection with dynamic attention,'' in \emph{ICCV}, 2021.

\bibitem{roh2022sparse}
B.~Roh, J.~Shin, W.~Shin, and S.~Kim, ``Sparse detr: Efficient end-to-end
  object detection with learnable sparsity,'' in \emph{ICLR}, 2022.

\bibitem{heo2022vita}
M.~Heo, S.~Hwang, S.~W. Oh, J.-Y. Lee, and S.~J. Kim, ``{VITA}: Video instance
  segmentation via object token association,'' \emph{NeurIPS}, 2022.

\bibitem{zou2022xdecoder}
X.~Zou, Z.-Y. Dou, J.~Yang, Z.~Gan, L.~Li, C.~Li, X.~Dai, J.~Wang, L.~Yuan,
  N.~Peng, L.~Wang, Y.~J. Lee, and J.~Gao, ``Generalized decoding for pixel,
  image and language,'' in \emph{CVPR}, 2023.

\bibitem{cai2022x}
Z.~Cai, G.~Kwon, A.~Ravichandran, E.~Bas, Z.~Tu, R.~Bhotika, and S.~Soatto,
  ``{X-DETR}: A versatile architecture for instance-wise vision-language
  tasks,'' in \emph{ECCV}, 2022.

\bibitem{shin2023video}
I.~Shin, D.~Kim, Q.~Yu, J.~Xie, H.-S. Kim, B.~Green, I.~S. Kweon, K.-J. Yoon,
  and L.-C. Chen, ``{Video-kMaX}: A simple unified approach for online and
  near-online video panoptic segmentation,'' \emph{arXiv preprint
  arXiv:2304.04694}, 2023.

\bibitem{tubelink}
X.~Li, H.~Yuan, W.~Zhang, J.~Pang, G.~Cheng, and C.~C. Loy, ``Tube-link: A
  flexible cross tube baseline for universal video segmentation,'' \emph{ICCV},
  2023.

\bibitem{yao2021efficientdetr}
Z.~Yao, J.~Ai, B.~Li, and C.~Zhang, ``Efficient {DETR}: improving end-to-end
  object detector with dense prior,'' \emph{arXiv preprint arXiv:2104.01318},
  2021.

\bibitem{mp_former}
H.~Zhang, F.~Li, H.~Xu, S.~Huang, S.~Liu, L.~M. Ni, and L.~Zhang,
  ``{MP-Former}: Mask-piloted transformer for image segmentation,''
  \emph{CVPR}, 2023.

\bibitem{wang2022towards}
W.~Wang, J.~Zhang, Y.~Cao, Y.~Shen, and D.~Tao, ``Towards data-efficient
  detection transformers,'' in \emph{ECCV}, 2022.

\bibitem{li2023panopticpartformer++}
X.~Li, S.~Xu, Y.~Yang, H.~Yuan, G.~Cheng, Y.~Tong, Z.~Lin, M.-H. Yang, and
  D.~Tao, ``Panopticpartformer++: A unified and decoupled view for panoptic
  part segmentation,'' \emph{arXiv preprint arXiv:2301.00954}, 2023.

\bibitem{DsHmp}
S.~He and H.~Ding, ``Decoupling static and hierarchical motion perception for
  referring video segmentation,'' in \emph{CVPR}, 2024.

\bibitem{D2Zero}
S.~He, H.~Ding, and W.~Jiang, ``Semantic-promoted debiasing and background
  disambiguation for zero-shot instance segmentation,'' in \emph{CVPR}, 2023.

\bibitem{RIE}
C.~Liu, X.~Li, and H.~Ding, ``Referring image editing: Object-level image
  editing via referring expressions,'' in \emph{CVPR}, 2024.

\bibitem{ISFP}
C.~Liu, X.~Jiang, and H.~Ding, ``Instance-specific feature propagation for
  referring segmentation,'' \emph{TMM}, 2022.

\bibitem{MCN}
G.~Luo, Y.~Zhou, X.~Sun, L.~Cao, C.~Wu, C.~Deng, and R.~Ji, ``Multi-task
  collaborative network for joint referring expression comprehension and
  segmentation,'' in \emph{CVPR}, 2020.

\bibitem{wu2022multilevel_ref_video_seg}
D.~Wu, X.~Dong, L.~Shao, and J.~Shen, ``Multi-level representation learning
  with semantic alignment for referring video object segmentation,'' in
  \emph{CVPR}, 2022.

\bibitem{MDETR}
A.~Kamath, M.~Singh, Y.~LeCun, I.~Misra, G.~Synnaeve, and N.~Carion,
  ``{MDETR}--modulated detection for end-to-end multi-modal understanding,''
  \emph{ICCV}, 2021.

\bibitem{cao2022prototype}
L.~Cao, Y.~Guo, Y.~Yuan, and Q.~Jin, ``Prototype as query for few shot semantic
  segmentation,'' \emph{arXiv preprint arXiv:2211.14764}, 2022.

\bibitem{DIIM}
H.~Ding, H.~Zhang, C.~Liu, and X.~Jiang, ``Deep interactive image matting with
  feature propagation,'' \emph{TIP}, vol.~31, pp. 2421--2432, 2022.

\bibitem{han2023referencetwice}
Y.~Han, J.~Zhang, Z.~Xue, C.~Xu, X.~Shen, Y.~Wang, C.~Wang, Y.~Liu, and X.~Li,
  ``Reference twice: A simple and unified baseline for few-shot instance
  segmentation,'' \emph{arXiv preprint arXiv:2301.01156}, 2023.

\bibitem{guo2021pct}
M.-H. Guo, J.-X. Cai, Z.-N. Liu, T.-J. Mu, R.~R. Martin, and S.-M. Hu, ``{PCT}:
  Point cloud transformer,'' \emph{Computational Visual Media}, 2021.

\bibitem{lai2022stratified}
X.~Lai, J.~Liu, L.~Jiang, L.~Wang, H.~Zhao, S.~Liu, X.~Qi, and J.~Jia,
  ``Stratified transformer for 3d point cloud segmentation,'' in \emph{CVPR},
  2022.

\bibitem{yu2022pointBERT}
X.~Yu, L.~Tang, Y.~Rao, T.~Huang, J.~Zhou, and J.~Lu, ``Point-{BERT}:
  Pre-training {3D} point cloud transformers with masked point modeling,'' in
  \emph{CVPR}, 2022.

\bibitem{pang2022masked}
Y.~Pang, W.~Wang, F.~E.~H. Tay, W.~Liu, Y.~Tian, and L.~Yuan, ``Masked
  autoencoders for point cloud self-supervised learning,'' in \emph{ECCV},
  2022.

\bibitem{zhang2022point_m2ae}
R.~Zhang, Z.~Guo, P.~Gao, R.~Fang, B.~Zhao, D.~Wang, Y.~Qiao, and H.~Li,
  ``{Point-M2AE}: Multi-scale masked autoencoders for hierarchical point cloud
  pre-training,'' \emph{arXiv preprint arXiv:2205.14401}, 2022.

\bibitem{Schult23ICRAMask3D}
J.~Schult, F.~Engelmann, A.~Hermans, O.~Litany, S.~Tang, and B.~Leibe,
  ``{Mask3D for 3D Semantic Instance Segmentation},'' in \emph{ICRA}, 2023.

\bibitem{sun2022superpoint}
J.~Sun, C.~Qing, J.~Tan, and X.~Xu, ``Superpoint transformer for 3d scene
  instance segmentation,'' \emph{AAAI}, 2023.

\bibitem{landrieu2018large_supper_points}
L.~Landrieu and M.~Simonovsky, ``Large-scale point cloud semantic segmentation
  with superpoint graphs,'' in \emph{CVPR}, 2018.

\bibitem{su2023pups}
S.~Su, J.~Xu, H.~Wang, Z.~Miao, X.~Zhan, D.~Hao, and X.~Li, ``{PUPS}: Point
  cloud unified panoptic segmentation,'' \emph{AAAI}, 2023.

\bibitem{semantic_kitti}
J.~Behley, M.~Garbade, A.~Milioto, J.~Quenzel, S.~Behnke, C.~Stachniss, and
  J.~Gall, ``A dataset for semantic segmentation of point cloud sequences,''
  \emph{arXiv preprint arXiv:1904.01416}, vol.~2, no.~3, 2019.

\bibitem{zhou2022cocoop}
K.~Zhou, J.~Yang, C.~C. Loy, and Z.~Liu, ``Conditional prompt learning for
  vision-language models,'' in \emph{CVPR}, 2022.

\bibitem{zhang2021tip_adapter}
R.~Zhang, R.~Fang, P.~Gao, W.~Zhang, K.~Li, J.~Dai, Y.~Qiao, and H.~Li,
  ``{Tip-Adapter}: Training-free clip-adapter for better vision-language
  modeling,'' \emph{ECCV}, 2022.

\bibitem{lin2022frozen}
Z.~Lin, S.~Geng, R.~Zhang, P.~Gao, G.~de~Melo, X.~Wang, J.~Dai, Y.~Qiao, and
  H.~Li, ``Frozen clip models are efficient video learners,'' in \emph{ECCV},
  2022.

\bibitem{chen2022vitadapter}
Z.~Chen, Y.~Duan, W.~Wang, J.~He, T.~Lu, J.~Dai, and Y.~Qiao, ``Vision
  transformer adapter for dense predictions,'' in \emph{ICLR}, 2023.

\bibitem{rao2022denseclip}
Y.~Rao, W.~Zhao, G.~Chen, Y.~Tang, Z.~Zhu, G.~Huang, J.~Zhou, and J.~Lu,
  ``{DenseCLIP}: Language-guided dense prediction with context-aware
  prompting,'' in \emph{CVPR}, 2022.

\bibitem{lueddecke22_clip_seg_prompt}
T.~L\"uddecke and A.~Ecker, ``Image segmentation using text and image
  prompts,'' in \emph{CVPR}, 2022.

\bibitem{jain2022oneformer}
J.~Jain, J.~Li, M.~Chiu, A.~Hassani, N.~Orlov, and H.~Shi, ``{OneFormer}: One
  transformer to rule universal image segmentation,'' in \emph{CVPR}, 2023.

\bibitem{kirillov2023segment}
A.~Kirillov, E.~Mintun, N.~Ravi, H.~Mao, C.~Rolland, L.~Gustafson, T.~Xiao,
  S.~Whitehead, A.~C. Berg, W.-Y. Lo, P.~Dollár, and R.~Girshick, ``Segment
  anything,'' in \emph{ICCV}, 2023.

\bibitem{zareian2021open}
A.~Zareian, K.~D. Rosa, D.~H. Hu, and S.-F. Chang, ``Open-vocabulary object
  detection using captions,'' \emph{CVPR}, 2021.

\bibitem{ViLD}
X.~Gu, T.-Y. Lin, W.~Kuo, and Y.~Cui, ``Open-vocabulary object detection via
  vision and language knowledge distillation,'' in \emph{ICLR}, 2022.

\bibitem{detic}
X.~Zhou, R.~Girdhar, A.~Joulin, P.~Kr{\"a}henb{\"u}hl, and I.~Misra,
  ``Detecting twenty-thousand classes using image-level supervision,'' in
  \emph{ECCV}, 2022.

\bibitem{OV-DETR}
Y.~Zang, W.~Li, K.~Zhou, C.~Huang, and C.~C. Loy, ``Open-vocabulary detr with
  conditional matching,'' in \emph{ECCV}, 2022.

\bibitem{PAD}
S.~He, H.~Ding, and W.~Jiang, ``Primitive generation and semantic-related
  alignment for universal zero-shot segmentation,'' in \emph{CVPR}, 2023.

\bibitem{zhou2023rethinking}
H.~Zhou, T.~Shen, X.~Yang, H.~Huang, X.~Li, L.~Qi, and M.-H. Yang, ``Rethinking
  evaluation metrics of open-vocabulary segmentaion,'' \emph{arXiv preprint
  arXiv:2311.03352}, 2023.

\bibitem{kuo2022fvlm}
W.~Kuo, Y.~Cui, X.~Gu, A.~Piergiovanni, and A.~Angelova, ``{F-VLM}:
  Open-vocabulary object detection upon frozen vision and language models,'' in
  \emph{ICLR}, 2023.

\bibitem{Maaz2022Multimodal}
M.~Maaz, H.~Rasheed, S.~Khan, F.~S. Khan, R.~M. Anwer, and M.-H. Yang,
  ``Class-agnostic object detection with multi-modal transformer,'' in
  \emph{ECCV}, 2022.

\bibitem{OpenSeg}
G.~Ghiasi, X.~Gu, Y.~Cui, and T.-Y. Lin, ``Scaling open-vocabulary image
  segmentation with image-level labels,'' in \emph{ECCV}, 2022.

\bibitem{LSeg}
B.~Li, K.~Q. Weinberger, S.~Belongie, V.~Koltun, and R.~Ranftl,
  ``Language-driven semantic segmentation,'' in \emph{ICLR}, 2022.

\bibitem{wu2023betrayed}
J.~Wu, X.~Li, H.~Ding, X.~Li, G.~Cheng, Y.~Tong, and C.~C. Loy, ``Betrayed by
  captions: Joint caption grounding and generation for open vocabulary instance
  segmentation,'' in \emph{ICCV}, 2023.

\bibitem{qin2023freeseg}
J.~Qin, J.~Wu, P.~Yan, M.~Li, R.~Yuxi, X.~Xiao, Y.~Wang, R.~Wang, S.~Wen,
  X.~Pan \emph{et~al.}, ``{FreeSeg}: Unified, universal and open-vocabulary
  image segmentation,'' \emph{CVPR}, 2023.

\bibitem{wang2021unidentified}
W.~Wang, M.~Feiszli, H.~Wang, and D.~Tran, ``Unidentified video objects: A
  benchmark for dense, open-world segmentation,'' in \emph{ICCV}, 2021.

\bibitem{xu2023odise}
J.~Xu, S.~Liu, A.~Vahdat, W.~Byeon, X.~Wang, and S.~De~Mello,
  ``{Open-Vocabulary Panoptic Segmentation with Text-to-Image Diffusion
  Models},'' \emph{CVPR}, 2023.

\bibitem{gupta2021ow}
A.~Gupta, S.~Narayan, K.~Joseph, S.~Khan, F.~S. Khan, and M.~Shah, ``{OW-DETR}:
  Open-world detection transformer,'' in \emph{CVPR}, 2022.

\bibitem{xu2023side}
M.~Xu, Z.~Zhang, F.~Wei, H.~Hu, and X.~Bai, ``Side adapter network for
  open-vocabulary semantic segmentation,'' \emph{CVPR}, 2023.

\bibitem{liu2020open}
Z.~Liu, Z.~Miao, X.~Pan, X.~Zhan, D.~Lin, S.~X. Yu, and B.~Gong, ``Open
  compound domain adaptation,'' in \emph{CVPR}, 2020.

\bibitem{yang2020fda}
Y.~Yang and S.~Soatto, ``{FDA}: Fourier domain adaptation for semantic
  segmentation,'' in \emph{CVPR}, 2020.

\bibitem{hoyer2022daformer}
L.~Hoyer, D.~Dai, and L.~Van~Gool, ``{DAFormer}: Improving network
  architectures and training strategies for domain-adaptive semantic
  segmentation,'' in \emph{CVPR}, 2022.

\bibitem{hoyer2022hrda}
------, ``{HRDA}: Context-aware high-resolution domain-adaptive semantic
  segmentation,'' in \emph{ECCV}, 2022.

\bibitem{hoyer2022mic}
L.~Hoyer, D.~Dai, H.~Wang, and L.~Van~Gool, ``{MIC}: Masked image consistency
  for context-enhanced domain adaptation,'' in \emph{CVPR}, 2023.

\bibitem{wang2021exploring}
W.~Wang, Y.~Cao, J.~Zhang, F.~He, Z.-J. Zha, Y.~Wen, and D.~Tao, ``Exploring
  sequence feature alignment for domain adaptive detection transformers,'' in
  \emph{ACM-MM}, 2021.

\bibitem{zhang2021da_detr}
J.~Zhang, J.~Huang, Z.~Luo, G.~Zhang, and S.~Lu, ``{DA-DETR}: Domain adaptive
  detection transformer by hybrid attention,'' \emph{arXiv preprint
  arXiv:2103.17084}, 2021.

\bibitem{yu2022mttrans}
J.~Yu, J.~Liu, X.~Wei, H.~Zhou, Y.~Nakata, D.~Gudovskiy, T.~Okuno, J.~Li,
  K.~Keutzer, and S.~Zhang, ``{MTTrans}: Cross-domain object detection with
  mean teacher transformer,'' in \emph{ECCV}, 2022.

\bibitem{xie2023sepico}
B.~Xie, S.~Li, M.~Li, C.~H. Liu, G.~Huang, and G.~Wang, ``Sepico:
  Semantic-guided pixel contrast for domain adaptive semantic segmentation,''
  \emph{PAMI}, 2023.

\bibitem{lambert2020mseg}
J.~Lambert, Z.~Liu, O.~Sener, J.~Hays, and V.~Koltun, ``{MSeg}: A composite
  dataset for multi-domain semantic segmentation,'' in \emph{CVPR}, 2020.

\bibitem{yin2022devil}
W.~Yin, Y.~Liu, C.~Shen, A.~v.~d. Hengel, and B.~Sun, ``The devil is in the
  labels: Semantic segmentation from sentences,'' \emph{arXiv preprint
  arXiv:2202.02002}, 2022.

\bibitem{lmseg}
Z.~Qiang, L.~Yuang, L.~Yuang, Y.~Chaohui, L.~Jingliang, W.~Zhibin, and W.~Fan,
  ``{LMSeg}: Language-guided multi-dataset segmentation,'' \emph{ICLR}, 2023.

\bibitem{Zhou_2022_CVPR_mutl_data_det}
X.~Zhou, V.~Koltun, and P.~Kr\"ahenb\"uhl, ``Simple multi-dataset detection,''
  in \emph{CVPR}, 2022.

\bibitem{meng2022detectionhub}
L.~Meng, X.~Dai, Y.~Chen, P.~Zhang, D.~Chen, M.~Liu, J.~Wang, Z.~Wu, L.~Yuan,
  and Y.-G. Jiang, ``Detection hub: Unifying object detection datasets via
  query adaptation on language embedding,'' \emph{CVPR}, 2023.

\bibitem{hoyer2023domain}
L.~Hoyer, D.~Dai, and L.~Van~Gool, ``Domain adaptive and generalizable network
  architectures and training strategies for semantic image segmentation,''
  \emph{arXiv preprint arXiv:2304.13615}, 2023.

\bibitem{zhao2023style}
Y.~Zhao, Z.~Zhong, N.~Zhao, N.~Sebe, and G.~H. Lee, ``Style-hallucinated dual
  consistency learning: A unified framework for visual domain generalization,''
  \emph{IJCV}, 2023.

\bibitem{xu2022multitoke_wsss}
L.~Xu, W.~Ouyang, M.~Bennamoun, F.~Boussaid, and D.~Xu, ``Multi-class token
  transformer for weakly supervised semantic segmentation,'' in \emph{CVPR},
  2022.

\bibitem{wang2020self}
Y.~Wang, J.~Zhang, M.~Kan, S.~Shan, and X.~Chen, ``Self-supervised equivariant
  attention mechanism for weakly supervised semantic segmentation,'' in
  \emph{CVPR}, 2020.

\bibitem{rossetti2022max}
S.~Rossetti, D.~Zappia, M.~Sanzari, M.~Schaerf, and F.~Pirri, ``Max pooling
  with vision transformers reconciles class and shape in weakly supervised
  semantic segmentation,'' in \emph{ECCV}, 2022.

\bibitem{hsu2019bbtp}
C.-C. Hsu, K.-J. Hsu, C.-C. Tsai, Y.-Y. Lin, and Y.-Y. Chuang, ``Weakly
  supervised instance segmentation using the bounding box tightness prior,''
  \emph{NeurIPS}, 2019.

\bibitem{lan2021discobox}
S.~Lan, Z.~Yu, C.~Choy, S.~Radhakrishnan, G.~Liu, Y.~Zhu, L.~S. Davis, and
  A.~Anandkumar, ``{DiscoBox}: Weakly supervised instance segmentation and
  semantic correspondence from box supervision,'' in \emph{CVPR}, 2021.

\bibitem{tian2021boxinst}
Z.~Tian, C.~Shen, X.~Wang, and H.~Chen, ``{BoxInst}: High-performance instance
  segmentation with box annotations,'' in \emph{CVPR}, 2021.

\bibitem{xu2022groupvit}
J.~Xu, S.~De~Mello, S.~Liu, W.~Byeon, T.~Breuel, J.~Kautz, and X.~Wang,
  ``Groupvit: Semantic segmentation emerges from text supervision,'' in
  \emph{CVPR}, 2022.

\bibitem{yi2023simple}
M.~Yi, Q.~Cui, H.~Wu, C.~Yang, O.~Yoshie, and H.~Lu, ``A simple framework for
  text-supervised semantic segmentation,'' in \emph{CVPR}, 2023.

\bibitem{maskfreevis}
L.~Ke, M.~Danelljan, H.~Ding, Y.-W. Tai, C.-K. Tang, and F.~Yu, ``Mask-free
  video instance segmentation,'' in \emph{CVPR}, 2023.

\bibitem{van2021unsupervised}
W.~Van~Gansbeke, S.~Vandenhende, S.~Georgoulis, and L.~Van~Gool, ``Unsupervised
  semantic segmentation by contrasting object mask proposals,'' in \emph{ICCV},
  2021.

\bibitem{simeoni2021localizingLOST}
O.~Sim{\'e}oni, G.~Puy, H.~V. Vo, S.~Roburin, S.~Gidaris, A.~Bursuc,
  P.~P{\'e}rez, R.~Marlet, and J.~Ponce, ``Localizing objects with
  self-supervised transformers and no labels,'' in \emph{BMVC}, 2021.

\bibitem{hamilton2022unsupervisedSETGO}
M.~Hamilton, Z.~Zhang, B.~Hariharan, N.~Snavely, and W.~T. Freeman,
  ``Unsupervised semantic segmentation by distilling feature correspondences,''
  \emph{ICLR}, 2022.

\bibitem{shin2022reco}
G.~Shin, W.~Xie, and S.~Albanie, ``{ReCo}: Retrieve and co-segment for
  zero-shot transfer,'' in \emph{NeurIPS}, 2022.

\bibitem{van2022discoveringMaskDistill}
W.~Van~Gansbeke, S.~Vandenhende, and L.~Van~Gool, ``Discovering object masks
  with transformers for unsupervised semantic segmentation,'' \emph{arXiv
  preprint arXiv:2206.06363}, 2022.

\bibitem{wang2022freesolo}
X.~Wang, Z.~Yu, S.~De~Mello, J.~Kautz, A.~Anandkumar, C.~Shen, and J.~M.
  Alvarez, ``{FreeSOLO}: Learning to segment objects without annotations,'' in
  \emph{CVPR}, 2022.

\bibitem{wang2023cut}
X.~Wang, R.~Girdhar, S.~X. Yu, and I.~Misra, ``Cut and learn for unsupervised
  object detection and instance segmentation,'' in \emph{CVPR}, 2023.

\bibitem{zhao2017icnet}
H.~Zhao, X.~Qi, X.~Shen, J.~Shi, and J.~Jia, ``{ICNet} for real-time semantic
  segmentation on high-resolution images,'' \emph{ECCV}, 2018.

\bibitem{maaz2023edgenext}
M.~Maaz, A.~Shaker, H.~Cholakkal, S.~Khan, S.~W. Zamir, R.~M. Anwer, and
  F.~Shahbaz~Khan, ``{EdgeNeXt}: efficiently amalgamated cnn-transformer
  architecture for mobile vision applications,'' in \emph{ECCV Workshops},
  2022.

\bibitem{mehta2021mobilevit}
S.~Mehta and M.~Rastegari, ``Mobilevit: light-weight, general-purpose, and
  mobile-friendly vision transformer,'' in \emph{ICLR}, 2022.

\bibitem{zhang2023rethinking}
J.~Zhang, X.~Li, J.~Li, L.~Liu, Z.~Xue, B.~Zhang, Z.~Jiang, T.~Huang, Y.~Wang,
  and C.~Wang, ``Rethinking mobile block for efficient neural models,''
  \emph{arXiv preprint arXiv:2301.01146}, 2023.

\bibitem{liang2022expediting}
W.~Liang, Y.~Yuan, H.~Ding, X.~Luo, W.~Lin, D.~Jia, Z.~Zhang, C.~Zhang, and
  H.~Hu, ``Expediting large-scale vision transformer for dense prediction
  without fine-tuning,'' in \emph{NeurIPS}, 2022.

\bibitem{zhou2023edgesam}
C.~Zhou, X.~Li, C.~C. Loy, and B.~Dai, ``Edgesam: Prompt-in-the-loop
  distillation for on-device deployment of sam,'' \emph{arXiv preprint
  arXiv:2312.06660}, 2023.

\bibitem{xu2024rapsam}
S.~Xu, H.~Yuan, Q.~Shi, L.~Qi, J.~Wang, Y.~Yang, Y.~Li, K.~Chen, Y.~Tong,
  B.~Ghanem, X.~Li, and M.-H. Yang, ``Rap-sam: Towards real-time all-purpose
  segment anything,'' \emph{arXiv preprint}, 2024.

\bibitem{zhang2022topformer}
W.~Zhang, Z.~Huang, G.~Luo, T.~Chen, X.~Wang, W.~Liu, G.~Yu, and C.~Shen,
  ``{TopFormer}: Token pyramid transformer for mobile semantic segmentation,''
  in \emph{CVPR}, 2022.

\bibitem{wan2023seaformer}
Q.~Wan, Z.~Huang, J.~Lu, G.~Yu, and L.~Zhang, ``{SeaFormer}: Squeeze-enhanced
  axial transformer for mobile semantic segmentation,'' in \emph{ICLR}, 2023.

\bibitem{cheng2020cascadepsp}
H.~K. Cheng, J.~Chung, Y.-W. Tai, and C.-K. Tang, ``{CascadePSP}: Toward
  class-agnostic and very high-resolution segmentation via global and local
  refinement,'' in \emph{CVPR}, 2020.

\bibitem{transfiner}
L.~Ke, M.~Danelljan, X.~Li, Y.-W. Tai, C.-K. Tang, and F.~Yu, ``Mask transfiner
  for high-quality instance segmentation,'' in \emph{CVPR}, 2022.

\bibitem{video_transfiner}
L.~Ke, H.~Ding, M.~Danelljan, Y.-W. Tai, C.-K. Tang, and F.~Yu, ``Video mask
  transfiner for high-quality video instance segmentation,'' in \emph{ECCV},
  2022.

\bibitem{liu2022simpleclick}
Q.~Liu, Z.~Xu, G.~Bertasius, and M.~Niethammer, ``{SimpleClick}: Interactive
  image segmentation with simple vision transformers,'' \emph{arXiv preprint
  arXiv:2210.11006}, 2022.

\bibitem{SegRefiner}
M.~Wang, H.~Ding, J.~H. Liew, J.~Liu, Y.~Zhao, and Y.~Wei, ``Segrefiner:
  Towards model-agnostic segmentation refinement with discrete diffusion
  process,'' in \emph{NeurIPS}, 2023.

\bibitem{song2024basam}
Y.~Song, Q.~Zhou, X.~Li, D.-P. Fan, X.~Lu, and L.~Ma, ``Ba-sam: Scalable
  bias-mode attention mask for segment anything model,'' \emph{arXiv preprint
  arXiv:2401.02317}, 2024.

\bibitem{wen2023patchdct}
Q.~Wen, J.~Yang, X.~Yang, and K.~Liang, ``Patchdct: Patch refinement for high
  quality instance segmentation,'' \emph{ICLR}, 2023.

\bibitem{shen2021dct}
X.~Shen, J.~Yang, C.~Wei, B.~Deng, J.~Huang, X.-S. Hua, X.~Cheng, and K.~Liang,
  ``Dct-mask: Discrete cosine transform mask representation for instance
  segmentation,'' in \emph{CVPR}, 2021.

\bibitem{qi2022openentity}
L.~Qi, J.~Kuen, Y.~Wang, J.~Gu, H.~Zhao, P.~Torr, Z.~Lin, and J.~Jia, ``Open
  world entity segmentation,'' \emph{TPAMI}, 2022.

\bibitem{stm_vos}
S.~W. Oh, J.-Y. Lee, N.~Xu, and S.~J. Kim, ``Video object segmentation using
  space-time memory networks,'' in \emph{ICCV}, 2019.

\bibitem{cheng2022xmem}
H.~K. Cheng and A.~G. Schwing, ``{XMem}: Long-term video object segmentation
  with an atkinson-shiffrin memory model,'' in \emph{ECCV}, 2022.

\bibitem{park2022perclip_vos}
K.~Park, S.~Woo, S.~W. Oh, I.~S. Kweon, and J.-Y. Lee, ``Per-clip video object
  segmentation,'' in \emph{CVPR}, 2022.

\bibitem{Wang2022LookBY}
J.~Wang, D.~Chen, Z.~Wu, C.~Luo, C.~Tang, X.~Dai, Y.~Zhao, Y.~Xie, L.~Yuan, and
  Y.-G. Jiang, ``Look before you match: Instance understanding matters in video
  object segmentation,'' in \emph{CVPR}, 2023.

\bibitem{ronneberger2015_unet}
O.~Ronneberger, P.~Fischer, and T.~Brox, ``{U-Net}: Convolutional networks for
  biomedical image segmentation,'' in \emph{MICCAI}, 2015.

\bibitem{isensee2021nnu}
F.~Isensee, P.~F. Jaeger, S.~A. Kohl, J.~Petersen, and K.~H. Maier-Hein,
  ``{nnU-Net}: a self-configuring method for deep learning-based biomedical
  image segmentation,'' \emph{Nature methods}, 2021.

\bibitem{chen2021transunet}
J.~Chen, Y.~Lu, Q.~Yu, X.~Luo, E.~Adeli, Y.~Wang, L.~Lu, A.~L. Yuille, and
  Y.~Zhou, ``Transunet: Transformers make strong encoders for medical image
  segmentation,'' \emph{arXiv preprint arXiv:2102.04306}, 2021.

\bibitem{cao2021swinunet}
H.~Cao, Y.~Wang, J.~Chen, D.~Jiang, X.~Zhang, Q.~Tian, and M.~Wang,
  ``{Swin-Unet}: Unet-like pure transformer for medical image segmentation,''
  in \emph{ECCV Workshops}, 2022.

\bibitem{zhang2021transfuse}
Y.~Zhang, H.~Liu, and Q.~Hu, ``Transfuse: Fusing transformers and cnns for
  medical image segmentation,'' in \emph{MICCAI}.\hskip 1em plus 0.5em minus
  0.4em\relax Springer, 2021.

\bibitem{hatamizadeh2022unetr}
A.~Hatamizadeh, Y.~Tang, V.~Nath, D.~Yang, A.~Myronenko, B.~Landman, H.~R.
  Roth, and D.~Xu, ``{UNETR}: Transformers for 3d medical image segmentation,''
  in \emph{WACV}, 2022.

\bibitem{zhang2023simple}
H.~Zhang, F.~Li, X.~Zou, S.~Liu, C.~Li, J.~Yang, and L.~Zhang, ``A simple
  framework for open-vocabulary segmentation and detection,'' in \emph{ICCV},
  2023.

\bibitem{zou2023segment}
X.~Zou, J.~Yang, H.~Zhang, F.~Li, L.~Li, J.~Gao, and Y.~J. Lee, ``Segment
  everything everywhere all at once,'' \emph{arXiv preprint arXiv:2304.06718},
  2023.

\bibitem{HIPIE}
X.~Wang, S.~Li, K.~Kallidromitis, Y.~Kato, K.~Kozuka, and T.~Darrell,
  ``Hierarchical open-vocabulary universal image segmentation,''
  \emph{NeurIPS}, 2023.

\bibitem{objects365}
S.~Shao, Z.~Li, T.~Zhang, C.~Peng, G.~Yu, X.~Zhang, J.~Li, and J.~Sun,
  ``Objects365: A large-scale, high-quality dataset for object detection,'' in
  \emph{ICCV}, 2019.

\bibitem{liang2023clustseg}
J.~Liang, T.~Zhou, D.~Liu, and W.~Wang, ``Clustseg: Clustering for universal
  segmentation,'' \emph{ICML}, 2023.

\bibitem{sun2022vss}
G.~Sun, Y.~Liu, H.~Ding, T.~Probst, and L.~Van~Gool, ``Coarse-to-fine feature
  mining for video semantic segmentation,'' in \emph{CVPR}, 2022.

\bibitem{sun2022mining}
G.~Sun, Y.~Liu, H.~Tang, A.~Chhatkuli, L.~Zhang, and L.~Van~Gool, ``Mining
  relations among cross-frame affinities for video semantic segmentation,''
  \emph{ECCV}, 2022.

\bibitem{hu2023you}
J.~Hu, L.~Huang, T.~Ren, S.~Zhang, R.~Ji, and L.~Cao, ``You only segment once:
  Towards real-time panoptic segmentation,'' in \emph{CVPR}, 2023.

\bibitem{ying2023ctvis}
K.~Ying, Q.~Zhong, W.~Mao, Z.~Wang, H.~Chen, L.~Y. Wu, Y.~Liu, C.~Fan,
  Y.~Zhuge, and C.~Shen, ``Ctvis: Consistent training for online video instance
  segmentation,'' in \emph{ICCV}, 2023.

\bibitem{heo2022generalized}
M.~Heo, S.~Hwang, J.~Hyun, H.~Kim, S.~W. Oh, J.-Y. Lee, and S.~J. Kim, ``A
  generalized framework for video instance segmentation,'' in \emph{CVPR},
  2023.

\bibitem{zhou2022slot}
Y.~Zhou, H.~Zhang, H.~Lee, S.~Sun, P.~Li, Y.~Zhu, B.~Yoo, X.~Qi, and J.-J. Han,
  ``{Slot-VPS}: Object-centric representation learning for video panoptic
  segmentation,'' in \emph{CVPR}, 2022.

\bibitem{phraseclick}
H.~Ding, S.~Cohen, B.~Price, and X.~Jiang, ``Phraseclick: toward achieving
  flexible interactive segmentation by phrase and click,'' in
  \emph{ECCV}.\hskip 1em plus 0.5em minus 0.4em\relax Springer, 2020, pp.
  417--435.

\bibitem{lu2022unified}
J.~Lu, C.~Clark, R.~Zellers, R.~Mottaghi, and A.~Kembhavi, ``{Unified-IO}: A
  unified model for vision, language, and multi-modal tasks,'' \emph{ICLR},
  2023.

\bibitem{qi2024generalizable}
L.~Qi, Y.-W. Chen, L.~Yang, T.~Shen, X.~Li, W.~Guo, Y.~Xu, and M.-H. Yang,
  ``Generalizable entity grounding via assistance of large language model,''
  \emph{arXiv preprint arXiv:2402.02555}, 2024.

\bibitem{yuan2024open}
H.~Yuan, X.~Li, C.~Zhou, Y.~Li, K.~Chen, and C.~C. Loy, ``Open-vocabulary sam:
  Segment and recognize twenty-thousand classes interactively,'' \emph{arXiv
  preprint arXiv:2401.02955}, 2024.

\bibitem{zhang2021prototypical}
H.~Zhang and H.~Ding, ``Prototypical matching and open set rejection for
  zero-shot semantic segmentation,'' in \emph{ICCV}, 2021.

\bibitem{zhou2024dvis}
Y.~Zhou, T.~Zhang, S.~Ji, S.~Yan, and X.~Li, ``Dvis-daq: Improving video
  segmentation via dynamic anchor queries,'' \emph{arXiv preprint
  arXiv:2404.00086}, 2024.

\bibitem{chen2022generalist}
T.~Chen, L.~Li, S.~Saxena, G.~Hinton, and D.~J. Fleet, ``A generalist framework
  for panoptic segmentation of images and videos,'' \emph{arXiv preprint
  arXiv:2210.06366}, 2022.

\bibitem{wang2024explore}
C.~Wang, X.~Li, H.~Ding, L.~Qi, J.~Zhang, Y.~Tong, C.~C. Loy, and S.~Yan,
  ``Explore in-context segmentation via latent diffusion models,'' \emph{arXiv
  preprint arXiv:2403.09616}, 2024.

\bibitem{xie2023mosaicfusion}
J.~Xie, W.~Li, X.~Li, Z.~Liu, Y.~S. Ong, and C.~C. Loy, ``Mosaicfusion:
  Diffusion models as data augmenters for large vocabulary instance
  segmentation,'' \emph{arXiv preprint arXiv:2309.13042}, 2023.

\bibitem{rombach2021highresolution}
R.~Rombach, A.~Blattmann, D.~Lorenz, P.~Esser, and B.~Ommer, ``High-resolution
  image synthesis with latent diffusion models,'' in \emph{CVPR}, 2022.

\bibitem{johnson2015image}
J.~Johnson, R.~Krishna, M.~Stark, L.-J. Li, D.~Shamma, M.~Bernstein, and
  L.~Fei-Fei, ``Image retrieval using scene graphs,'' in \emph{CVPR}, 2015.

\bibitem{pvsg}
J.~Yang, W.~Peng, X.~Li, Z.~Guo, L.~Chen, B.~Li, Z.~Ma, K.~Zhou, W.~Zhang,
  C.~C. Loy, and Z.~Liu, ``Panoptic video scene graph generation,'' in
  \emph{CVPR}, 2023.

\bibitem{yang_psg}
J.~Yang, Y.~Z. Ang, Z.~Guo, K.~Zhou, W.~Zhang, and Z.~Liu, ``Panoptic scene
  graph generation,'' in \emph{ECCV}, 2022.

\bibitem{psg4d}
J.~Yang, J.~CEN, W.~PENG, S.~Liu, F.~Hong, X.~Li, K.~Zhou, Q.~Chen, and Z.~Liu,
  ``4d panoptic scene graph generation,'' in \emph{NeurIPS}, 2023.

\bibitem{wang2023pair}
J.~Wang, Z.~Wen, X.~Li, Z.~Guo, J.~Yang, and Z.~Liu, ``Pair then relation:
  Pair-net for panoptic scene graph generation,'' \emph{arXiv preprint
  arXiv:2307.08699}, 2023.

\bibitem{luiten2021hota}
J.~Luiten, A.~Osep, P.~Dendorfer, P.~Torr, A.~Geiger, L.~Leal-Taix{\'e}, and
  B.~Leibe, ``{HOTA}: A higher order metric for evaluating multi-object
  tracking,'' \emph{IJCV}, 2021.

\bibitem{kolodiazhnyi2023oneformer3d}
M.~Kolodiazhnyi, A.~Vorontsova, A.~Konushin, and D.~Rukhovich, ``Oneformer3d:
  One transformer for unified point cloud segmentation,'' \emph{arXiv preprint
  arXiv:2311.14405}, 2023.

\bibitem{zhao2021point}
H.~Zhao, L.~Jiang, J.~Jia, P.~H. Torr, and V.~Koltun, ``Point transformer,'' in
  \emph{ICCV}, 2021.

\bibitem{wu2022point}
X.~Wu, Y.~Lao, L.~Jiang, X.~Liu, and H.~Zhao, ``Point transformer v2: Grouped
  vector attention and partition-based pooling,'' in \emph{NeurIPS}, 2022.

\bibitem{schult2023mask3d}
J.~Schult, F.~Engelmann, A.~Hermans, O.~Litany, S.~Tang, and B.~Leibe,
  ``Mask3d: Mask transformer for 3d semantic instance segmentation,'' in
  \emph{ICRA}, 2023.

\bibitem{sun2023superpoint}
J.~Sun, C.~Qing, J.~Tan, and X.~Xu, ``Superpoint transformer for 3d scene
  instance segmentation,'' in \emph{AAAI}, 2023.

\bibitem{wu2023open}
J.~Wu, X.~Li, S.~Xu, H.~Yuan, H.~Ding, Y.~Yang, X.~Li, J.~Zhang, Y.~Tong,
  X.~Jiang, B.~Ghanem, and D.~Tao, ``Towards open vocabulary learning: A
  survey,'' \emph{arXiv pre-print}, 2023.

\bibitem{ding2022decoupling}
J.~Ding, N.~Xue, G.-S. Xia, and D.~Dai, ``Decoupling zero-shot semantic
  segmentation,'' in \emph{CVPR}, 2022.

\bibitem{xu2022_simple_baseline_open_voc}
M.~Xu, Z.~Zhang, F.~Wei, Y.~Lin, Y.~Cao, H.~Hu, and X.~Bai, ``A simple baseline
  for open-vocabulary semantic segmentation with pre-trained vision-language
  model,'' in \emph{ECCV}, 2022.

\bibitem{zhou2021denseclip}
C.~Zhou, C.~C. Loy, and B.~Dai, ``{DenseCLIP}: Extract free dense labels from
  clip,'' \emph{ECCV}, 2022.

\bibitem{panoptic-MaskCLIP}
Z.~Ding, J.~Wang, and Z.~Tu, ``Open-vocabulary panoptic segmentation with
  maskclip,'' \emph{ICML}, 2023.

\bibitem{Mask-adapted-clip}
F.~Liang, B.~Wu, X.~Dai, K.~Li, Y.~Zhao, H.~Zhang, P.~Zhang, P.~Vajda, and
  D.~Marculescu, ``Open-vocabulary semantic segmentation with mask-adapted
  clip,'' \emph{arXiv preprint arXiv:2210.04150}, 2022.

\bibitem{global-knowledge-calibration}
K.~Han, Y.~Liu, J.~H. Liew, H.~Ding, Y.~Wei, J.~Liu, Y.~Wang, Y.~Tang, Y.~Yang,
  J.~Feng \emph{et~al.}, ``Global knowledge calibration for fast
  open-vocabulary segmentation,'' \emph{ICCV}, 2023.

\bibitem{side-adapter}
M.~Xu, Z.~Zhang, F.~Wei, H.~Hu, and X.~Bai, ``Side adapter network for
  open-vocabulary semantic segmentation,'' \emph{CVPR}, 2023.

\bibitem{conceptual-captions}
P.~Sharma, N.~Ding, S.~Goodman, and R.~Soricut, ``Conceptual captions: A
  cleaned, hypernymed, image alt-text dataset for automatic image captioning,''
  in \emph{ACL)}, 2018.

\bibitem{SBU}
V.~Ordonez, G.~Kulkarni, and T.~Berg, ``Im2text: Describing images using 1
  million captioned photographs,'' \emph{NeurIPS}, 2011.

\bibitem{visual-genome}
R.~Krishna, Y.~Zhu, O.~Groth, J.~Johnson, K.~Hata, J.~Kravitz, S.~Chen,
  Y.~Kalantidis, L.-J. Li, D.~A. Shamma \emph{et~al.}, ``Visual genome:
  Connecting language and vision using crowdsourced dense image annotations,''
  \emph{IJCV}, 2017.

\bibitem{coco-captions}
X.~Chen, H.~Fang, T.-Y. Lin, R.~Vedantam, S.~Gupta, P.~Doll{\'a}r, and C.~L.
  Zitnick, ``Microsoft coco captions: Data collection and evaluation server,''
  \emph{CVPR}, 2015.

\bibitem{XPM}
D.~Huynh, J.~Kuen, Z.~Lin, J.~Gu, and E.~Elhamifar, ``Open-vocabulary instance
  segmentation via robust cross-modal pseudo-labeling,'' in \emph{CVPR}, 2022.

\bibitem{mask-free-OVIS}
V.~VS, N.~Yu, C.~Xing, C.~Qin, M.~Gao, J.~C. Niebles, V.~M. Patel, and R.~Xu,
  ``Mask-free ovis: Open-vocabulary instance segmentation without manual mask
  annotations,'' \emph{CVPR}, 2023.

\bibitem{xu2022simple}
M.~Xu, Z.~Zhang, F.~Wei, Y.~Lin, Y.~Cao, H.~Hu, and X.~Bai, ``A simple baseline
  for open-vocabulary semantic segmentation with pre-trained vision-language
  model,'' in \emph{ECCV}, 2022.

\bibitem{ru2022weakly}
L.~Ru, B.~Du, Y.~Zhan, and C.~Wu, ``Weakly-supervised semantic segmentation
  with visual words learning and hybrid pooling,'' \emph{IJCV}, 2022.

\bibitem{xu2023mctformer_p}
L.~Xu, M.~Bennamoun, F.~Boussaid, H.~Laga, W.~Ouyang, and D.~Xu, ``Mctformer+:
  Multi-class token transformer for weakly supervised semantic segmentation,''
  \emph{arXiv preprint arXiv:2308.03005}, 2023.

\bibitem{ru2023token}
L.~Ru, H.~Zheng, Y.~Zhan, and B.~Du, ``Token contrast for weakly-supervised
  semantic segmentation,'' in \emph{CVPR}, 2023.

\bibitem{zhu2023weaktr}
L.~Zhu, Y.~Li, J.~Fang, Y.~Liu, H.~Xin, W.~Liu, and X.~Wang, ``Weaktr:
  Exploring plain vision transformer for weakly-supervised semantic
  segmentation,'' \emph{arXiv preprint arXiv:2304.01184}, 2023.

\bibitem{kim2023causal}
J.~Kim, B.-K. Lee, and Y.~M. Ro, ``Causal unsupervised semantic segmentation,''
  \emph{arXiv preprint arXiv:2310.07379}, 2023.

\bibitem{ghiasi2021simple}
G.~Ghiasi, Y.~Cui, A.~Srinivas, R.~Qian, T.-Y. Lin, E.~D. Cubuk, Q.~V. Le, and
  B.~Zoph, ``Simple copy-paste is a strong data augmentation method for
  instance segmentation,'' in \emph{CVPR}, 2021.

\bibitem{mmseg2020}
M.~Contributors, ``{MMSegmentation}: Openmmlab semantic segmentation toolbox
  and benchmark,'' \url{https://github.com/open-mmlab/mmsegmentation}, 2020.

\bibitem{chen2019mmdetection}
K.~Chen, J.~Wang, J.~Pang, Y.~Cao, Y.~Xiong, X.~Li, S.~Sun, W.~Feng, Z.~Liu,
  J.~Xu \emph{et~al.}, ``Mmdetection: Open mmlab detection toolbox and
  benchmark,'' \emph{arXiv preprint}, 2019.

\bibitem{dong2023afformer}
D.~Bo, W.~Pichao, and F.~Wang, ``Afformer: Head-free lightweight semantic
  segmentation with linear transformer,'' in \emph{AAAI}, 2023.

\bibitem{lan2023vision}
S.~Lan, X.~Yang, Z.~Yu, Z.~Wu, J.~M. Alvarez, and A.~Anandkumar, ``Vision
  transformers are good mask auto-labelers,'' \emph{arXiv preprint
  arXiv:2301.03992}, 2023.

\bibitem{zhao2022shade}
Y.~Zhao, Z.~Zhong, N.~Zhao, N.~Sebe, and G.~H. Lee, ``Style-hallucinated dual
  consistency learning for domain generalized semantic segmentation,'' in
  \emph{ECCV}, 2022.

\bibitem{choi2021robustnet}
S.~Choi, S.~Jung, H.~Yun, J.~T. Kim, S.~Kim, and J.~Choo, ``Robustnet:
  Improving domain generalization in urban-scene segmentation via instance
  selective whitening,'' in \emph{CVPR}, 2021.

\bibitem{zhao2022shadevdg}
Y.~Zhao, Z.~Zhong, N.~Zhao, N.~Sebe, and G.~H. Lee, ``Style-hallucinated dual
  consistency learning: A unified framework for visual domain generalization,''
  \emph{arXiv preprint arXiv:2212.09068}, 2022.

\bibitem{hong20224ddsnet}
F.~Hong, H.~Zhou, X.~Zhu, H.~Li, and Z.~Liu, ``Lidar-based 4d panoptic
  segmentation via dynamic shifting network,'' \emph{arXiv preprint
  arXiv:2203.07186}, 2022.

\end{thebibliography}
}

\newpage
\end{document}